\documentclass[11pt]{article}
\usepackage[preprint]{acl}
\usepackage{inconsolata}

\usepackage[utf8]{inputenc}
\usepackage[T1]{fontenc}
\usepackage{booktabs}
\usepackage{amsfonts}
\usepackage{amsmath, amssymb}
\usepackage{amsthm}
\usepackage{nicefrac}
\usepackage{microtype}
\usepackage{graphicx}
\usepackage{multirow}
\usepackage{float}
\usepackage{enumitem}
\usepackage{array}
\usepackage{etoolbox}

\usepackage{bm}
\usepackage{tikz}
\usetikzlibrary{arrows.meta, positioning, shapes.geometric, fit, calc}

\definecolor{promptframe}{HTML}{3B4CCA}
\definecolor{promptback}{HTML}{F4F5FC}
\definecolor{prompttitle}{HTML}{2F3EA8}
\newsavebox{\promptboxsave}
\newenvironment{promptbox}[1]
  {\par\addvspace{8pt}\noindent
   \setlength{\fboxsep}{4pt}%
   \colorbox{prompttitle}{\color{white}\bfseries\footnotesize\strut #1}%
   \par\nobreak\vspace{-0.4pt}%
   \begin{lrbox}{\promptboxsave}%
   \begin{minipage}{\dimexpr\columnwidth-2\fboxsep-2\fboxrule\relax}%
   \ttfamily\scriptsize}
  {\end{minipage}%
   \end{lrbox}%
   \noindent\setlength{\fboxrule}{0.5pt}%
   \fcolorbox{promptframe}{promptback}{\usebox{\promptboxsave}}%
   \par\addvspace{8pt}}

\newenvironment{squishitemize}
{\begin{itemize}[label=\textbullet,itemsep=0pt,parsep=0pt,topsep=0pt,partopsep=0pt,leftmargin=.15in,labelsep=0.05in]}
  {\end{itemize}}

\newcounter{paperpanel}
\renewcommand{\thepaperpanel}{\alph{paperpanel}}
\AtBeginEnvironment{figure}{\setcounter{paperpanel}{0}}
\newenvironment{subfigure}[2][t]{%
  \stepcounter{paperpanel}%
  \begin{minipage}[#1]{#2}\centering%
}{%
  \end{minipage}%
}
\newcommand{\subcaptiontext}[1]{\par\vspace{0.25em}{\small(\thepaperpanel)~#1\par}}

\usepackage{breqn}

\renewenvironment{equation*}{\begin{dmath*}}{\end{dmath*}}
\renewcommand{\[}{\begin{dmath*}}
\renewcommand{\]}{\end{dmath*}}
\usepackage[capitalize,noabbrev]{cleveref}

\title{Same Answer, Different Reasons: Trace-Level Divergence in Multilingual Large Language Models}

\author{%
  Rui Sun\textsuperscript{1,\dag} \quad
  Peizhao Mei\textsuperscript{1,*} \quad
  Xiwei Cheng\textsuperscript{1,*} \quad
  Kaizhuo Chen\textsuperscript{1,*} \\
  \textbf{Taoran Li\textsuperscript{2} \quad
  Zhiyuan Yu\textsuperscript{2} \quad
  Varun Chandrasekaran\textsuperscript{1}} \\[4pt]
  \textsuperscript{1}University of Illinois Urbana-Champaign \quad
  \textsuperscript{2}Texas A\&M University
}

\begin{document}

\maketitle

{\let\thefootnote\relax\footnotetext{\textsuperscript{\dag}Project lead. \textsuperscript{*}Equal contribution. Correspondence to: \texttt{ruis7@illinois.edu}, \texttt{varunc@illinois.edu}.}}

\begin{abstract}
\label{sec:abstract}

Multilingual large language models are increasingly deployed in settings where visible reasoning traces support verification, education, and safety auditing, yet standard evaluations typically measure only final-answer agreement or holistic response similarity. As a result, they establish outcome consistency without determining whether users across languages receive comparable supporting evidence. We introduce a trace-level divergence framework that compares visible reasoning traces as structured sequences, decomposing agreement into structural similarity, order consistency, and holistic similarity, with pair-specific calibration to enable comparison across languages. Across three reasoning benchmarks, seven languages, and six frontier LLMs, the natural baseline---embedding each whole response and comparing the two by cosine---systematically overstates agreement: $52.1\%$--$67.3\%$ of examples in the top quartile of holistic similarity are not in the top quartile of structural similarity. This dissociation is stable across sentence encoders and step segmentations, and blind bilingual annotation studies in two English-anchored language pairs show that human judgments align with structural similarity rather than with the holistic baseline. A same-language control shows that the level of divergence is language-specific: two English traces for paraphrases of the same question agree more closely than an English trace and its counterparts in other languages. Our results show that answer-level and holistic evaluations can substantially overestimate the consistency of multilingual explanatory artifacts, motivating trace-level evaluation for reliable multilingual deployment.
\end{abstract}

\section{Introduction}
\label{sec:intro}

Large language models (LLMs) are increasingly deployed in settings where they produce not only final answers but also visible reasoning traces intended to justify those answers. These traces are consumed by users, educators, auditors, and process-supervision systems to understand, verify, and improve model behavior~\citep{wei2022chain}. As multilingual LLMs become widely deployed, an implicit expectation follows: semantically equivalent questions posed in different languages should produce not only compatible answers but also broadly comparable supporting explanations. In this paper, we study the \emph{visible} chain of thought emitted by the model, not its hidden internal computation. A chain of thought need not faithfully expose the model's internal reasoning process~\citep{turpin2023language,lanham2023measuring,chen2025reasoning}, and we make no claim that it does. Rather, we study the explanatory artifact that downstream users, verifiers, and chain-of-thought monitors directly observe and act upon; our claims concern what that artifact contains in each language, not whether it reflects the computation behind it.

Existing multilingual evaluations largely measure \emph{outcome consistency}: whether translated prompts produce the same or correct final answer~\citep{shi2022multilingual,mmluprox2025}. More recent approaches compare complete responses using holistic semantic similarity, but still collapse an entire reasoning trace into a single representation. As a result, two multilingual responses may be judged equivalent even when they expose different intermediate claims, assumptions, problem decompositions, or execution order. Some of these differences may simply reflect natural linguistic variation; others may represent substantively different supporting evidence. Existing evaluations cannot distinguish between them. We therefore argue that multilingual systems should also be evaluated for \emph{evidentiary consistency}: whether users receive comparable supporting evidence for the same conclusion regardless of the language they ask in.

To address this problem, we introduce a trace-level divergence framework that compares multilingual reasoning traces as structured sequences rather than as single holistic responses. Our framework aligns step-structured chains of thought using multilingual sentence embeddings and decomposes cross-lingual trace consistency into three complementary metrics: structural similarity $S$, measuring preservation of reasoning content; order consistency $\tau$, measuring preservation of reasoning order; and holistic similarity $H$, a whole-response baseline. Because multilingual sentence embeddings exhibit language-pair-dependent similarity floors and ceilings, a pair-specific calibration rescales sentence similarities onto a readable scale within each language pair before aggregation.

We evaluate the proposed framework on three reasoning benchmarks, seven languages, and six frontier LLMs spanning four model families. We find that whole-answer similarity systematically overstates agreement between multilingual reasoning traces. Across all model--dataset evaluations, holistic similarity remains consistently high ($H=0.770$--$0.857$), while structural similarity is substantially lower ($S=0.502$--$0.629$), with order consistency lower still ($\tau=0.307$--$0.470$), even when both language outputs produce the same verified-correct answer. Rank-based analyses reach the same conclusion: within each model--dataset evaluation and language pair, $52.1\%$--$67.3\%$ of examples in the top quartile of holistic similarity are not also in the top quartile of structural similarity. Finally, blind bilingual annotation studies in two English-anchored language pairs show that human judgments of reasoning-trace equivalence align with structural similarity rather than holistic similarity, indicating that whole-response embeddings systematically overestimate agreement in the explanatory artifacts multilingual users actually observe.

Further analyses locate the divergence. In a same-language control, where each question is paraphrased in English and run through the identical pipeline, two English traces agree more closely than an English trace and its counterparts in other languages in every cell tested, although the holistic--structural dissociation itself also appears within English. Within a trace, alignment is weakest at the step stating the shared answer, and divergence grows on harder items. Throughout, we read low structural or order similarity as divergence, not failure.

Our contributions are threefold. First, we formulate multilingual evaluation as the problem of measuring the consistency of visible reasoning traces, complementing traditional outcome-based evaluation. Second, we introduce a calibrated trace-level framework that decomposes agreement into content preservation, execution order, and holistic similarity. Third, through a comprehensive empirical study, we show that answer-level and holistic evaluations systematically overstate the consistency of multilingual explanatory artifacts, motivating trace-level evaluation for reliable multilingual deployment.

\section{Problem Formulation}
\label{sec:problem_setup}

We study whether multilingual LLMs produce consistent \emph{visible reasoning traces} when given semantically equivalent prompts in different languages. Our goal is to evaluate two complementary notions of consistency: (i) whether multilingual prompts produce the same final answer, and (ii) whether they produce comparable supporting explanations. The former captures \emph{outcome consistency}; the latter captures \emph{trace consistency}. This section formalizes both notions and introduces the calibration required to compare multilingual reasoning traces across language pairs.

\noindent{\bf Setup.}
Let $f_\theta$ denote a fixed LLM and let $\mathcal{Q}=\{q_i\}_{i=1}^{N}$ be a set of reasoning problems specified at the level of semantic intent. For each $q_i$, we collect $K$ semantic-preserving translations $\mathcal{L}_i=\{\ell_{i,1},\ldots,\ell_{i,K}\}$ and obtain responses
\begin{align*}
y_{i,k} &= f_\theta(\ell_{i,k}).
\end{align*}
Each response is parsed into a final answer $\alpha_{i,k}$ and a step-structured visible reasoning trace
\begin{align*}
r_{i,k} &= (s_{i,k,1},\ldots,s_{i,k,t_{i,k}}),
\end{align*}
where $t_{i,k}$ denotes the number of reasoning steps. Each step is itself a sequence of sentences,
\begin{align*}
s_{i,k,t} &= (u_{i,k,t,1},\ldots,u_{i,k,t,n_{i,k,t}}),
\end{align*}
where $n_{i,k,t}$ is the number of sentences in step $t$. Let $\alpha_i^\star$ denote the reference answer.

\subsection{Outcome and Trace Consistency}
\label{sec:problem_formulation}

We formalize multilingual consistency at two complementary levels. Outcome consistency asks whether translated prompts produce the same final answer. Trace consistency asks whether they produce comparable visible reasoning traces. These notions are complementary: multilingual responses may reach the same answer while exposing different supporting claims, decompositions, or execution order.

\noindent{\bf 1. Outcome Consistency.}
Following prior multilingual evaluation work~\citep{qi2023cross,shi2022multilingual}, we define
\begin{align}
\Delta_{\text{acc}}(q_i)
&=
\max_k
\mathbf{1}[\alpha_{i,k}\neq\alpha_i^\star] \nonumber\\
&\quad
-
\min_k
\mathbf{1}[\alpha_{i,k}\neq\alpha_i^\star]
\;\in\;\{0,1\}.
\label{eq:delta_acc}
\end{align}
Here, $\Delta_{\text{acc}}=0$ iff all language variants are uniformly correct or uniformly incorrect, and $\Delta_{\text{acc}}=1$ otherwise. While this captures multilingual agreement at the level of outcomes, it provides no information about whether the accompanying visible reasoning traces remain consistent across languages.

\noindent{\bf 2. Trace Divergence.}
To quantify divergence between visible reasoning traces, we define
\begin{align}
D_{\text{trace}}(q_i)
&=
\frac{2}{K(K-1)}
\!\!\sum_{1\le k<k'\le K}\!\!
d(r_{i,k},r_{i,k'}).
\label{eq:d_reason}
\end{align}
Rather than a single scalar, we represent trace divergence using three complementary dimensions,
\begin{align}
d(r_a,r_b) &=
\Big(
1-S(r_a,r_b),\;
\tfrac{1-\tau(r_a,r_b)}{2},\nonumber\\
&\qquad\quad\;
1-H(r_a,r_b)
\Big).
\label{eq:trace_divergence_vector}
\end{align}
Thus, Eq.~\eqref{eq:d_reason} yields a three-dimensional expectation over all language pairs. The three components capture complementary aspects of multilingual trace consistency:

\begin{squishitemize}
\item \emph{Structural similarity} $S(\cdot,\cdot)\in[0,1]$: preservation of step-level reasoning content across traces (\S~\ref{sec:structural_score}).

\item \emph{Order consistency} $\tau(\cdot,\cdot)\in[-1,1]$: preservation of the ordering of corresponding reasoning steps (\S~\ref{sec:soft_tau}).

\item \emph{Whole-answer cosine} $H(\cdot,\cdot)\in[-1,1]$: a holistic embedding-based baseline that compares complete responses without localizing where divergence occurs (formally defined in Appendix~\ref{sec:raw_semantic}).
\end{squishitemize}

\noindent
Raw multilingual cosines carry language-pair-specific floors and ceilings, so the sentence similarities behind $S$ and $\tau$ are calibrated per language pair before aggregation (\S~\ref{sec:calibration}, Appendix~\ref{app:per_pair_calibration}).

\section{Method}
\label{sec:method}

Given two visible reasoning traces produced for semantically equivalent prompts in different languages, our framework measures whether they preserve the same step-level content and ordering. The pipeline first parses each response into steps (\S~\ref{sec:step_parsing}), calibrates multilingual sentence similarities for each language pair (\S~\ref{sec:calibration}), and constructs a step-pair similarity matrix through two-level greedy alignment (\S~\ref{sec:structural_score}). This matrix yields the structural similarity score $S$ and the order-consistency score $\tau$ (\S~\ref{sec:soft_tau}). We compute the holistic baseline $H$ separately over complete responses (Appendix~\ref{sec:raw_semantic}).

\subsection{Step Parsing}
\label{sec:step_parsing}

We segment each response using line-initial step headers. The parser first matches the prompted English pattern \texttt{Step}~$n$:, followed by its equivalents in the seven evaluated languages, a plain numbered-list pattern, and finally paragraph boundaries. The same ordered rule set is applied to every language, and parsing terminates at the final-answer line. Each non-empty step is further divided into sentences using a Unicode-aware regular expression. No LLM is used during parsing. Appendix~\ref{app:parser_details} provides the complete parsing hierarchy and validation details.

\subsection{Per-Pair Calibration}
\label{sec:calibration}

Raw multilingual cosine similarities $c_{a,b}(u,v)=\cos(\phi(u),\phi(v))$, for a multilingual sentence encoder $\phi$, exhibit language-pair-specific floors and ceilings. For each language pair $(a,b)$ we therefore estimate an unrelated-sentence floor $\widehat{\mathrm{floor}}_{a,b}$, the mean cosine between sentences drawn from traces of different questions, and a self-translation ceiling $\widehat{\mathrm{ceiling}}_{a,b}$, a high percentile of the cosine between trace sentences and their LLM translations into the other language (estimators and constants in Appendix~\ref{app:calibration_details}). Each raw sentence cosine is then rescaled as
\begin{align}
\widetilde{c}_{a,b}(u,v)
&=
\operatorname{clip}\!\left(
\frac{
c_{a,b}(u,v)
-
\widehat{\mathrm{floor}}_{a,b}
}{
\widehat{\mathrm{ceiling}}_{a,b}
-
\widehat{\mathrm{floor}}_{a,b}
},
0,1
\right).
\label{eq:calibrated_similarity}
\end{align}
Calibration aligns the empirical endpoints of each language pair, so that within a pair $0$ marks the unrelated-sentence floor and $1$ the self-translation ceiling; it does not assume that intermediate calibrated values are semantically identical across pairs. Within a pair it is a fixed monotone map, so it cannot reorder instances inside a pair, and the within-pair rank comparisons that carry our results are unaffected by it (\S~\ref{sec:exp_baseline_comparison}). Appendix~\ref{app:calibration_details} gives the estimators, the motivation for the constants, and translator-independence checks.

\subsection{Two-Level Greedy Alignment}
\label{sec:structural_score}

A trace is an ordered sequence of steps, and each step may contain multiple sentences. We therefore align traces hierarchically: an inner alignment compares sentences to score each step pair, and an outer alignment aggregates step-pair scores into a trace-level structural similarity.

\paragraph{Inner sentence alignment.}
For steps $s_{a,i}$ and $s_{b,j}$ containing $n_{a,i}$ and $n_{b,j}$ sentences, respectively, let
$\widetilde{\mathbf{C}}\in[0,1]^{n_{a,i}\times n_{b,j}}$
contain the calibrated sentence similarities from
Eq.~\eqref{eq:calibrated_similarity}. We define the precision-style score $p^{\mathrm{in}}_{i,j}=\frac{1}{n_{a,i}}\sum_{m=1}^{n_{a,i}}\max_n\widetilde{\mathbf{C}}_{m,n}$, the recall-style score $r^{\mathrm{in}}_{i,j}=\frac{1}{n_{b,j}}\sum_{n=1}^{n_{b,j}}\max_m\widetilde{\mathbf{C}}_{m,n}$, and the step-pair similarity $\sigma_{i,j}=F_1(p^{\mathrm{in}}_{i,j},r^{\mathrm{in}}_{i,j})$.
This BERTScore-style greedy alignment~\citep{zhang2020bertscore} is high only when the sentences in each step are covered by the other.

\paragraph{Outer step alignment.}
Collect the step-pair similarities into
$\mathbf{A}\in[0,1]^{t_a\times t_b}$, where
$\mathbf{A}_{i,j}=\sigma_{i,j}$. The structural similarity is
{\small
\begin{align}
S(r_a,r_b)
&=
F_1\!\left(
\frac{1}{t_a}\sum_i\max_j\mathbf{A}_{i,j},\;
\frac{1}{t_b}\sum_j\max_i\mathbf{A}_{i,j}
\right).
\label{eq:structural_similarity}
\end{align}
}
The bidirectional maxima permit many-to-one correspondences, allowing traces to express the same content at different step granularities. Because $S$ ignores the locations of matched steps, it measures content preservation independently of order.

\subsection{Soft Kendall-\texorpdfstring{$\tau$}{tau} Order Consistency}
\label{sec:soft_tau}

Structural similarity does not account for the positions of matched steps. We therefore derive an order-consistency score from the same step-pair matrix $\mathbf{A}$.

\paragraph{Soft assignment.}
For the directed comparison $a\rightarrow b$, we sharpen the step similarities and normalize each source-step row:
\begin{align}
\widetilde{\mathbf{A}}_{i,j}
&=
\mathbf{A}_{i,j}^{\,3},
\label{eq:sharpened_matrix}\\
P^{a\rightarrow b}_{i,j}
&=
\begin{cases}
\dfrac{\widetilde{\mathbf{A}}_{i,j}}
{\sum_{j'}\widetilde{\mathbf{A}}_{i,j'}},
&
\sum_{j'}\widetilde{\mathbf{A}}_{i,j'}>0,\\[3mm]
\dfrac{1}{t_b},
&
\text{otherwise}.
\end{cases}
\label{eq:soft_assignment}
\end{align}
Thus, $P^{a\rightarrow b}_{i,j}$ represents the soft correspondence between source step $i$ and target step $j$. Sharpening concentrates mass on stronger matches without imposing a hard one-to-one assignment. Appendix~\ref{app:sharpening_ablation} reports sensitivity to the sharpening choice.

\paragraph{Order mass.}
For each ordered source-step pair $i<k$, we accumulate probability mass according to whether their matched target positions are concordant, discordant, or tied:
\begin{align}
M^{a\rightarrow b}_{r}
&=
\frac{1}{\binom{t_a}{2}}
\sum_{i<k}
\sum_{(j,\ell)\in\Omega_r}
P^{a\rightarrow b}_{i,j}
P^{a\rightarrow b}_{k,\ell},
\nonumber\\
&\qquad
r\in\{\mathrm{con},\mathrm{dis},\mathrm{tie}\},
\label{eq:order_mass}
\end{align}
where $\Omega_{\mathrm{con}}=\{(j,\ell):j<\ell\}$, $\Omega_{\mathrm{dis}}=\{(j,\ell):j>\ell\}$, and $\Omega_{\mathrm{tie}}=\{(j,\ell):j=\ell\}$. The directed order score is
\begin{align}
\tau^{a\rightarrow b}
&=
\frac{
M^{a\rightarrow b}_{\mathrm{con}}
-
M^{a\rightarrow b}_{\mathrm{dis}}
}{
M^{a\rightarrow b}_{\mathrm{con}}
+
M^{a\rightarrow b}_{\mathrm{dis}}
+
M^{a\rightarrow b}_{\mathrm{tie}}
}.
\label{eq:directed_tau}
\end{align}

Because this score is directional, we symmetrize it:
\begin{align}
\tau(r_a,r_b)
&=
\frac{1}{2}
\left(
\tau^{a\rightarrow b}
+
\tau^{b\rightarrow a}
\right),
\label{eq:soft_tau}
\end{align}
using $\mathbf{A}$ for $a\rightarrow b$ and
$\mathbf{A}^{\top}$ for $b\rightarrow a$. The score is undefined when either trace has fewer than two steps, when either trace is empty, or when either directed denominator is zero. We interpret $\tau$ jointly with $S$: high $\tau$ with low $S$ indicates that the matched content is ordered consistently but structurally incomplete.

\section{Experimental Setup}
\label{sec:experiments}
\label{sec:exp_setup}

We evaluate on three multilingual reasoning benchmarks: \textsc{MMLU-ProX}~\citep{mmluprox2025}, \textsc{StrategyQA}~\citep{geva2021strategyqa}, and \textsc{SciBench}~\citep{ye2024scibench}. Each run uses seven languages (Arabic, English, Spanish, Hindi, Russian, Vietnamese, and Chinese), yielding 21 unordered language pairs per model--dataset evaluation. The benchmark suite spans distinct reasoning regimes: \textsc{MMLU-ProX} tests multiple-choice academic reasoning, where answer options and domain knowledge constrain the response; \textsc{StrategyQA} tests yes/no commonsense reasoning, where premise interpretation and implicit world knowledge are central; and \textsc{SciBench} tests quantitative STEM reasoning, where traces often combine formula selection, unit conversion, and numerical computation. This spread lets us distinguish dataset effects from model-specific behavior under fixed metric definitions.

We evaluate six models from four families: Claude Opus~4.6, Claude Haiku~4.5, DeepSeek-V3.2, GPT-OSS-120B, Qwen3-235B, and Qwen3-32B, sampling 500 problems per dataset--model run. Trace metrics are reported only after the answer- and trace-validity cascade; one-language-correct cases contribute only to $\Delta_{\mathrm{acc}}$ (Appendix~\ref{app:exp_setup_details}).

Metrics follow \S~\ref{sec:method}, using LaBSE~\citep{feng2022labse} for step scores, BGE-M3~\citep{chen2024bgem3} for $H$, and soft Kendall-$\tau$ sharpening power 3. Table~\ref{tab:aggregate_results} reports problem-level $95\%$ bootstrap confidence intervals (2{,}000 resamples), resampling each question's 21 language pairs jointly. Appendices~\ref{app:accuracy_summary}, \ref{app:calibration}, \ref{app:sharpening_ablation}, and \ref{app:parser_coverage} provide filtering details, audits, and ablations.

\section{Results}
\label{sec:results}

Our evaluation asks whether final-answer agreement implies trace agreement across languages (\S~\ref{sec:exp_main}), whether whole-answer similarity can substitute for step-level alignment (\S~\ref{sec:exp_baseline_comparison}), whether structural, order, and holistic similarity provide complementary signals (\S~\ref{sec:exp_decomposition}), whether the divergence is specific to crossing a language (\S~\ref{sec:within_language}), where in a trace it appears and what a low score signals (\S~\ref{sec:where}), and whether the resulting metrics are robust and agree with human judgment (\S~\ref{sec:validation}).

\begin{table*}[t]
\caption{\small Aggregate divergence analyses. $Q$ is the per-evaluation question count after the trace-validity cascade (Appendix~\ref{app:exp_setup_details}); $P$ is retained pair--question instances; ``pairs'' is the number of language pairs. $S$ (structural similarity), $\tau$ (soft Kendall-$\tau$), and $H$ (whole-answer BGE-M3 cosine) are macro-averaged over $Q$; brackets are half-widths of problem-level percentile bootstrap $95\%$ CIs (2{,}000 resamples). $H{-}S$ is a diagnostic gap, not a calibrated distance. Per-language outcome accuracies (Fig.~\ref{fig:app_accuracy_summary}) are computed on the full 500-problem inference set before this filter and are not bounded by $Q$.}

\label{tab:aggregate_results}
\centering
\scriptsize
\resizebox{\textwidth}{!}{%
\begin{tabular}{llrrrrrrrr}
\toprule
{\bf Dataset} & {\bf Model} & {\bf \# samples} & $Q$ & $P$ & {\bf \# pairs} & $S$ & $\tau$ & $H$ & $H{-}S$ \\
\midrule
\textsc{MMLU-ProX} & Claude Opus 4.6  & 500 & 431 & 8{,}470 & 21 & 0.569~[$\pm$0.007] & 0.414~[$\pm$0.014] & 0.800~[$\pm$0.004] & 0.231 \\
\textsc{MMLU-ProX} & Claude Haiku 4.5 & 500 & 321 & 5{,}459 & 21 & 0.514~[$\pm$0.009] & 0.393~[$\pm$0.016] & 0.787~[$\pm$0.004] & 0.273 \\
\textsc{MMLU-ProX} & DeepSeek-V3.2    & 500 & 394 & 7{,}179 & 21 & 0.537~[$\pm$0.007] & 0.356~[$\pm$0.014] & 0.794~[$\pm$0.004] & 0.257 \\
\textsc{MMLU-ProX} & GPT-OSS-120B     & 500 & 372 & 6{,}552 & 21 & 0.542~[$\pm$0.008] & 0.469~[$\pm$0.016] & 0.791~[$\pm$0.004] & 0.249 \\
\textsc{MMLU-ProX} & Qwen3-235B       & 500 & 370 & 6{,}656 & 21 & 0.540~[$\pm$0.006] & 0.348~[$\pm$0.015] & 0.793~[$\pm$0.004] & 0.253 \\
\textsc{MMLU-ProX} & Qwen3-32B        & 500 & 295 & 4{,}714 & 21 & 0.513~[$\pm$0.008] & 0.412~[$\pm$0.015] & 0.791~[$\pm$0.005] & 0.278 \\
\textsc{SciBench}  & Claude Opus 4.6  & 500 & 362 & 7{,}062 & 21 & 0.611~[$\pm$0.006] & 0.447~[$\pm$0.013] & 0.780~[$\pm$0.004] & 0.169 \\
\textsc{SciBench}  & Claude Haiku 4.5 & 500 & 296 & 4{,}739 & 21 & 0.572~[$\pm$0.009] & 0.470~[$\pm$0.016] & 0.770~[$\pm$0.004] & 0.197 \\
\textsc{SciBench}  & DeepSeek-V3.2    & 500 & 357 & 5{,}977 & 21 & 0.502~[$\pm$0.009] & 0.326~[$\pm$0.012] & 0.785~[$\pm$0.004] & 0.283 \\
\textsc{SciBench}  & GPT-OSS-120B     & 500 & 324 & 5{,}435 & 21 & 0.592~[$\pm$0.011] & 0.321~[$\pm$0.015] & 0.796~[$\pm$0.004] & 0.204 \\
\textsc{SciBench}  & Qwen3-235B       & 500 & 336 & 5{,}686 & 21 & 0.522~[$\pm$0.009] & 0.339~[$\pm$0.013] & 0.783~[$\pm$0.004] & 0.261 \\
\textsc{SciBench}  & Qwen3-32B        & 500 & 268 & 3{,}874 & 21 & 0.629~[$\pm$0.012] & 0.332~[$\pm$0.015] & 0.781~[$\pm$0.005] & 0.152 \\
\textsc{StrategyQA} & Claude Opus 4.6 & 500 & 451 & 9{,}039 & 21 & 0.598~[$\pm$0.007] & 0.418~[$\pm$0.012] & 0.818~[$\pm$0.004] & 0.220 \\
\textsc{StrategyQA} & Claude Haiku 4.5 & 500 & 379 & 6{,}857 & 21 & 0.530~[$\pm$0.008] & 0.400~[$\pm$0.013] & 0.857~[$\pm$0.004] & 0.327 \\
\textsc{StrategyQA} & DeepSeek-V3.2   & 500 & 414 & 7{,}813 & 21 & 0.560~[$\pm$0.006] & 0.324~[$\pm$0.010] & 0.795~[$\pm$0.004] & 0.235 \\
\textsc{StrategyQA} & GPT-OSS-120B    & 500 & 357 & 6{,}751 & 21 & 0.541~[$\pm$0.007] & 0.380~[$\pm$0.013] & 0.816~[$\pm$0.004] & 0.275 \\
\textsc{StrategyQA} & Qwen3-235B      & 500 & 402 & 7{,}591 & 21 & 0.560~[$\pm$0.007] & 0.355~[$\pm$0.012] & 0.801~[$\pm$0.004] & 0.240 \\
\textsc{StrategyQA} & Qwen3-32B       & 500 & 301 & 5{,}177 & 21 & 0.515~[$\pm$0.008] & 0.307~[$\pm$0.012] & 0.793~[$\pm$0.005] & 0.278 \\
\bottomrule
\end{tabular}%
}
\end{table*}

Appendix~\ref{app:additional_results} provides the supporting result details: pair-level tables (Appendix~\ref{app:pair_tables}), cross-dataset and cross-model patterns (Appendix~\ref{app:replication}), qualitative verified traces (Appendix~\ref{app:qualitative}), and rejected-sample audits outside the primary verified-correct set (Appendix~\ref{app:rejected_case_studies}).

\subsection{Correct Answers Still Hide Trace Divergence}
\label{sec:exp_main}

We first test whether agreement on the final answer implies agreement in the emitted reasoning trace by restricting to language pairs where both answers are verified correct. If multilingual reasoning were invariant, such pairs would exhibit high structural similarity $S$, high order consistency $\tau$, and only a small gap between the holistic score $H$ and the step-level score $S$.

However, across the evaluations in Table~\ref{tab:aggregate_results}, $H$ remains consistently high ($0.770$--$0.857$), while $S$ is substantially lower ($0.502$--$0.629$) and $\tau$ lower still ($0.307$--$0.470$). The same pattern holds on \textsc{MMLU-ProX}: across six 500-sample evaluations (21 language pairs), $S{=}0.513$--$0.569$, $\tau{=}0.348$--$0.469$, and $H{=}0.787$--$0.800$.

Taken together, these results contradict the invariance hypothesis. Even when answers agree, reasoning traces diverge: responses remain globally similar as full strings (high $H$) while differing in step content ($S$) and, more markedly, in step ordering ($\tau$), particularly on \textsc{MMLU-ProX}.

\subsection{Holistic Similarity Overstates Consistency}
\label{sec:exp_baseline_comparison}

\begin{figure}[t]
  \centering
  \includegraphics[width=\columnwidth]{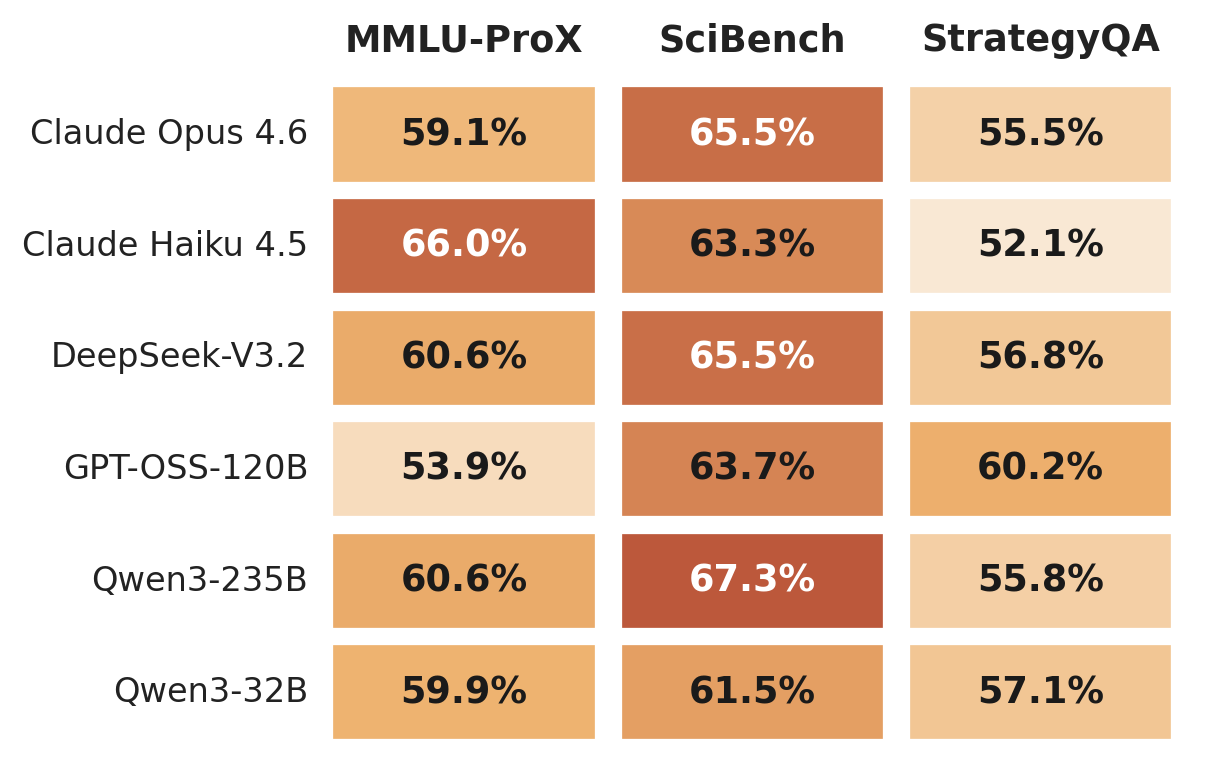}
\caption{\small Rank-based holistic-to-structural mismatch across all 18 model--dataset evaluations. Each cell reports the share of pair--question instances in the top quartile of whole-answer cosine $H$ that are not also in the top quartile of structural score $S$, with quartiles computed within the same evaluation and language pair. Appendix~\ref{app:rank_h_s} reports the per-dataset breakdown, including the stricter share of top-$H$ instances falling in the bottom structural quartile.}
  \label{fig:H_vs_S}
\end{figure}

We test whether a whole-answer embedding baseline can substitute for step-level alignment by comparing $H$ against $S$. Because $H$ uses BGE-M3 (8K context) while $S$ uses LaBSE for trace-length reasons (Appendix~\ref{sec:raw_semantic}), we treat $H{-}S$ as a diagnostic contrast and rely on a scale-free rank comparison. If a holistic embedding captures cross-lingual reasoning structure, high $H$ should coincide with high $S$ within the same evaluation and language pair.

In practice, Table~\ref{tab:aggregate_results} shows aggregate $H{-}S$ gaps of $0.152$ to $0.327$ across the 21 language-pair evaluations; we do not interpret these absolute magnitudes as calibrated distances given the encoder asymmetry. Figure~\ref{fig:H_vs_S} provides the scale-free rank comparison (computed within each (evaluation, pair, encoder) cell, hence invariant to per-pair monotone rescaling): within each evaluation and language pair, $52.1\%$--$67.3\%$ of top-quartile $H$ examples fall outside the top quartile of $S$, and a nontrivial fraction land in the bottom structural quartile. Appendix~\ref{app:rank_h_s} reports the full rank table and representative raw point-cloud panels.

This comparison is unaffected by the per-pair calibration: quartile thresholds are computed within each language pair, where calibration is a single monotone map, and recomputing the statistic on the uncalibrated scale reproduces it to four decimal places on all three Claude Opus~4.6 evaluations (Appendix~\ref{app:calibration_rank_invariance}). The reported mismatch is therefore a property of the traces rather than of the calibration constants.

These results do not support replacing step-level alignment with a holistic baseline. While $H$ serves as a coarse semantic signal, it is structurally under-informative, failing to detect missing, merged, or reordered reasoning steps: a consistency check built on whole-response cosine returns a high score on precisely the pairs where step content and order have moved.

\subsection{Structural Coverage and Order Are Non-Redundant}
\label{sec:exp_decomposition}

We finally test whether the decomposition into $S$, $\tau$, and $H$ is necessary by examining whether the three metrics capture distinct variation. If a single trace-level score were sufficient, then $S$, $\tau$, and $H$ would be strongly correlated after controlling for the remaining metric.

\begin{table}[t]
\centering
\caption{Pairwise Pearson and partial correlations among the three metrics at the pair--question level for Claude Opus~4.6. For each metric pair, the first value is the Pearson correlation and the second is the partial correlation after controlling for the remaining metric.}
\label{tab:correlations}
\small
\setlength{\tabcolsep}{3pt}
\begin{tabular}{@{}lrrrr@{}}
\toprule
\textbf{Dataset} & $n$ & $S$--$\tau$ & $S$--$H$ & $\tau$--$H$ \\
\midrule
\textsc{MMLU-ProX}  & 8{,}470 & .29\,/\,.29 & .29\,/\,.29 & .05\,/\,$-$.04 \\
\textsc{StrategyQA} & 9{,}039 & .30\,/\,.31 & .33\,/\,.33 & .04\,/\,$-$.07 \\
\textsc{SciBench}   & 7{,}062 & .17\,/\,.17 & .29\,/\,.29 & .02\,/\,$-$.03 \\
\bottomrule
\end{tabular}
\end{table}

In practice, Table~\ref{tab:correlations} shows that $\tau$ and $H$ are nearly uncorrelated, while $S$ has only moderate independent association with both. The same separation holds at the aggregate level: evaluations with similar $S$ can exhibit markedly different order consistency, and large $H{-}S$ gaps or top-$H$ mismatches do not align with $\tau$. Appendix~\ref{app:replication} provides a complementary cross-dataset check (Claude Opus~4.6): language-pair rankings of $H$ replicate strongly across datasets ($\rho=0.78$--$0.98$), $\tau$ rankings replicate moderately ($\rho=0.52$--$0.76$), while $S$ is more task-sensitive ($\rho=0.11$--$0.55$).

These results reject the single-score hypothesis. The three metrics capture complementary aspects of reasoning: $S$ reflects step coverage, $\tau$ captures execution order, and $H$ measures only holistic semantic similarity. The cross-dataset pattern further highlights this distinction: holistic similarity is relatively stable across tasks, order behaviour is partly language-pair dependent, and structural coverage is most sensitive to how reasoning is elicited. Collapsing these axes would obscure both task- and model-specific forms of trace divergence.

\subsection{Divergence Is Larger Across Languages Than Within One}
\label{sec:within_language}

Every comparison so far crosses a language boundary, so on its own it cannot separate divergence caused by changing language from ordinary variation in generated reasoning. As a same-language control, we rewrite each question as a meaning-preserving English paraphrase and run it through the identical pipeline as a condition $\text{en}_{\text{para}}$ that receives the same appended instruction as a translated condition, for GPT-OSS-120B and DeepSeek-V3.2 on all three benchmarks, about 100 questions each (Appendix~\ref{app:within_language}). Pairing within question, we compare $S(\text{en},\text{en}_{\text{para}})$ with the same question's mean $S$ over its English-anchored cross-lingual pairs. Two English traces agree more closely in all six dataset--model cells and in 272 of 337 questions (Table~\ref{tab:within_language}; pooled $\Delta S=+0.090$, $95\%$ CI $[+0.077,+0.103]$, $p<10^{-33}$), and order consistency moves the same way in every cell. For this cross-pair comparison of levels, calibration favours the same-language pair by at most $0.024$; subtracting that bias cell by cell leaves $\Delta S$ between $+0.046$ and $+0.124$, above thresholds fixed before the run.

\begin{table}[t]
\centering
\caption{Same-language control (GPT-OSS: GPT-OSS-120B; DeepSeek: DeepSeek-V3.2). $\Delta S$ is $S(\text{en},\text{en}_{\text{para}})$ minus the same question's mean $S$ over its six English-anchored cross-lingual pairs; brackets are half-widths of $95\%$ $t$-intervals. ``$\#{>}0$'' counts questions with $\Delta S>0$.}
\label{tab:within_language}
\small
\setlength{\tabcolsep}{2.5pt}
\begin{tabular}{@{}llrrr@{}}
\toprule
\textbf{Dataset} & \textbf{Model} & $n$ & $\Delta S$ & $\#{>}0$ \\
\midrule
\textsc{MMLU-ProX}  & GPT-OSS  & 59 & $+0.120$~[$\pm$0.029] & 49 \\
\textsc{MMLU-ProX}  & DeepSeek & 68 & $+0.071$~[$\pm$0.024] & 56 \\
\textsc{SciBench}   & GPT-OSS  & 41 & $+0.136$~[$\pm$0.048] & 36 \\
\textsc{SciBench}   & DeepSeek & 47 & $+0.083$~[$\pm$0.037] & 39 \\
\textsc{StrategyQA} & GPT-OSS  & 57 & $+0.041$~[$\pm$0.029] & 36 \\
\textsc{StrategyQA} & DeepSeek & 65 & $+0.101$~[$\pm$0.028] & 56 \\
\midrule
\multicolumn{2}{@{}l}{Pooled} & 337 & $+0.090$~[$\pm$0.013] & 272 \\
\bottomrule
\end{tabular}
\end{table}

The rank dissociation of \S~\ref{sec:exp_baseline_comparison}, by contrast, persists within English: $41.2\%$--$60.0\%$ of top-quartile-$H$ instances are not top-quartile-$S$ between two English traces, against $50.5\%$--$64.6\%$ across languages in the same run. Holistic similarity thus overstates step-level agreement whether or not the traces share a language, a property of the metrics; what is specific to crossing a language is the level of divergence.

\subsection{Where Divergence Appears and What It Signals}
\label{sec:where}

\noindent{\bf Position.} Scoring each English step by its best match in the other trace (English-anchored pairs, all 18 evaluations; Appendix~\ref{app:positional}), alignment is weakest at the end: the final step is less similar than the middle steps by $0.042$ ($95\%$ CI $[0.038,0.046]$; lower in 16 of 18 evaluations), and the gap widens to $0.058$ when both traces have the same number of steps, so it is not an unmatched extra step. Every pair here reaches the same correct answer; the step stating it is the least aligned. Where traces first part depends on the task, in an order stable across thresholds: \textsc{SciBench} and \textsc{MMLU-ProX} traces often part at the setup step, \textsc{StrategyQA} traces mid-trace. The profile has the same shape in all six target languages, which differ only in level.

\noindent{\bf Difficulty and correctness.} Divergence is larger on harder items: indexing difficulty by the number $k$ of the seven languages that answer correctly, mean $S$ over correct EN--ZH and EN--ES pairs rises from $0.526$ at $k=2$ to $0.579$ at $k=7$ ($\rho=0.10$, $p<10^{-25}$; Appendix~\ref{app:correctness}). Scoring the pairs the main analyses exclude, among answer-agreeing pairs the $225$ that share a wrong answer have lower $S$ than the $11{,}176$ that share the correct one (Cliff's $\delta=+0.293$, $95\%$ CI $[+0.217,+0.366]$), while $H$ does not separate them ($\delta=-0.027$, $[-0.114,+0.055]$). Against correct pairs of equal $k$, however, the $S$ contrast falls to $+0.073$ ($[-0.061,+0.209]$): low $S$ marks hard items, on which shared errors concentrate, rather than errors as such.

\noindent{\bf Reading a low score.} Low $S$ or $\tau$ is divergence, not failure: two correct traces can legitimately differ, and the step matrix shows where. It still helps allocate review. Answer agreement cannot flag answer-agreeing pairs that share a wrong answer; ranking such pairs by $S$ within each evaluation and language pair, the least similar quarter contains $44.0\%$ of these errors, against $28.9\%$ for $H$ and $25\%$ at random (paired difference $+15.1$ points, $[+6.2,+23.9]$). At a $2\%$ base rate this prioritizes a fixed review budget; it is not an error detector.

\subsection{Robustness and Human Corroboration}
\label{sec:validation}

The analyses above read low $S$ as genuine trace-level divergence. We check that this reading does not depend on the choice of sentence encoder or on step segmentation, and that it is corroborated by human judgment. Protocols, sampling, and full statistics are in Appendices~\ref{app:granularity}, \ref{app:human_validation}, and~\ref{app:encoder_sensitivity}.

\noindent{\bf Encoder robustness.} Recomputing $S$ and $\tau$ under two alternative sentence encoders---BGE-M3 (568M parameters), the encoder $H$ already uses, so that $H$ and $S$ share an encoder, and a distilled MiniLM (118M)---leaves the rank-based finding intact: across three encoders and three datasets, every evaluation falls between $50.1\%$ and $65.5\%$ of top-quartile-$H$ examples not being top-quartile-$S$, closely tracking the $52.1\%$--$67.3\%$ range reported above. Absolute calibrated levels are less stable, with the $H{-}S$ gap widening under one encoder and narrowing under the other. We therefore interpret $S$ and $\tau$ by rank within a fixed encoder rather than by absolute level, and report them that way throughout. Because the framework reads only the emitted traces and a sentence encoder, its requirements are modest: it needs no access to weights or activations---two of the six evaluated models are closed-weight---and the 118M encoder runs on CPU.

\noindent{\bf Step segmentation.} Restricting each evaluation to pairs whose traces have equal step counts lowers the statistic by $1.0$--$2.9$ points on the three Claude Opus~4.6 evaluations and leaves all 18 evaluations between $49.5\%$ and $71.0\%$ (Appendix~\ref{app:granularity}); the parser recovers steps from $99.8\%$--$100\%$ of traces in every language.

\begin{table}[t]
\centering
\caption{Spearman correlations between human-judged reasoning equivalence and each metric ($N=60$ per study). Partial correlations control for the other metric. Brackets are bootstrap 95\% CIs; they were not computed for the EN--ZH study. Ordinal Krippendorff's $\alpha$ is $0.41$ (EN--ZH) and $0.58$ (EN--ES).}
\label{tab:hv_correlations}
\small
\setlength{\tabcolsep}{3pt}
\begin{tabular}{@{}lrrcr@{}}
\toprule
\textbf{Quantity} & \multicolumn{2}{c}{\textbf{EN--ZH}} & \multicolumn{2}{c}{\textbf{EN--ES}} \\
\cmidrule(lr){2-3}\cmidrule(lr){4-5}
 & $\rho$ & $p$ & $\rho$ [95\% CI] & $p$ \\
\midrule
$\rho(\text{equiv}, S)$      & .371     & .004     & .44 [.23, .62]      & $<$.001 \\
$\rho(\text{equiv}, \tau)$   & .421     & $<$.001  & .38 [.16, .57]      & .001 \\
$\rho(\text{equiv}, H)$      & $-$.066  & n.s.     & .03 [$-$.21, .26]   & .80 \\
\midrule
$\rho(\text{equiv}, S \mid H)$ & .391    & .002     & .45 [.24, .63]      & $<$.001 \\
$\rho(\text{equiv}, H \mid S)$ & $-$.147 & n.s.     & $-$.08 [$-$.31, .16] & .51 \\
\bottomrule
\end{tabular}
\end{table}

\noindent{\bf Human corroboration.} To check that the divergence $S$ measures is one a reader would recognize, bilingual annotators rated verified-correct Claude Opus~4.6 trace pairs on a three-point reasoning-equivalence scale, blind to all metric values, in two language pairs (English--Chinese and English--Spanish, $N=60$ each, sampled so that $H$ and $S$ both vary over their full ranges). Table~\ref{tab:hv_correlations} reports the correlations: equivalence judgments track $S$ and $\tau$ in both pairs and are unrelated to $H$, and the separation survives controlling each metric for the other. The quantity we test directly is the difference between $S$ and $H$ as predictors of the same judgments. Williams' test for dependent correlations gives $\Delta\rho = 0.44$, $t(57) = 2.80$, $p = 0.007$ for English--Chinese and $\Delta\rho = 0.41$, $t(57) = 2.81$, $p = 0.007$ for English--Spanish, and a Fisher-$z$ test finds no difference between the two language pairs ($p = 0.65$) even though Spanish is typologically the closer of the two to English. Asked to name the type of difference, annotators point to missing reasoning content, granularity, and step order rather than wording---the profile a content-sensitive measure should produce, and the opposite of what a verbalization detector would. A blind LLM judge given both traces and the same rubric tracks the human consensus less closely than $S$ does, and never applies the \emph{Substantially different} label.

These studies are small, and we present them as corroboration of the automated results rather than as independent evidence that $S$ is superior to alternatives. Their role is to check that the metric responds to something human readers also see. They cover one model's traces and two English-anchored pairs, and we do not claim the agreement extends beyond them.

\section{Related Work}
\label{sec:related_work}

\noindent{\bf Multilingual reasoning evaluation.}
Chain-of-thought prompting elicits step-by-step reasoning~\citep{wei2022chain}, but multilingual evaluations largely stop at outputs. \citet{shi2022multilingual} measure multilingual CoT accuracy and \citet{barua2025long} extend this to long-form reasoning, while benchmarks such as MMLU-ProX~\citep{mmluprox2025} and metrics like RankC~\citep{qi2023cross} assess consistency at the level of final answers. \citet{qi2025xreasoning} instead control the language a model thinks in and measure the resulting accuracy cost, which is complementary along a distinct axis: they vary the trace language and observe correctness, whereas we hold correctness fixed and measure how far the traces diverge. None of these localizes divergence within a trace.

\noindent{\bf Trace-level reasoning evaluation.}
ROSCOE~\citep{golovneva2023roscoe} scores reasoning chains at step level in monolingual settings, and we extend that perspective to the cross-lingual case, where similarity must be calibrated per language pair. ROSCOE is not an available baseline here: it scores a \emph{single} chain against a reference to judge that chain's quality, whereas we compare two equally valid traces whose quality is held fixed by the validity cascade, and its monolingual encoders would have to be replaced by a multilingual one---reducing the comparison to the uncalibrated step cosines our calibration exists to correct (Appendix~\ref{app:calibration}). Concurrent work~\citep{multilingual_cot_eval2025} studies cross-lingual transfer of rationales, but without step- or order-level alignment.

\noindent{\bf Faithfulness of the visible trace.}
A separate line of work asks whether an emitted chain of thought reflects the computation that produced the answer: models can rationalize a predetermined answer~\citep{turpin2023language,lanham2023measuring,chen2025reasoning}, and consequential computation can leave no interpretable trace in the output tokens at all~\citep{baherwani2026invisible}. Our measurement is orthogonal to this question: we compare two \emph{emitted} traces and make no claim that either exhausts internal computation. The visible trace is nonetheless what verification, education, and chain-of-thought monitoring pipelines consume~\citep{korbak2025monitorability}, so our claim concerns a monitor's input, not its faithfulness: whatever a monitor extracts from a trace, it extracts from a different artifact in each language. This needs only the emitted text, so the framework also applies to closed-weight models.

\noindent{\bf Multilingual embeddings and alignment.}
We use LaBSE~\citep{feng2022labse} for sentence embeddings and BGE-M3~\citep{chen2024bgem3} for holistic similarity. Our alignment extends embedding-based matching~\citep{zhang2020bertscore} to sentence- and step-level structure, and uses a soft Kendall-$\tau$ inspired by continuous relaxations of permutations~\citep{mena2018sinkhorn,grover2019neuralsort}. Appendix~\ref{sec:relation} gives a detailed comparison with prior step-level metrics.

\section{Conclusion}
\label{sec:conclusion}

We show that agreement on final answers does not imply agreement in the reasoning traces a model emits across languages: holistic similarity can remain high even when step-level structure and order diverge. The consequence is at the level of the artifact a user reads: for the same correct answer, the supporting evidence a reader is given can depend on the language they asked in---steps another reader never sees, or the same steps in a different order---more than on how the question is worded within one language. It appears in every one of the $21$ language pairs we evaluate, including the typologically closest. By decomposing trace consistency into structural similarity $S$, order consistency $\tau$, and holistic similarity $H$, our framework reveals complementary axes that single-score evaluations obscure and locates where traces part. The rank-based finding is stable across sentence encoders and step segmentations, and small blind annotation studies on one model's traces in two English-anchored language pairs are consistent with it. Because the dissociation also appears between two English traces, the decomposition extends to process-level consistency beyond cross-lingual settings, including across paraphrases, prompts, and decoding strategies.

\clearpage
\section*{Limitations}
\label{sec:limitations}

\noindent{\bf Visible traces, not internal computation.} A chain of thought need not be a faithful record of the computation that produced the answer~\citep{turpin2023language,lanham2023measuring,chen2025reasoning}, and we link the divergence we measure to no internal quantity. The visible trace is what verification, education, and auditing pipelines consume, and that is the scope of our claims. Activation-level analysis would address faithfulness directly, but it would exclude closed-weight models, including two of the six we evaluate.

\noindent{\bf Ranks, not levels.} What survives an encoder change is rank structure rather than absolute scale (Appendix~\ref{app:encoder_sensitivity}), and per-pair calibration aligns only the \emph{endpoints} of the similarity scale. We therefore compare $S$ and $\tau$ by rank within a fixed encoder and read absolute cross-pair values as descriptive only. The one cross-pair comparison of levels we draw, the same-language control of \S~\ref{sec:within_language}, is paired within question and its calibration exposure is measured, but it remains a level comparison. Ceiling estimates additionally rely on LLM translation, whose co-dependence with the evaluated model we bound in Appendix~\ref{app:translator-independence}.

\noindent{\bf Scope of the evidence.} The human validation covers two language pairs, both including English, with $60$ verified-correct trace pairs each, all from one model; agreement is moderate, which attenuates rather than inflates the reported correlations. We do not establish that the metrics track human judgment for other models or for pairs without English. The same-language control covers two of the six models, about $100$ questions per benchmark, and one paraphrase per question, and it does not resample an identical prompt at non-zero temperature, which would isolate decoding variation; its effect is unanimous in direction but varies about threefold in size across cells. Step boundaries are verified automatically and by a manual spot check too small to bound the residual error rate (Appendix~\ref{app:parser_validation}).

\noindent{\bf Divergence and outcomes.} The association between low $S$ and shared wrong answers rests on $225$ pairs from two language pairs and is largely carried by item difficulty (\S~\ref{sec:where}). We draw no causal conclusion from it, and we do not show that divergence impairs verification, comprehension, or safety; that would require a study with human readers.

\section*{Ethical Considerations}
\label{sec:ethics}

\noindent{\bf Data and annotation.} Questions come from public reasoning benchmarks and the traces are model-generated, so no personal or sensitive data is involved. Annotators were bilingual speakers informed of the purpose of the study, and no personal data was collected from them; Appendix~\ref{app:hv_annotators} reports recruitment and consent in full.

\noindent{\bf Artifact release.} We will release the evaluation code under a permissive open-source license. It builds on publicly released models and benchmarks---LaBSE, BGE-M3, and \texttt{paraphrase-multilingual-MiniLM-L12-v2}, together with \textsc{MMLU-ProX}, \textsc{StrategyQA}, and \textsc{SciBench}---and our use and redistribution are consistent with their respective licenses. We redistribute no benchmark data and no annotation labels.

\noindent{\bf Intended use.} The framework diagnoses whether a system reasons the same way in different languages; it does not certify reasoning quality. Scores are relative within a fixed encoder and language pair and should not be used to rank languages, to gate access to a service, or as a training target---traces can be made to agree without any improvement in the underlying reasoning. Because all pairs in the main analyses are verified-correct, a low score there marks a different route to the same answer, not a wrong one; its association with shared errors supports prioritizing human review, not detecting errors.

\noindent{\bf Fairness.} Equal accuracy can mask unequal traces, so users who read a model's explanation may receive systematically different support depending on the language they use.

\bibliography{references}

\clearpage
\appendix
\raggedbottom

\newcommand{\appendixsectionbreak}{\newpage}

\appendixsectionbreak
\section{The Necessity of Per-Language-Pair Calibration}
\label{app:per_pair_calibration}

The trace-divergence measure $D_{\text{trace}}$ relies on cross-lingual sentence similarities that must be comparable across language pairs. In practice, raw cosine similarities violate this requirement because they conflate semantic relatedness with language-pair-specific properties of the embedding space. These differences manifest as pair-dependent similarity floors and ceilings (reported in Appendix~\ref{app:calibration}), making raw cosine scores unsuitable for direct comparison. We therefore normalize sentence similarities using pair-specific calibration before constructing the trace-level metrics.

Formally, the structural and order metrics are built from the sentence-level cosine

\begin{gather*}
c_{a,b}(u,u')
=
\cos\!\big(\phi(u),\phi(u')\big),\\
u\in\mathcal{U}_a,\;
u'\in\mathcal{U}_b,
\end{gather*}

where $\phi$ is a multilingual sentence encoder and $\mathcal{U}_a$ denotes the space of sentences appearing in language-$a$ reasoning traces.

Raw cosine similarity is not directly comparable across language pairs because both the expected similarity between unrelated sentences (the \emph{floor}) and the attainable similarity between semantically equivalent sentences (the \emph{ceiling}) vary across languages. These shifts arise from anisotropy and uneven cross-lingual alignment in multilingual embedding spaces~\citep{feng2022labse,chen2024bgem3}. We define

\begin{align*}
\mathrm{floor}_{a,b}
&=
\mathbb{E}
\!\left[
c_{a,b}(u,u')
\mid
u\perp u'
\right],\\
\mathrm{ceiling}_{a,b}
&=
\mathbb{E}
\!\left[
c_{a,b}(u,T_{a\rightarrow b}(u))
\right],
\end{align*}

where $u\perp u'$ denotes semantic independence and
$T_{a\rightarrow b}:\mathcal{U}_a\rightarrow\mathcal{U}_b$
is a semantics-preserving translator.

The calibrated similarity is then

\begin{gather*}
\widetilde{c}_{a,b}(u,u')
=
\frac{
c_{a,b}(u,u')
-
\mathrm{floor}_{a,b}
}{
\mathrm{ceiling}_{a,b}
-
\mathrm{floor}_{a,b}
}
\in[0,1],\\
u\in\mathcal{U}_a,\;
u'\in\mathcal{U}_b.
\end{gather*}

After calibration, sentence similarities depend primarily on semantic relatedness rather than language-pair-specific embedding geometry, enabling meaningful comparison and aggregation across multilingual reasoning traces.

\appendixsectionbreak
\section{Supplementary Method Details}
\label{app:method}
\label{app:method_details}

\subsection{Step-Parsing Details}
\label{app:parser_details}

Each response is segmented using the following ordered fallback hierarchy:

\begin{enumerate}
    \item the prompted English line-initial pattern \texttt{Step}~$n$:;
    \item line-initial equivalents of the step marker in the seven evaluated languages;
    \item a plain line-initial numbered-list pattern; and
    \item paragraph boundaries.
\end{enumerate}

Parsing terminates when the final-answer marker is encountered. Empty blocks are discarded, and each remaining step is divided into sentences using a Unicode-aware regular expression. The same ordered rule set is applied in every language, and the parser records which tier was used for each response. No LLM or language-specific semantic parser is invoked.

We distinguish two notions of parser quality. \emph{Coverage} asks whether a response yields a non-empty sequence of steps. \emph{Boundary fidelity} asks whether the extracted step count agrees with the explicit step markers in the raw output. The corresponding diagnostics by language, model, and dataset, together with a manual spot check of the recovered boundaries, are reported in Appendix~\ref{app:parser_validation}.

\subsubsection{Parser Validation}
\label{app:parser_validation}

The automatic marker-count comparison evaluates whether the implementation faithfully recovers the boundaries explicitly emitted by the model. It does not establish that those model-provided boundaries constitute the uniquely correct semantic segmentation of the trace. We therefore also read the parsed output manually, inspecting roughly $20$ randomly drawn traces per model--dataset cell against the corresponding raw responses. In every trace inspected, the recovered boundaries matched the segmentation a reader would impose, and we found no case of a step being merged or split against the evident structure of the trace. This audit is small: manual boundary inspection is costly, and we did not extend it to a sample large enough to bound the residual error rate. We report it as a spot check supporting the automatic diagnostics rather than as an independent measurement of boundary fidelity.

\subsection{Calibration Details and Robustness}
\label{app:calibration_details}

\subsubsection{Floor and Ceiling Estimators}

The calibration interval of \S\ref{sec:calibration} is estimated independently for each language pair $(a,b)$.

Let $\mathcal{U}_a$ denote the sentences appearing in language-$a$ traces, and let $q(u)$ identify the source question from which sentence $u$ was extracted. We estimate the unrelated-sentence floor using pairs sampled from distinct questions:
\begin{align}
\widehat{\mathrm{floor}}_{a,b}
&=
\frac{1}{|\mathcal{P}_{a,b}|}
\!\!\sum_{(u,u')\in\mathcal{P}_{a,b}}\!\!
c_{a,b}(u,u'),
\nonumber\\
&\qquad\qquad\qquad
q(u)\neq q(u'),
\label{eq:floor_estimator}
\end{align}
where
\begin{align*}
c_{a,b}(u,u')
&=
\cos\!\big(\phi(u),\phi(u')\big)
\end{align*}
and $\phi$ is a multilingual sentence encoder.

To estimate the attainable similarity between semantically equivalent sentences, we sample
$\mathcal{V}_a\subseteq\mathcal{U}_a$ and
$\mathcal{V}_b\subseteq\mathcal{U}_b$ from traces whose final answers are verified as correct. Let
$T_{a\rightarrow b}$ and $T_{b\rightarrow a}$ denote an LLM translator in the two directions. We pool the resulting self-translation similarities:
\begin{multline}
\mathcal{C}^{\mathrm{trans}}_{a,b}
=
\big\{
c_{a,b}(u,T_{a\rightarrow b}(u)):
u\in\mathcal{V}_a
\big\}\\
\cup
\big\{
c_{a,b}(T_{b\rightarrow a}(v),v):
v\in\mathcal{V}_b
\big\}.
\label{eq:translation_pool}
\end{multline}
We define the empirical ceiling as
\begin{align}
\widehat{\mathrm{ceiling}}_{a,b}
&=
\operatorname{clip}\big(
\operatorname{perc}_{90}
(\mathcal{C}^{\mathrm{trans}}_{a,b}),
\nonumber\\
&\qquad\quad
\widehat{\mathrm{floor}}_{a,b}
+\epsilon_{\mathrm{cal}},\;
c_{\max}
\big),
\label{eq:ceiling_estimator}
\end{align}
where $\epsilon_{\mathrm{cal}}=0.10$ and $c_{\max}=0.98$. Each raw sentence cosine is then rescaled by Eq.~\eqref{eq:calibrated_similarity}.

\subsubsection{Motivation for the Calibration Constants}

The empirical ceiling in Eq.~\eqref{eq:ceiling_estimator} uses the $90$th percentile of bidirectional self-translation similarities rather than the maximum. The percentile reduces sensitivity to occasional near-duplicate or unusually literal translations that would otherwise compress the calibrated dynamic range.

The minimum margin
\[
\epsilon_{\mathrm{cal}}=0.10
\]
prevents a nearly degenerate calibration interval when the estimated floor and ceiling are too close. The ceiling cap
\[
c_{\max}=0.98
\]
prevents isolated near-identity translations from setting an unrealistically high endpoint.

The floor and ceiling are estimated independently for each language pair. All remaining constants---the ceiling percentile, $\epsilon_{\mathrm{cal}}$, $c_{\max}$, and the sharpening power used in Eq.~\eqref{eq:sharpened_matrix}---were fixed before evaluating the reported trace-level results and were held identical across all model--dataset evaluations. They were not selected against $S$, $\tau$, $H$, or the top-quartile rank statistic.

\subsubsection{What Calibration Does and Does Not Establish}

The floor anchors the calibrated scale at an empirical neutral level between unrelated content. The ceiling anchors it at an empirical estimate of the encoder's attainable similarity between semantically equivalent content for the same language pair. Calibration therefore removes pair-dependent drift at these two endpoints.

This procedure does not establish that an intermediate calibrated value has exactly the same semantic interpretation across different language pairs. Translation quality and embedding geometry may vary nonlinearly between the estimated endpoints. Accordingly, pooled absolute values should be interpreted descriptively.

The paper's central comparison is less demanding: examples are ranked within a fixed model--dataset--language-pair cell. For a fixed embedding space, any strictly monotone pair-specific rescaling preserves this ordering except where clipping creates ties. The rank-based comparison therefore does not require intermediate calibrated values to be globally semantically identical.

\subsubsection{Rank Invariance of the Headline Statistic}
\label{app:calibration_rank_invariance}

The rank comparison reported in \S\ref{sec:exp_baseline_comparison} is the share of pair--question instances in the top quartile of $H$ that do not fall in the top quartile of $S$, with both thresholds taken within a single evaluation and language pair. Within such a cell the calibration of Eq.~\eqref{eq:calibrated_similarity} is a fixed affine rescaling followed by clipping to $[0,1]$. It is monotone, so it preserves the ordering of instances in the cell and leaves the quantile thresholds at the same instances; the greedy alignment of \S\ref{sec:structural_score} selects matches by argmax and is likewise invariant. The only way calibration could move the statistic is by clipping distinct values onto a shared endpoint and creating ties. $H$ is an uncalibrated cosine, so any movement would fall entirely on $S$.

The residual tie effect is nil in practice. Recomputing the statistic on the uncalibrated scale---recovered by inverting the stored per-pair floor and ceiling---reproduces the calibrated value on all three Claude Opus~4.6 evaluations, with every difference below $10^{-4}$ (Table~\ref{tab:calibration_rank_invariance}). All three reproduced values lie inside the $52.1\%$--$67.3\%$ range reported in \S\ref{sec:exp_baseline_comparison}.

\begin{table}[h]
\caption{Headline rank statistic recomputed without calibration. ``top-$H$'' is the number of instances in the top quartile of $H$ within their language pair; the reported value is the share of those not in the top quartile of $S$.}
\label{tab:calibration_rank_invariance}
\centering
\small
\setlength{\tabcolsep}{4pt}
\begin{tabular}{@{}lrrrr@{}}
\toprule
\textbf{Dataset} & $P$ & top-$H$ & \textbf{calibrated} & $\Delta$ \\
\midrule
\textsc{StrategyQA} & 9{,}039 & 2{,}267 & 0.5554 & 0.0000 \\
\textsc{SciBench}   & 7{,}062 & 1{,}772 & 0.6552 & 0.0000 \\
\textsc{MMLU-ProX}  & 8{,}470 & 2{,}125 & 0.5906 & 0.0000 \\
\bottomrule
\end{tabular}
\end{table}

Pooling the quartile thresholds across the $21$ language pairs of an evaluation, rather than computing them within each pair, defines a different statistic: it folds between-pair differences in level into a comparison that is otherwise made within a pair. Under that definition the calibrated and uncalibrated values differ by $+1.0$, $-0.4$, and $-1.5$ percentage points on \textsc{StrategyQA}, \textsc{SciBench}, and \textsc{MMLU-ProX} respectively---small, and inconsistent in sign.

Calibration is therefore not what produces the reported divergence. What it provides is a readable scale: within a language pair, $0$ marks the unrelated-pair floor and $1$ the attainable self-translation ceiling, and $S$ and $H$ are placed on a common footing so that the $H{-}S$ gap is a well-defined quantity at all. Both are statements about levels, and the rank comparison the paper relies on does not depend on either.

\subsubsection{Ceiling-Percentile Sensitivity}

We assess the stability of the estimated ceilings by varying the self-translation percentile while holding the sampled translation pool fixed. This analysis measures sensitivity of the endpoint estimator rather than sensitivity to a change in encoder.

\subsubsection{Translator Independence}

The ceiling translator can, in some evaluations, belong to the same broad model family as the model whose traces are being analyzed. We test whether this creates a family-specific calibration artifact by re-estimating the language-pair ceilings with an independent-family translator and recomputing the metrics; per-pair values and rank correlations are reported in Appendix~\ref{app:translator-independence}. The experiment tests the dependence of the estimated calibration endpoints on the translation model. It does not imply that machine translation is equally reliable for every language, particularly for lower-resource languages.

\subsection{Sharpening Sensitivity}

The soft assignment sharpens step-pair similarities as $\widetilde{\mathbf{A}}_{i,j}=\mathbf{A}_{i,j}^{\,p}$ with $p=3$, concentrating each source step's assignment mass on its strongest candidate matches while retaining uncertainty over multiple candidates. We compare this default against a power family, $p\in\{1,2,3,5,10\}$, and a softmax-temperature family, $P_{i,j}\propto\exp(\mathbf{A}_{i,j}/T)$ with $T\in\{0.05,0.1,0.2,0.5,1.0\}$. Because different schemes distribute probability mass differently, their absolute $\tau$ values need not coincide, so robustness is assessed through preservation of rankings across evaluation cells. Full results are reported in Appendix~\ref{app:sharpening_ablation}.

\subsection{Alternative No-Tie Order Score}
\label{app:no_tie_tau}

The main paper reports the tie-aware order score from Eq.~\eqref{eq:directed_tau}. For completeness, we also define a variant that conditions only on concordant and discordant mass:
\begin{align}
\tau^{a\rightarrow b}_{\mathrm{no\text{-}tie}}
=
\frac{
M^{a\rightarrow b}_{\mathrm{con}}
-
M^{a\rightarrow b}_{\mathrm{dis}}
}{
M^{a\rightarrow b}_{\mathrm{con}}
+
M^{a\rightarrow b}_{\mathrm{dis}}
}.
\label{eq:no_tie_tau_directed}
\end{align}
The corresponding symmetrized score is
\begin{align}
\tau_{\mathrm{no\text{-}tie}}(r_a,r_b)
=
\frac{1}{2}
\left(
\tau^{a\rightarrow b}_{\mathrm{no\text{-}tie}}
+
\tau^{b\rightarrow a}_{\mathrm{no\text{-}tie}}
\right).
\label{eq:no_tie_tau}
\end{align}

The no-tie score measures relative ordering conditional on the two source steps being assigned to distinct target positions. The tie-aware score additionally penalizes concentration of multiple source steps on the same target step and is therefore used as the primary metric.

Both directed variants are undefined when their denominators are zero. A symmetrized score is reported as undefined if either direction is undefined.

\subsection{Controlled Metric Checks}

We use controlled perturbations to verify that the metrics respond to their intended dimensions. Removing a reasoning step should reduce structural similarity, because one trace no longer covers all content in the other. Rewriting a step without altering its semantic content should preserve structural similarity, subject to the multilingual encoder's robustness. Dividing one reasoning move into multiple steps, or combining adjacent moves, should be tolerated by the many-to-one correspondences in the structural score. Permuting otherwise unchanged steps should leave structural similarity relatively stable while reducing order consistency. Appendix~\ref{sec:verification} reports $S$, $\tau$, and $H$ for each perturbation family. These checks establish dimension-specific metric behavior under known transformations; they complement, but do not replace, validation on naturally generated multilingual traces.

\subsection{Whole-Answer Holistic Similarity}
\label{sec:raw_semantic}

The holistic baseline embeds each complete visible reasoning trace as a single sequence. Let $\psi$ denote the whole-response encoder. We define
\begin{align}
H(r_a,r_b)
=
\cos\!\big(
\psi(r_a),
\psi(r_b)
\big).
\label{eq:holistic_similarity}
\end{align}

Unlike $S$, the holistic baseline does not align individual steps or identify where two traces differ. Unlike $\tau$, it does not explicitly measure preservation of step order. It serves as a comparison against the common practice of representing an entire response with a single embedding.

\paragraph{Why BGE-M3 for $H$ and LaBSE for $S$.}
Step-level structural similarity $S$ uses LaBSE~\citep{feng2022labse} because it is the standard cross-lingual sentence encoder and individual reasoning steps comfortably fit its 256-token context. The whole-trace baseline $H$ instead uses BGE-M3~\citep{chen2024bgem3} because LaBSE's 256-token limit truncates a substantial fraction of our reasoning traces (especially on \textsc{SciBench} and \textsc{StrategyQA}, where a single verified trace frequently exceeds 1{,}000 tokens), and in preliminary tests the resulting LaBSE-on-truncated-trace cosines were uninformative as a holistic signal. BGE-M3's 8K-token window covers the entire trace without truncation in our evaluations, which is the property a holistic baseline is meant to provide. Adopting LaBSE on both sides would therefore not isolate the aggregation effect from $H{-}S$; it would replace the encoder asymmetry with a truncation artifact whose magnitude depends on dataset-specific trace-length distributions. We accordingly do not interpret the absolute magnitude of $H{-}S$ in Table~\ref{tab:aggregate_results} as a calibrated distance, and rely on the scale-free rank comparison in Figure~\ref{fig:H_vs_S} for the qualitative claim that holistic similarity overstates structural consistency.

\subsection{Interpretation of Structural and Order Scores}
\label{app:metric_interpretation}

The two primary trace-level metrics capture complementary properties.

\paragraph{High $S$, high $\tau$.}
The traces contain similar step-level content in a similar order.

\paragraph{High $S$, low $\tau$.}
The traces contain similar reasoning moves but arrange them differently.

\paragraph{Low $S$, high $\tau$.}
The subset of content that can be aligned appears in a consistent order, but substantial content is absent or unmatched.

\paragraph{Low $S$, low $\tau$.}
The traces differ both in step-level content and in the ordering of their matched components.

Neither metric determines whether a difference is harmful. They identify and decompose observable differences in the visible explanatory artifacts, which can then be inspected by human or automated evaluators.

\subsection{Relation to Prior Step-Level Metrics}
\label{sec:relation}

We compare against prior step-level metrics to clarify which parts of the framework are reused and which are specific to cross-lingual reasoning. The method recomposes BERTScore-style greedy matching~\citep{zhang2020bertscore}, ROSCOE-style step alignment~\citep{golovneva2023roscoe}, multilingual sentence embeddings~\citep{feng2022labse}, and soft ordering ideas from continuous sorting~\citep{mena2018sinkhorn,grover2019neuralsort}.

The main difference is that our alignment is calibrated per language pair before aggregation. Prior step metrics typically use raw within-language cosine scores, while cross-lingual traces require a pair-specific unrelated-sentence floor and self-translation ceiling so that a score for, for example, \texttt{en-es} has the same interpretation as a score for \texttt{hi-zh}.

The second difference is that we report structure and order separately. A single alignment score can hide whether two traces contain different reasoning moves or contain similar moves in a different sequence; $S$ and $\tau$ are kept separate so these two kinds of divergence remain visible.

\appendixsectionbreak
\section{Supplementary Experimental Details}
\label{app:experiments}

This appendix collects the experimental details that support the three main results without expanding Section~\ref{sec:results}. We place the dataset, model, verification, accuracy-audit, and calibration materials next to the discussion they support.

\subsection{Datasets, Models, and Inference}
\label{app:exp_setup_details}

All reported evaluations use the same seven languages: Arabic (ar), English (en), Spanish (es), Hindi (hi), Russian (ru), Vietnamese (vi), and Chinese (zh). This gives $\binom{7}{2}=21$ unordered language pairs for each model--dataset run. We sample 500 problems per run from each benchmark.

\textsc{MMLU-ProX}~\citep{mmluprox2025} supplies parallel multiple-choice questions across academic and professional domains. We verify traces by extracting the final option and comparing it with the dataset answer key; ambiguous outputs are checked by an answer-consistency judge that sees the question, reference answer, and model final answer, but not any trace-similarity metric. \textsc{StrategyQA}~\citep{geva2021strategyqa} provides yes/no commonsense questions requiring implicit multi-step reasoning; we verify the final yes/no answer against the dataset label, with the same answer-consistency judge used for ambiguous outputs. \textsc{SciBench}~\citep{ye2024scibench} contains open-ended STEM problems with fixed numeric or symbolic answers; the verification step checks whether the model final answer is equivalent to the reference answer before the trace is used for divergence analysis.

The model suite comprises Claude Opus~4.6, Claude Haiku~4.5, DeepSeek-V3.2, GPT-OSS-120B, Qwen3-235B, and Qwen3-32B. All reported runs use greedy decoding.

\noindent{\bf Retention cascade.} For each 500-problem dataset--model run, we first verify each language-specific final answer against the reference. For every language whose answer is correct on a given question, an LLM trace judge then inspects the reasoning trace and emits \texttt{VALID} or \texttt{INVALID}. If any answer-correct trace for that question is judged \texttt{INVALID}, the question is dropped from trace aggregation, even if other answer-correct languages would otherwise form valid pairs. The surviving pair--question instances enter $S$, $\tau$, and $H$; one-language-correct cases are reserved for $\Delta_{\mathrm{acc}}$. Table~\ref{tab:aggregate_results} reports the retained question counts $Q$ and pair--question counts $P$ after this cascade.

\noindent{\bf Metric settings.} Step-level metrics use LaBSE~\citep{feng2022labse}, the holistic baseline $H$ uses BGE-M3~\citep{chen2024bgem3}, unrelated-pair floor sampling draws up to $10^{4}$ sentence pairs per language pair, the self-translation ceiling samples up to 500 sentences per direction, and the soft Kendall-$\tau$ sharpening power is 3. Appendix~\ref{app:sharpening_ablation} shows that the cross-evaluation $\bar\tau$ ranking is preserved with Spearman $\geq0.85$ across other powers and a softmax-temperature baseline. Aggregate means in Table~\ref{tab:aggregate_results} use $95\%$ percentile bootstrap intervals with 2{,}000 problem-level resamples, jointly resampling the 21 language-pair observations for each question.

\noindent{\bf Compute.} All LLM trace generation and ceiling-translation calls use hosted APIs (AWS Bedrock for Claude, DeepSeek, GPT-OSS, and Qwen endpoints); we therefore report no local GPU hours for inference. Sentence-level embedding, per-pair calibration, and $(S,\tau,H)$ computation for the full 18 model--dataset evaluations run on a single NVIDIA A100 and take roughly $25$ hours end-to-end, dominated by LaBSE and BGE-M3 encoding of the retained traces.

\subsection{Prompt Templates}
\label{app:prompt_templates}

Because the prompt elicits the step structure on which the parser and both step-level metrics depend, we give the templates in full. All inference uses a single fixed template family across every model, dataset, and language; no prompt is tuned per language. Decoding is greedy (temperature $0$) with an identical token budget and retry policy across all runs.

\noindent{\bf Shared output-format block.} Appended to every system prompt:

\begin{promptbox}{System prompt: shared output-format block}
\begin{verbatim}
OUTPUT FORMAT REQUIREMENTS:
- Write the reasoning as a numbered
  sequence of steps.
- Every reasoning line must begin with an
  English step header in exactly this
  format: Step 1:, Step 2:, Step 3:, ...
- Keep the step labels exactly as
  `Step N:` even when the reasoning
  content is in another language.
- Put only one logical step under each
  step header; multi-line steps are
  allowed when needed.
- End with exactly one final-answer line
  in this format:

  Final Answer: [your answer here]

- Do not wrap the final answer in a
  sentence.
- Do not omit the step headers.
\end{verbatim}
\end{promptbox}

\noindent{\bf Task-specific system prompts.} The preamble differs only in role and emphasis. \textsc{MMLU-ProX}: ``You are an expert problem solver. Answer the following multiple-choice question,'' followed by four instructions---analyze the options systematically; explain your reasoning step by step; end with the single best option letter; follow the output format exactly---then the block above, then a note that the final answer must be a single option label such as A, B, C, or D. \textsc{SciBench}: ``You are an expert mathematician and scientist. Solve the given problem carefully,'' followed by five instructions---show your reasoning step by step; perform calculations explicitly and do not skip important intermediate steps; if the problem has sub-parts, solve them in a clean logical order; be concise but complete; follow the output format exactly---then the block above. \textsc{StrategyQA}: ``You are an expert at commonsense and multi-hop reasoning. Answer the given yes/no question,'' followed by six instructions---identify the implicit sub-questions that must be resolved; establish the relevant facts for each, one sub-question per step; combine the established facts explicitly into a single conclusion; rely on general world knowledge; be concise but complete; follow the output format exactly---then the block above, then a note that the final answer must be exactly \texttt{Yes} or \texttt{No}, written in English, with no additional words.

\newpage
\noindent{\bf User messages.}

\begin{promptbox}{User message}
\begin{verbatim}
MMLU-ProX:
  Question:\n{question}\n\n
  Options:\n{A. ...}\n{B. ...}
SciBench:
  Problem (unit: {unit}):\n{question}
StrategyQA:
  Question:\n{question}
\end{verbatim}
\end{promptbox}

\noindent{\bf Language-specific instructions.} English receives none. Every non-English language appends exactly one sentence to the system prompt:

\begin{promptbox}{Language-specific instruction (non-English)}
\begin{verbatim}
IMPORTANT: Write the content of every
reasoning step and the final answer in
{Language}. Keep the step headers exactly
as `Step 1:`, `Step 2:`, etc., and keep
`Final Answer:` in English.
\end{verbatim}
\end{promptbox}

\noindent where \texttt{\{Language\}} is the English language name. For \textsc{SciBench} items already supplied in the target language, the wording instead notes that the problem is already in that language and asks for the reasoning content and final answer in it. For \textsc{StrategyQA}, the sentence requests the reasoning content in the target language with the answer token as the English word \texttt{Yes} or \texttt{No}, which restores rule-based answer checking in all seven languages. This constraint touches only the final-answer line, at which the step parser terminates, so $S$ and $\tau$ are computed on identical step content either way.

\noindent{\bf Effect on cross-lingual comparison.} Prompting shapes granularity and parser compliance, and the design controls for this in the way that matters here: the segmentation cue is an English literal held constant across all seven languages. Step boundaries are therefore never recovered by language-specific heuristics, and differences in $S$ or $\tau$ across languages cannot arise from differing header conventions. The parser audit in Appendix~\ref{app:parser_coverage} bears this out, with effectively complete coverage in every language.

\subsection{Outcome-Level Accuracy Audit}
\label{app:accuracy_summary}

\begin{figure*}[h]
  \centering
  \includegraphics[width=0.92\textwidth]{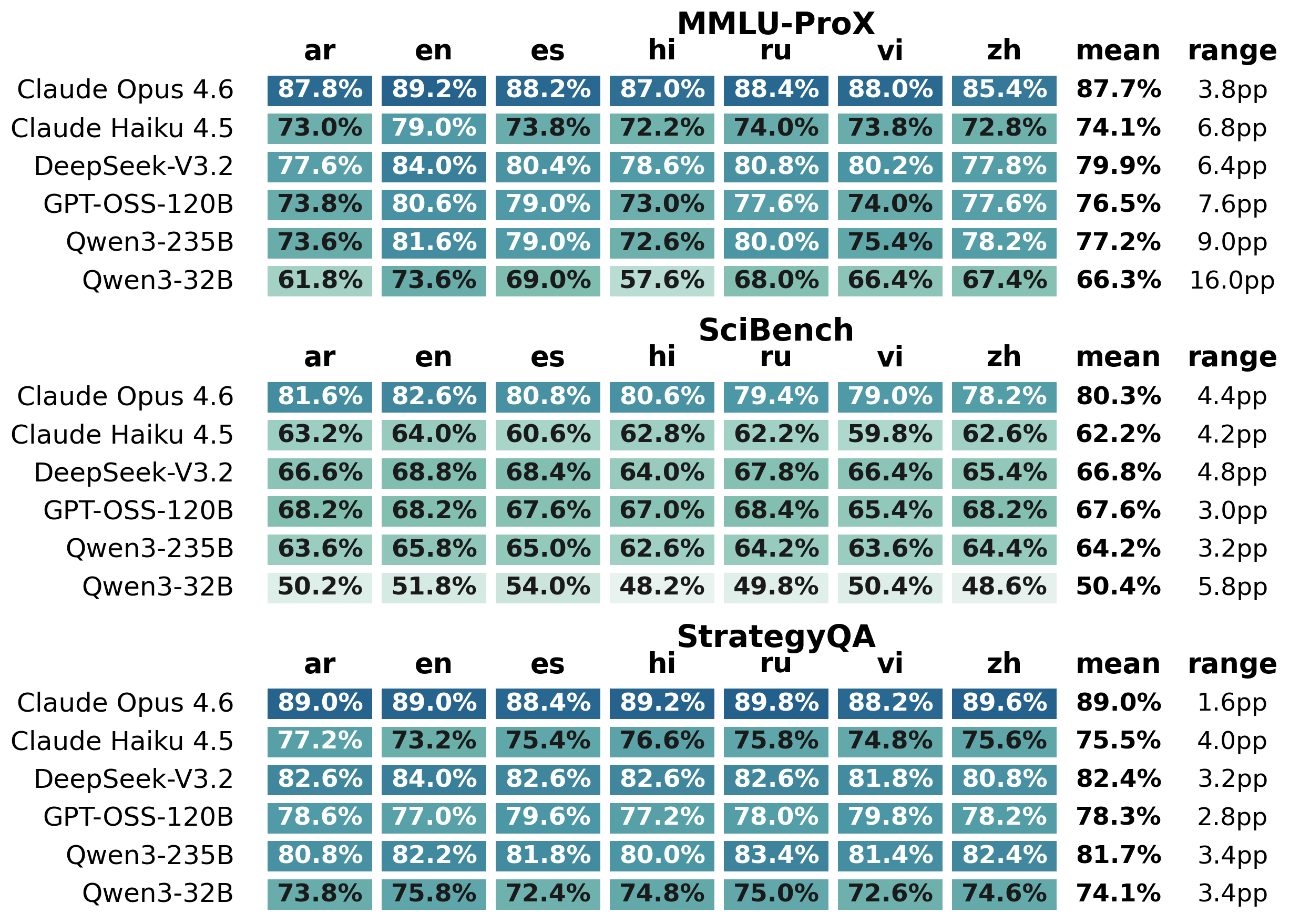}
  \caption{Outcome-level accuracy heatmap across the multilingual evaluations. Within each panel, columns are ar, en, es, hi, ru, vi, zh, mean, and range. Cell values are computed over the \emph{full 500-problem inference set} for each (model, dataset) pair, prior to the trace-validity filter that defines $Q$ in Table~\ref{tab:aggregate_results}. The figure audits final-answer performance, while Section~\ref{sec:results} focuses on trace divergence after correctness filtering.}

  \label{fig:app_accuracy_summary}
\end{figure*}

Figure~\ref{fig:app_accuracy_summary} reports per-language final-answer accuracy for the same model--dataset evaluations summarized in Table~\ref{tab:aggregate_results}. This audit contextualizes the correctness filter used before trace-level aggregation: accuracy varies across languages and datasets, but the main divergence metrics are computed only after both compared traces pass answer verification.

\subsection{Calibration Intervals}
\label{app:calibration}

We report calibration ranges to show how much the raw multilingual cosine scale varies before pair-specific normalization. Table~\ref{tab:calibration} summarises the sentence-level intervals used in the main analyses: \textsc{StrategyQA} has the largest floor spread, while \textsc{SciBench} and \textsc{MMLU-ProX} often hit the ceiling cap.

\begin{table*}[!htbp]
\caption{Sentence-level calibration intervals. Ranges are over language pairs in the corresponding evaluation. Ceiling values are capped at $c_{\max}=0.98$, so datasets whose self-translation similarities reach the cap can have zero or near-zero ceiling spread.}
\label{tab:calibration}
\centering
\small
\begin{tabular}{lrrrr}
\toprule
{\bf Evaluation} & {\bf floor range} & {\bf floor spread} & {\bf ceiling range} & {\bf ceiling spread} \\
\midrule
\textsc{MMLU-ProX}, Claude Opus 4.6 & 0.187--0.331 & 0.144 & 0.960--0.980 & 0.020 \\
\textsc{StrategyQA}, Claude Opus 4.6 & 0.157--0.381 & 0.224 & 0.912--0.968 & 0.056 \\
\textsc{SciBench}, Claude Opus 4.6 & 0.292--0.352 & 0.059 & 0.980--0.980 & 0.000 \\
\bottomrule
\end{tabular}
\end{table*}

\appendixsectionbreak
\section{Additional Results}
\label{app:additional_results}

\subsection{Translator-Independence Probe}
\label{app:translator-independence}

\noindent{\bf Methodological concern.} The default ceiling estimation in \S\ref{sec:calibration} uses an LLM translator from the same family (Claude Opus~4.6) as several of the evaluated models. A strong translator whose idiolect matches the evaluated model could inflate ceilings on exactly those pairs where the model is already consistent, partially undoing the calibration. We probe this co-dependence by re-estimating the per-pair sentence ceiling with an independent translator from a different model family and recomputing the divergence metrics on the same retained pair--question instances.

\noindent{\bf Setup.} The probe uses \textsc{StrategyQA} on Claude Opus~4.6 ($n=451$ retained verified-correct questions over 21 language pairs). The independent translator is \texttt{Mistral-Large-3-675B-Instruct}. The unrelated-pair sentence floor is shared between the two analyses (the floor does not depend on the translator); only the ceiling estimation changes. All other settings (LaBSE step embeddings, BGE-M3 holistic embedding, $90$th-percentile self-translation ceiling, $\epsilon_{\text{cal}}=0.10$, $c_{\max}=0.98$, sharpening power 3) are identical.

\noindent{\bf Result.} Per-pair rankings of the three metrics are stable under the alternative ceiling. Across 21 language pairs, the Spearman rank correlations between the original (Claude-Opus ceiling) and the independent-translator (Mistral-Large-3 ceiling) estimates are $\rho_{C}=+0.92$ on the ceiling itself, $\rho_{\bar S}=+0.89$ on mean structural similarity, $\rho_{\bar\tau}=+0.87$ on mean soft Kendall-$\tau$, and $\rho_{\bar H}=+0.99$ on the holistic baseline (sanity: $H$ does not use the ceiling and should be near-invariant). Pooled aggregate means change by $\Delta\bar S=+0.002$ and $\Delta\bar\tau=-0.003$. Per-pair values are reported in Table~\ref{tab:translator_independence}.

\begin{table}[h]
\caption{Translator-independence probe on \textsc{StrategyQA} Claude Opus~4.6 ($n=451$, 21 pairs). $C$ is the per-pair sentence ceiling. $\Delta\bar S$ and $\Delta\bar\tau$ are the (Mistral~$-$~Opus) shifts in mean structural similarity and mean soft Kendall-$\tau$ with-tie. $H$ does not depend on the ceiling and is unchanged.}
\label{tab:translator_independence}
\centering
\small
\begin{tabular}{lrrrrr}
\toprule
{\bf pair} & $C_{\text{Opus}}$ & $C_{\text{Mist}}$ & $\Delta C$ & $\Delta\bar S$ & $\Delta\bar\tau$ \\
\midrule
ar-en & 0.943 & 0.941 & $-0.002$ & $+0.001$ & $-0.013$ \\
ar-es & 0.960 & 0.957 & $-0.003$ & $+0.002$ & $+0.003$ \\
ar-hi & 0.935 & 0.934 & $-0.001$ & $+0.002$ & $+0.004$ \\
ar-ru & 0.943 & 0.935 & $-0.008$ & $+0.006$ & $-0.005$ \\
ar-vi & 0.922 & 0.923 & $+0.001$ & $-0.000$ & $-0.005$ \\
ar-zh & 0.950 & 0.947 & $-0.003$ & $+0.004$ & $-0.008$ \\
en-es & 0.962 & 0.962 & $-0.000$ & $+0.000$ & $+0.017$ \\
en-hi & 0.929 & 0.931 & $+0.003$ & $-0.001$ & $+0.012$ \\
en-ru & 0.934 & 0.941 & $+0.008$ & $-0.005$ & $+0.007$ \\
en-vi & 0.936 & 0.936 & $+0.001$ & $-0.001$ & $+0.011$ \\
en-zh & 0.933 & 0.941 & $+0.007$ & $-0.006$ & $+0.009$ \\
es-hi & 0.952 & 0.958 & $+0.006$ & $+0.002$ & $-0.007$ \\
es-ru & 0.968 & 0.970 & $+0.002$ & $-0.003$ & $-0.005$ \\
es-vi & 0.962 & 0.958 & $-0.004$ & $+0.002$ & $-0.007$ \\
es-zh & 0.945 & 0.941 & $-0.004$ & $+0.003$ & $-0.006$ \\
hi-ru & 0.963 & 0.955 & $-0.008$ & $+0.019$ & $-0.032$ \\
hi-vi & 0.968 & 0.963 & $-0.005$ & $+0.023$ & $-0.036$ \\
hi-zh & 0.912 & 0.925 & $+0.013$ & $-0.005$ & $-0.021$ \\
ru-vi & 0.960 & 0.959 & $-0.001$ & $+0.001$ & $-0.001$ \\
ru-zh & 0.926 & 0.917 & $-0.009$ & $+0.004$ & $+0.007$ \\
vi-zh & 0.924 & 0.933 & $+0.009$ & $-0.013$ & $+0.012$ \\
\midrule
mean  & 0.944 & 0.944 & $+0.000$ & $+0.002$ & $-0.003$ \\
\bottomrule
\end{tabular}
\end{table}

\noindent{\bf Interpretation.} Per-pair ceiling changes are bounded by $|\Delta C|\le 0.013$ and pooled-mean shifts are within $\pm 0.003$ for both $\bar S$ and $\bar\tau$. The largest individual shifts on the downstream metrics fall on Hindi-involving pairs: the top three $|\Delta\bar\tau|$ shifts are \texttt{hi-vi} ($0.036$), \texttt{hi-ru} ($0.032$), and \texttt{hi-zh} ($0.021$), and the top two $|\Delta\bar S|$ shifts are \texttt{hi-vi} ($0.023$) and \texttt{hi-ru} ($0.019$); on the ceiling itself, $|\Delta C|$ is largest for \texttt{hi-zh} ($+0.013$), with \texttt{ru-zh}, \texttt{vi-zh}, \texttt{ar-ru}, \texttt{en-ru}, and \texttt{hi-ru} clustered between $0.008$ and $0.009$, so $|\Delta C|$ alone does not single out Hindi. Together these patterns suggest that ceiling estimation for Hindi-side directions is the most translator-dependent of the seven languages tested, with the bidirectional soft Kendall-$\tau$ amplifying this sensitivity relative to the calibrated structural score. Relative pair rankings nonetheless remain stable ($\rho_{\bar S}=+0.89$, $\rho_{\bar\tau}=+0.87$). The headline conclusion---$H$ systematically exceeds $\bar S$ on verified-correct trace pairs---does not depend on the choice of ceiling translator, and the per-pair calibration findings are therefore not an artefact of using a Claude-family translator on Claude-family traces.

\subsection{Pair-Level Result Tables}
\label{app:pair_tables}

We keep pair-level values in the appendix so the main results can focus on aggregate hypotheses. Table~\ref{tab:opus_pair_results} gives the Claude Opus~4.6 pair-level views for all three evaluated datasets.

The pair-level rankings show that language-pair difficulty is not fixed across tasks. English--Spanish and Spanish--Russian are consistently among the strongest pairs, but Hindi- and Chinese-involving pairs shift more substantially across datasets, reinforcing the main claim that trace divergence reflects model--dataset--language interactions rather than a single language-pair ordering.

\begin{table*}[!htbp]
\caption{Claude Opus~4.6 pair-level divergence across datasets. Each dataset block is independently sorted by structural similarity $S$ in descending order. $N$ is the number of problems on which both languages produced an answer-verified response \emph{and} both reasoning traces were marked \texttt{VALID} by the LLM trace-judge in Appendix~\ref{app:exp_setup_details} (the same cascade used for $Q$ in Table~\ref{tab:aggregate_results}); it is therefore typically smaller than the count of problems on which both languages are merely answer-correct. $\Delta_{\mathrm{acc}}$ reports pair-level one-language-correct disagreement for \textsc{MMLU-ProX}.}

\label{tab:opus_pair_results}
\centering
\scriptsize
\setlength{\tabcolsep}{2.2pt}
\resizebox{\textwidth}{!}{%
\begin{tabular}{lrrrrr@{\hspace{1.0em}}lrrrr@{\hspace{1.0em}}lrrrr}
\toprule
\multicolumn{6}{c}{\bf {\textsc{MMLU-ProX}}} &
\multicolumn{5}{c}{\bf {\textsc{StrategyQA}}} &
\multicolumn{5}{c}{\bf {\textsc{SciBench}}} \\
\cmidrule(lr){1-6}\cmidrule(lr){7-11}\cmidrule(lr){12-16}
{\bf pair} & $N$ & $\Delta_{\mathrm{acc}}$ & $S$ & $\tau$ & $H$ &
{\bf pair} & $N$ & $S$ & $\tau$ & $H$ &
{\bf pair} & $N$ & $S$ & $\tau$ & $H$ \\
\midrule
es-ru & 408 & 4.2\% & 0.610 & 0.451 & 0.819 & es-ru & 430 & 0.625 & 0.462 & 0.836 & en-es & 338 & 0.669 & 0.492 & 0.863 \\
en-es & 411 & 3.8\% & 0.604 & 0.449 & 0.859 & ar-hi & 434 & 0.622 & 0.419 & 0.798 & es-ru & 336 & 0.645 & 0.483 & 0.812 \\
ru-vi & 409 & 4.8\% & 0.585 & 0.429 & 0.799 & ar-zh & 430 & 0.618 & 0.420 & 0.775 & es-vi & 336 & 0.638 & 0.471 & 0.784 \\
ar-es & 409 & 3.2\% & 0.583 & 0.423 & 0.789 & vi-zh & 431 & 0.614 & 0.410 & 0.813 & ru-vi & 334 & 0.625 & 0.464 & 0.778 \\
en-zh & 400 & 6.6\% & 0.583 & 0.416 & 0.813 & en-zh & 431 & 0.613 & 0.407 & 0.820 & en-vi & 335 & 0.621 & 0.441 & 0.801 \\
es-zh & 401 & 4.4\% & 0.583 & 0.425 & 0.787 & es-zh & 427 & 0.611 & 0.409 & 0.798 & en-ru & 340 & 0.620 & 0.456 & 0.821 \\
es-vi & 406 & 5.4\% & 0.581 & 0.422 & 0.803 & ar-es & 429 & 0.611 & 0.444 & 0.792 & es-hi & 340 & 0.610 & 0.440 & 0.767 \\
ar-hi & 399 & 5.6\% & 0.578 & 0.416 & 0.777 & ru-zh & 431 & 0.611 & 0.416 & 0.801 & vi-zh & 328 & 0.609 & 0.446 & 0.778 \\
es-hi & 397 & 7.2\% & 0.572 & 0.408 & 0.788 & hi-zh & 431 & 0.609 & 0.411 & 0.792 & ar-es & 339 & 0.608 & 0.451 & 0.777 \\
ar-en & 411 & 3.4\% & 0.570 & 0.417 & 0.794 & es-vi & 430 & 0.609 & 0.436 & 0.846 & es-zh & 334 & 0.607 & 0.450 & 0.761 \\
ar-ru & 407 & 4.6\% & 0.570 & 0.425 & 0.792 & ru-vi & 431 & 0.608 & 0.428 & 0.821 & en-zh & 332 & 0.606 & 0.429 & 0.783 \\
ar-zh & 398 & 5.6\% & 0.569 & 0.400 & 0.777 & ar-ru & 432 & 0.605 & 0.436 & 0.802 & hi-vi & 336 & 0.606 & 0.436 & 0.754 \\
en-ru & 411 & 4.4\% & 0.569 & 0.444 & 0.828 & en-es & 429 & 0.602 & 0.428 & 0.882 & ar-hi & 342 & 0.602 & 0.439 & 0.752 \\
hi-vi & 402 & 5.4\% & 0.568 & 0.399 & 0.782 & en-vi & 427 & 0.601 & 0.440 & 0.868 & ar-vi & 337 & 0.602 & 0.450 & 0.763 \\
hi-ru & 400 & 6.6\% & 0.566 & 0.408 & 0.788 & en-ru & 431 & 0.593 & 0.431 & 0.843 & en-hi & 341 & 0.598 & 0.420 & 0.788 \\
ar-vi & 405 & 5.0\% & 0.565 & 0.414 & 0.785 & ar-vi & 430 & 0.592 & 0.421 & 0.790 & ar-en & 338 & 0.597 & 0.427 & 0.783 \\
ru-zh & 397 & 7.0\% & 0.560 & 0.429 & 0.794 & es-hi & 430 & 0.589 & 0.421 & 0.823 & hi-ru & 339 & 0.595 & 0.432 & 0.761 \\
vi-zh & 400 & 5.8\% & 0.555 & 0.423 & 0.801 & ar-en & 431 & 0.577 & 0.398 & 0.799 & ru-zh & 331 & 0.594 & 0.444 & 0.770 \\
en-vi & 408 & 5.2\% & 0.548 & 0.421 & 0.820 & hi-ru & 431 & 0.564 & 0.414 & 0.822 & ar-zh & 334 & 0.594 & 0.417 & 0.751 \\
en-hi & 399 & 7.0\% & 0.546 & 0.406 & 0.811 & en-hi & 431 & 0.561 & 0.377 & 0.844 & hi-zh & 334 & 0.589 & 0.425 & 0.744 \\
hi-zh & 392 & 7.2\% & 0.546 & 0.398 & 0.772 & hi-vi & 432 & 0.545 & 0.400 & 0.817 & ar-ru & 338 & 0.587 & 0.427 & 0.777 \\
\bottomrule
\end{tabular}
}
\end{table*}

\subsection{Cross-Dataset and Cross-Model Patterns}
\label{app:replication}

We compare shared language pairs across datasets to check whether the aggregate conclusion depends on a single corpus. Table~\ref{tab:replication} shows that $H$ rankings replicate strongly across datasets, $\tau$ rankings replicate moderately, and $S$ is more task-sensitive, as expected for a metric affected by how each dataset encourages step granularity.

\begin{table*}[!htbp]
\caption{Cross-dataset replication on the 21 shared language pairs for Claude Opus~4.6. Each cell shows the Spearman rank correlation across language-pair means with a two-sided $p$-value.}
  \label{tab:replication}
  \centering
\small
  \begin{tabular}{lccc}
    \toprule
    {\bf metric} & {\bf \textsc{MMLU-ProX}/\textsc{StrategyQA}} & {\bf \textsc{MMLU-ProX}/\textsc{SciBench}} & {\bf \textsc{StrategyQA}/\textsc{SciBench}} \\
    \midrule
    $S$ & $0.37$ ($p{=}0.095$) & $0.55$ ($p{=}0.010$) & $0.11$ ($p{=}0.642$) \\
    $\tau$ & $0.52$ ($p{=}0.015$) & $0.76$ ($p{<}0.001$) & $0.62$ ($p{=}0.002$) \\
    $H$ & $0.80$ ($p{<}0.001$) & $\mathbf{0.98}$ ($p{<}0.001$) & $0.78$ ($p{<}0.001$) \\
    \bottomrule
  \end{tabular}
\end{table*}

Beyond rank replication, the datasets differ in the type of divergence they expose. Averaging over the six tested models, \textsc{MMLU-ProX} has the lowest mean structural similarity ($\bar S=0.536$) but the highest mean order consistency of the three ($\bar\tau=0.399$), suggesting that multiple-choice academic questions often preserve a broad procedural order while still varying the content or granularity of intermediate steps. \textsc{SciBench} has the highest mean structural similarity ($\bar S=0.571$) but lower mean order consistency ($\bar\tau=0.373$), consistent with quantitative STEM solutions that reuse similar formulas and quantities while allowing languages to reorder setup, substitution, conversion, and rounding. \textsc{StrategyQA} sits between them structurally ($\bar S=0.551$) but has the highest holistic similarity ($\bar H=0.813$) and the largest mean holistic-to-structural gap ($\overline{H-S}=0.263$), indicating that yes/no commonsense answers can sound semantically similar even when premise interpretation or evidence selection changes inside the trace.

Model comparisons within each dataset show that these are not purely dataset-level effects. On \textsc{MMLU-ProX}, Claude Opus~4.6 has the strongest structural alignment ($S=0.569$), while GPT-OSS-120B has the strongest order consistency ($\tau=0.469$); on \textsc{SciBench}, Qwen3-32B reaches the highest structural score ($S=0.629$) but much lower order consistency ($\tau=0.332$), whereas Claude Haiku~4.5 has the highest order score ($\tau=0.470$). On \textsc{StrategyQA}, Claude Opus~4.6 leads both $S$ and $\tau$, while Claude Haiku~4.5 has the highest holistic score ($H=0.857$) but also the largest diagnostic gap ($H-S=0.327$). Model quality under final-answer or whole-output similarity therefore does not translate monotonically into step-level consistency, and the same model family can shift its relative behaviour across datasets.

\subsection{Rank-Based Holistic--Structural Comparison}
\label{app:rank_h_s}

Because $S$ is computed from pair-calibrated sentence similarities while $H$ is an uncalibrated whole-answer cosine, we supplement the raw $H$-versus-$S$ plots with a rank-based comparison that uses only the within-evaluation, within-language-pair distributions. This avoids treating $H-S$ as a calibrated metric distance.

\begin{figure*}[!t]
  \centering
  \begin{subfigure}[t]{0.32\textwidth}
    \centering
    \includegraphics[width=\linewidth]{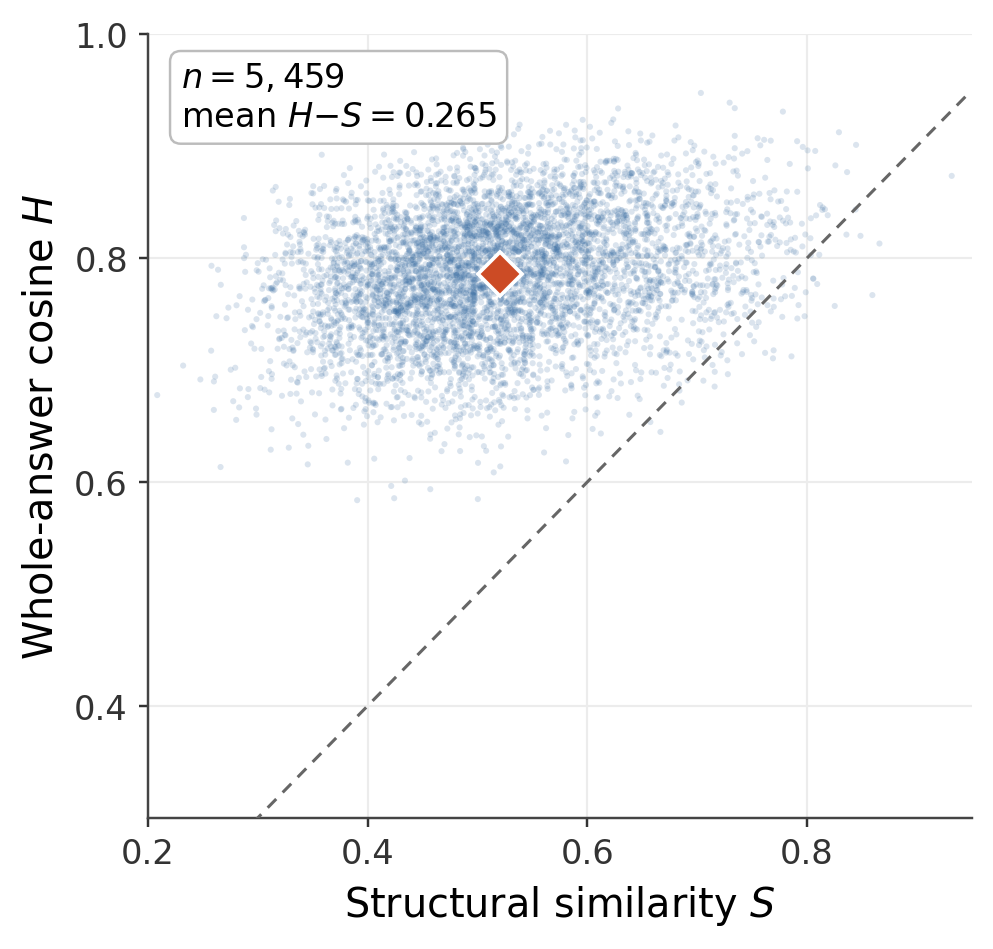}
    \subcaptiontext{\textsc{MMLU-ProX}, Claude Haiku~4.5}
  \end{subfigure}\hfill
  \begin{subfigure}[t]{0.32\textwidth}
    \centering
    \includegraphics[width=\linewidth]{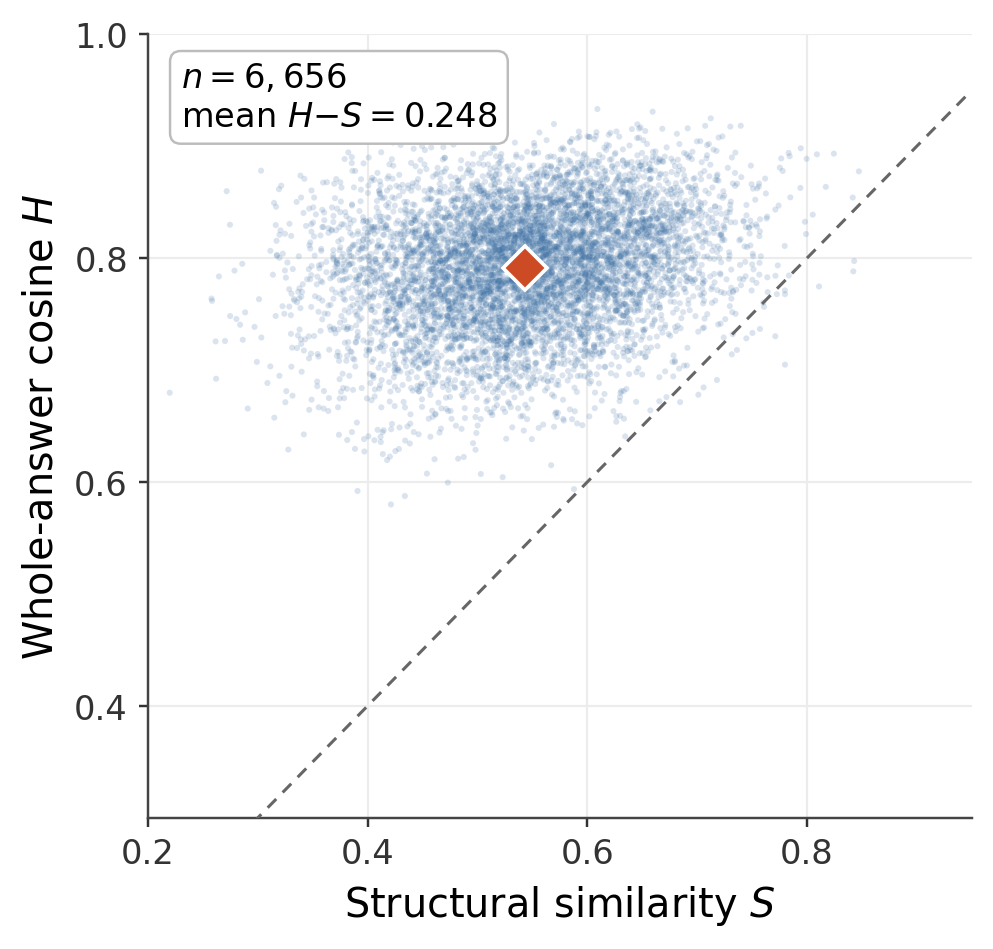}
    \subcaptiontext{\textsc{MMLU-ProX}, Qwen3-235B}
  \end{subfigure}\hfill
  \begin{subfigure}[t]{0.32\textwidth}
    \centering
    \includegraphics[width=\linewidth]{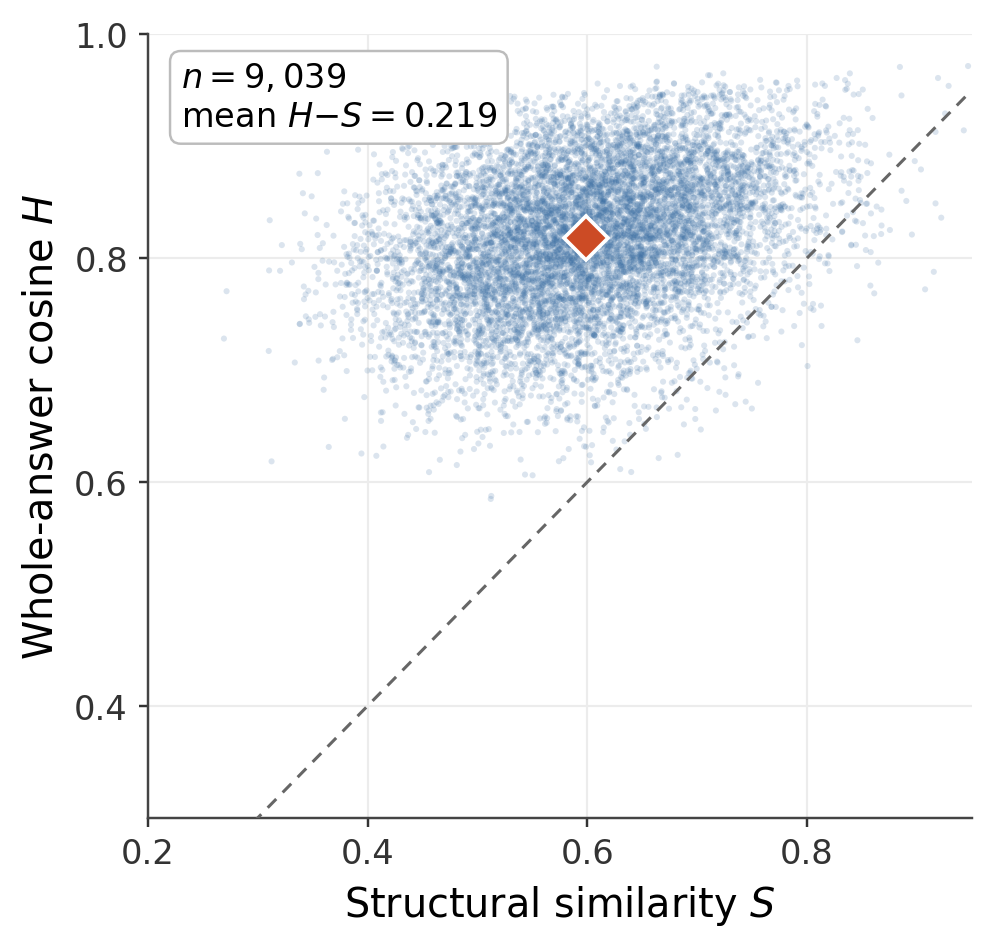}
    \subcaptiontext{\textsc{StrategyQA}, Claude Opus~4.6}
  \end{subfigure}

  \vspace{0.8em}
  \begin{subfigure}[t]{0.32\textwidth}
    \centering
    \includegraphics[width=\linewidth]{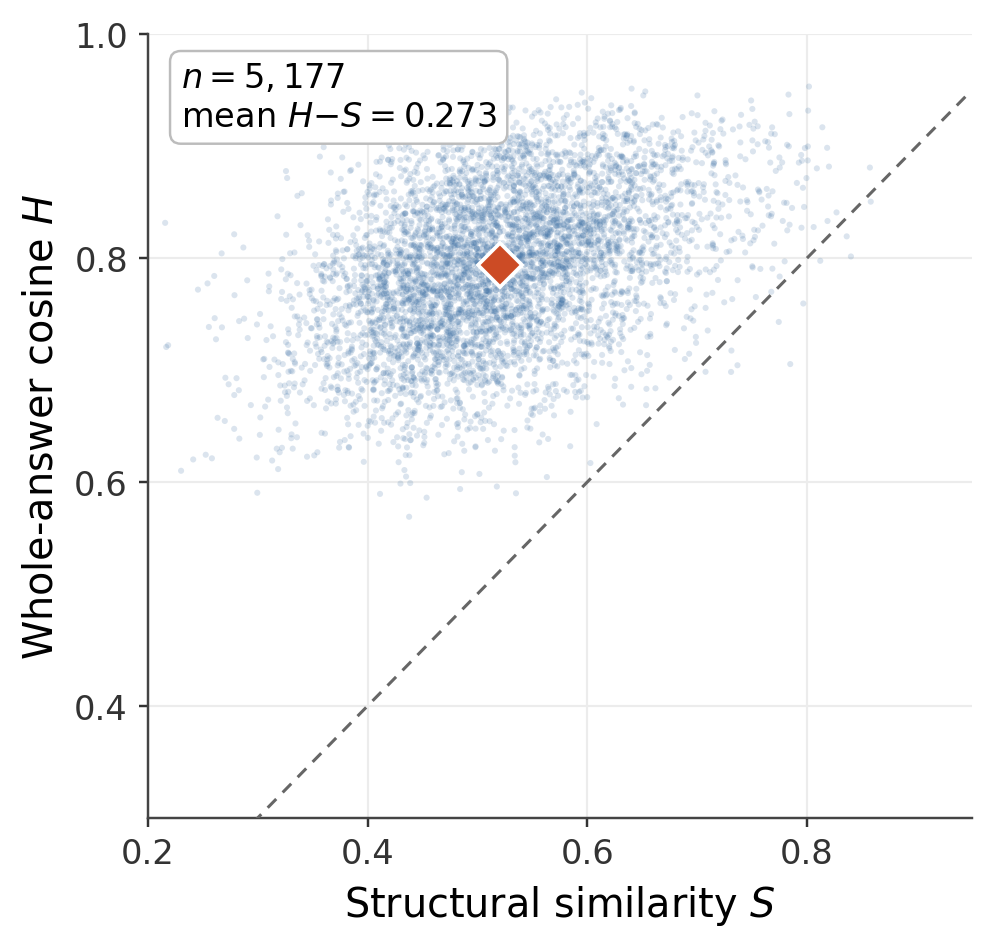}
    \subcaptiontext{\textsc{StrategyQA}, Qwen3-32B}
  \end{subfigure}\hfill
  \begin{subfigure}[t]{0.32\textwidth}
    \centering
    \includegraphics[width=\linewidth]{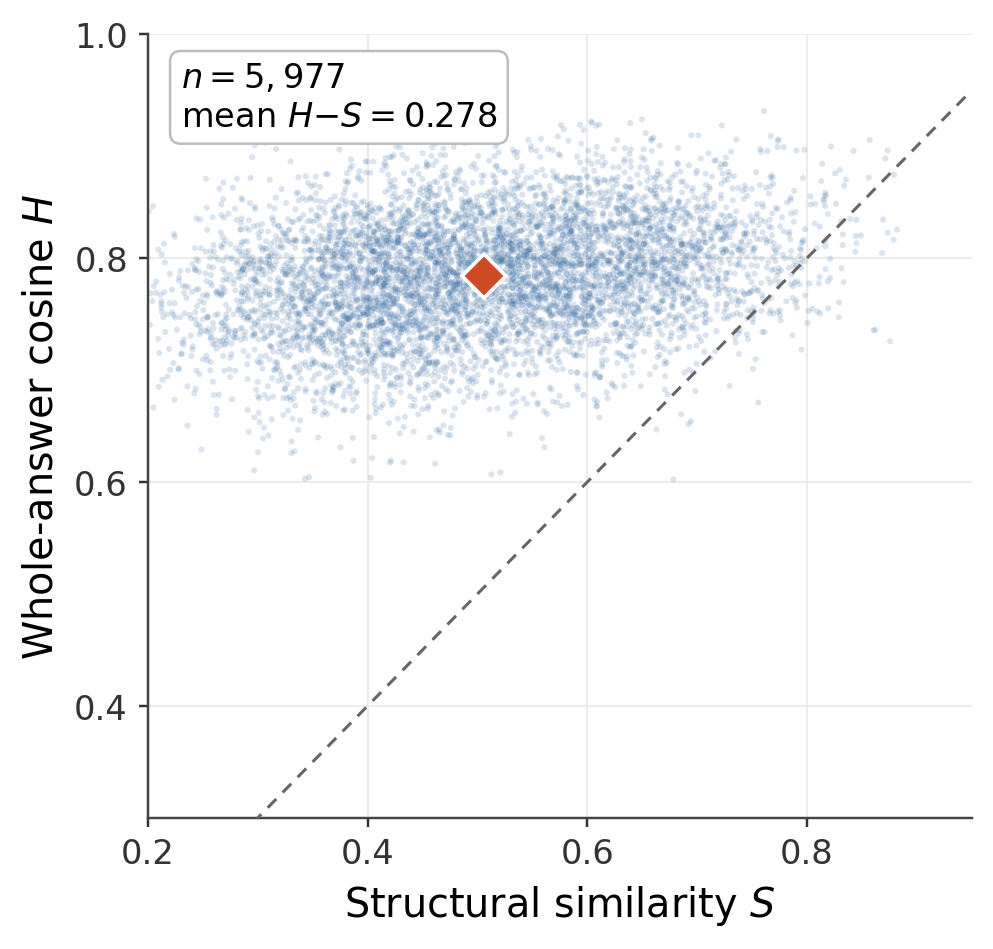}
    \subcaptiontext{\textsc{SciBench}, DeepSeek-V3.2}
  \end{subfigure}\hfill
  \begin{subfigure}[t]{0.32\textwidth}
    \centering
    \includegraphics[width=\linewidth]{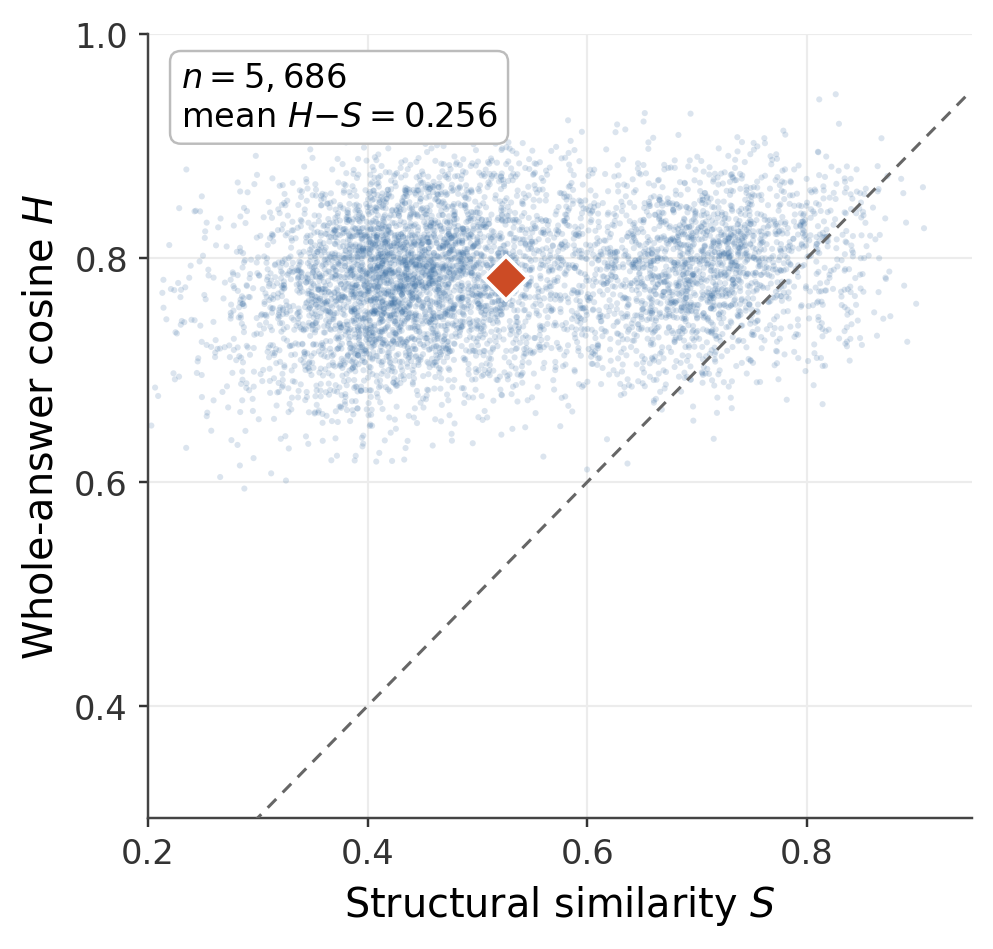}
    \subcaptiontext{\textsc{SciBench}, Qwen3-235B}
  \end{subfigure}
  \caption{Representative pair--question point clouds for whole-answer cosine $H$ against structural score $S$. These plots audit the rank-based result in Figure~\ref{fig:H_vs_S}: high whole-output similarity commonly coexists with substantially lower step-level structural alignment.}
  \label{fig:app_h_vs_s_panels}
\end{figure*}

Figure~\ref{fig:app_h_vs_s_panels} gives representative raw point clouds for the same diagnostic. The panels show that high whole-output similarity can coexist with much lower step-level structural alignment, motivating the rank-based summary in Table~\ref{tab:rank_h_s_comparison}.

\begin{table*}[!htbp]
\caption{Rank-based comparison of whole-answer cosine $H$ and structural score $S$. Percentile thresholds are computed within each model--dataset evaluation and language pair, so the table does not interpret raw $H-S$ as a calibrated distance. $\rho_{\mathrm{pair}}$ is the mean within-pair Spearman correlation. \emph{top-$H$ not top-$S$} is the share of instances in the top quartile of $H$ that are not also in the top quartile of $S$; \emph{top-$H$ bottom-$S$} is the stricter share that falls in the bottom structural quartile.}
\label{tab:rank_h_s_comparison}
\centering
\scriptsize
\resizebox{\textwidth}{!}{%
\begin{tabular}{llrrrrr}
\toprule
{\bf Dataset} & {\bf Model} & $P$ & {\bf pairs} & $\rho_{\mathrm{pair}}$ & {\bf top-$H$ not top-$S$} & {\bf top-$H$ bottom-$S$} \\
\midrule
\textsc{MMLU-ProX} & Claude Haiku 4.5 & 5,459 & 21 & 0.23 & 66.0\% & 13.8\% \\
\textsc{MMLU-ProX} & Claude Opus 4.6 & 8,470 & 21 & 0.29 & 59.1\% & 12.8\% \\
\textsc{MMLU-ProX} & DeepSeek-V3.2 & 7,179 & 21 & 0.30 & 60.6\% & 12.9\% \\
\textsc{MMLU-ProX} & GPT-OSS-120B & 6,552 & 21 & 0.42 & 53.9\% & 7.5\% \\
\textsc{MMLU-ProX} & Qwen3-235B & 6,656 & 21 & 0.30 & 60.6\% & 13.3\% \\
\textsc{MMLU-ProX} & Qwen3-32B & 4,714 & 21 & 0.30 & 59.9\% & 14.5\% \\
\textsc{SciBench} & Claude Haiku 4.5 & 4,739 & 21 & 0.24 & 63.3\% & 15.2\% \\
\textsc{SciBench} & Claude Opus 4.6 & 7,062 & 21 & 0.24 & 65.5\% & 15.0\% \\
\textsc{SciBench} & DeepSeek-V3.2 & 5,977 & 21 & 0.24 & 65.5\% & 14.7\% \\
\textsc{SciBench} & GPT-OSS-120B & 5,435 & 21 & 0.24 & 63.7\% & 16.2\% \\
\textsc{SciBench} & Qwen3-235B & 5,686 & 21 & 0.21 & 67.3\% & 16.4\% \\
\textsc{SciBench} & Qwen3-32B & 3,874 & 21 & 0.25 & 61.5\% & 17.1\% \\
\textsc{StrategyQA} & Claude Haiku 4.5 & 6,857 & 21 & 0.45 & 52.1\% & 7.8\% \\
\textsc{StrategyQA} & Claude Opus 4.6 & 9,039 & 21 & 0.39 & 55.5\% & 9.3\% \\
\textsc{StrategyQA} & DeepSeek-V3.2 & 7,813 & 21 & 0.36 & 56.8\% & 10.7\% \\
\textsc{StrategyQA} & GPT-OSS-120B & 6,751 & 21 & 0.29 & 60.2\% & 13.2\% \\
\textsc{StrategyQA} & Qwen3-235B & 7,591 & 21 & 0.39 & 55.8\% & 9.5\% \\
\textsc{StrategyQA} & Qwen3-32B & 5,177 & 21 & 0.38 & 57.1\% & 9.6\% \\
\bottomrule
\end{tabular}
}
\end{table*}

Across the evaluations in Table~\ref{tab:rank_h_s_comparison}, the mean within-pair Spearman correlation ranges from $0.21$ to $0.45$. Among examples in the top quartile of whole-answer cosine, a weighted average of 59.9\% are not also in the top structural quartile, and 12.4\% fall into the bottom structural quartile. Thus, high holistic similarity often fails to identify the examples that are most structurally aligned under the step-level metric, even when the comparison is made entirely in rank space.

\subsection{Parser Coverage}
\label{app:parser_coverage}

We measure the success rate of the regex-based step parser of \S\ref{sec:step_parsing} on the StrategyQA Claude Opus~4.6 traces from the post-judge verified question set ($|\text{verified}|{=}458$ questions, equal to $500$ minus the $42$ rejected-sample questions of \S\ref{app:rejected_case_studies} and seven greater than $Q{=}451$ in Table~\ref{tab:aggregate_results} because $Q$ additionally requires that at least two answer-correct languages survive the trace-judge per question). The parser is run on the trace produced for every language on each verified question regardless of whether that language is itself answer-correct, giving $458$ parser invocations per language; per-language answer accuracies on the underlying $500$-problem inference set are reported separately in Figure~\ref{fig:app_accuracy_summary}.

Coverage---the fraction of those $458$ traces yielding at least one parsed step---is $100.0\%$ for English, Spanish, Hindi, Russian, Vietnamese, and Chinese, and $99.8\%$ for Arabic, where exactly one trace produced no steps. No language required the generic numbered-list fallback in place of the language-specific \texttt{Step}~$n$: / \texttt{Paso}~$n$: / Chinese-character header: the fallback rate is $0.0\%$ in all seven languages. Parser failure is therefore not a material source of the divergence patterns reported in Section~\ref{sec:experiments}. Coverage records only whether a trace parses; Appendix~\ref{app:parser_validation} reports the marker-agreement diagnostic and the manual boundary spot check that address where the recovered boundaries fall.

\subsection{Step-Count-Matched Rank Statistic}
\label{app:granularity}

The many-to-one correspondences in $S$ tolerate a difference in step granularity, but they do not make $S$ blind to it (Appendix~\ref{sec:verification}: merging steps lowers $S$). If the headline rank statistic were driven by one language splitting the same reasoning into more steps than another, it would weaken on pairs whose traces have the same number of steps. Table~\ref{tab:granularity} recomputes it on exactly those pairs, with quartile thresholds recomputed within each language pair on the restricted subset, so that the restricted value is the same statistic on a smaller population.

\begin{table*}[!htbp]
\caption{Headline rank statistic (share of top-quartile-$H$ instances not in the top quartile of $S$, quartiles within each evaluation and language pair) on all pair--question instances and on those whose two traces have equal step counts. $\Delta$ is matched minus all, in percentage points, computed before rounding. Mean $S$ is pooled over pair--question instances and therefore differs slightly from the question-macro-averaged values of Table~\ref{tab:aggregate_results}.}
\label{tab:granularity}
\centering
\scriptsize
\resizebox{\textwidth}{!}{%
\begin{tabular}{llrrrrrrr}
\toprule
 & & & \multicolumn{2}{c}{\bf top-$H$ not top-$S$} & & & \multicolumn{2}{c}{\bf mean $S$} \\
\cmidrule(lr){4-5}\cmidrule(lr){8-9}
{\bf Dataset} & {\bf Model} & $P$ & {\bf all} & {\bf matched} & $\Delta$ & {\bf \% matched} & {\bf all} & {\bf matched} \\
\midrule
\textsc{MMLU-ProX} & Claude Haiku 4.5 & 5{,}459 & 66.0\% & 68.1\% & $+2.0$ & 37.9\% & 0.520 & 0.540 \\
\textsc{MMLU-ProX} & Claude Opus 4.6 & 8{,}470 & 59.1\% & 56.2\% & $-2.9$ & 38.0\% & 0.572 & 0.595 \\
\textsc{MMLU-ProX} & DeepSeek-V3.2 & 7{,}179 & 60.6\% & 58.0\% & $-2.6$ & 22.5\% & 0.542 & 0.565 \\
\textsc{MMLU-ProX} & GPT-OSS-120B & 6{,}552 & 53.9\% & 52.7\% & $-1.2$ & 41.0\% & 0.548 & 0.563 \\
\textsc{MMLU-ProX} & Qwen3-235B & 6{,}656 & 60.6\% & 61.5\% & $+0.9$ & 20.2\% & 0.544 & 0.566 \\
\textsc{MMLU-ProX} & Qwen3-32B & 4{,}714 & 59.9\% & 59.2\% & $-0.7$ & 37.3\% & 0.521 & 0.537 \\
\textsc{SciBench} & Claude Haiku 4.5 & 4{,}739 & 63.3\% & 65.7\% & $+2.4$ & 34.8\% & 0.576 & 0.591 \\
\textsc{SciBench} & Claude Opus 4.6 & 7{,}062 & 65.5\% & 62.8\% & $-2.8$ & 36.8\% & 0.611 & 0.621 \\
\textsc{SciBench} & DeepSeek-V3.2 & 5{,}977 & 65.5\% & 71.0\% & $+5.5$ & 16.6\% & 0.506 & 0.518 \\
\textsc{SciBench} & GPT-OSS-120B & 5{,}435 & 63.7\% & 61.9\% & $-1.9$ & 31.6\% & 0.591 & 0.600 \\
\textsc{SciBench} & Qwen3-235B & 5{,}686 & 67.3\% & 67.2\% & $-0.1$ & 20.3\% & 0.526 & 0.534 \\
\textsc{SciBench} & Qwen3-32B & 3{,}874 & 61.5\% & 57.8\% & $-3.7$ & 28.7\% & 0.631 & 0.634 \\
\textsc{StrategyQA} & Claude Haiku 4.5 & 6{,}857 & 52.1\% & 49.5\% & $-2.6$ & 56.8\% & 0.533 & 0.547 \\
\textsc{StrategyQA} & Claude Opus 4.6 & 9{,}039 & 55.5\% & 54.6\% & $-1.0$ & 51.9\% & 0.599 & 0.611 \\
\textsc{StrategyQA} & DeepSeek-V3.2 & 7{,}813 & 56.8\% & 57.8\% & $+1.0$ & 35.5\% & 0.563 & 0.569 \\
\textsc{StrategyQA} & GPT-OSS-120B & 6{,}751 & 60.2\% & 60.4\% & $+0.2$ & 47.2\% & 0.542 & 0.553 \\
\textsc{StrategyQA} & Qwen3-235B & 7{,}591 & 55.8\% & 55.9\% & $+0.1$ & 46.7\% & 0.562 & 0.576 \\
\textsc{StrategyQA} & Qwen3-32B & 5{,}177 & 57.1\% & 56.8\% & $-0.2$ & 46.3\% & 0.521 & 0.529 \\
\bottomrule
\end{tabular}%
}
\end{table*}

The ``all'' column reproduces Table~\ref{tab:rank_h_s_comparison}. Restricting to equal step counts lowers the statistic in 11 of the 18 evaluations and raises it in 7; the change averages $1.8$ points in magnitude, with a net shift of $-0.4$ points, and every matched value lies between $49.5\%$ and $71.0\%$. The largest change, $+5.5$ points on \textsc{SciBench} with DeepSeek-V3.2, falls on the evaluation where the fewest pairs have matched step counts ($16.6\%$). Mean $S$ rises slightly on matched pairs, by $0.003$--$0.023$, consistent with the sanity check that a granularity mismatch costs some structural similarity. Step-granularity mismatch therefore does not account for the dissociation. Part of what remains is arguably the construct rather than a confound: if a model systematically emits coarser steps in one language, that is a difference a reader of that trace experiences.

\subsection{Sanity-Check Validation}
\label{sec:verification}

We use controlled English--Chinese trace perturbations to verify that the metrics respond to the intended kinds of reasoning change. The source trace asks whether Aristotle could have used a laptop; the Chinese-side trace is either a direct translation or a controlled perturbation that reorders, merges, removes, abstracts, or replaces reasoning steps. Table~\ref{tab:verification_results} combines the metric outcomes with representative perturbation descriptions.

\begin{table*}[t]
\centering
\caption{Combined sanity-check results and illustrative perturbations for the en--zh validation set. $H$ is whole-answer cosine, $S$ is structural similarity, and $\tau$ is soft Kendall-$\tau$ order consistency.}
\label{tab:verification_results}
\small
\setlength{\tabcolsep}{4pt}
\begin{tabular}{@{}p{0.25\linewidth}ccc p{0.45\linewidth}@{}}
\toprule
\textbf{Case} & $H$ & $S$ & $\tau$ & \textbf{Chinese-side perturbation} \\
\midrule
base (aligned reasoning) & 0.851 & 0.827 & 0.960 & Same five reasoning moves in the same order, translated directly. \\
reversed order steps & 0.834 & 0.857 & $-$0.964 & Same five moves but in reversed order. \\
step merge (coarser steps) & 0.826 & 0.723 & 0.891 & The first two premises are merged, and feasibility plus conclusion are expressed in fewer coarser steps. \\
step skip (missing steps) & 0.848 & 0.746 & 0.884 & One intermediate historical comparison is omitted while the final answer is preserved. \\
abstraction (semantic drift) & 0.779 & 0.366 & 0.643 & The trace keeps the broad topic but replaces concrete date comparisons with higher-level historical claims. \\
granularity mismatch (uneven split) & 0.825 & 0.631 & 0.839 & The same content is split into uneven step units across the two traces. \\
low-entropy granularity & 0.811 & 0.433 & 0.000 & Multiple source steps collapse onto a single broad target step. \\
unrelated (semantic failure) & 0.346 & 0.035 & 0.495 & A different corporate-governance reasoning trace is substituted while preserving a final ``no'' answer string. \\
\bottomrule
\end{tabular}
\end{table*}

The sanity check confirms that the decomposition separates content coverage from order. Reversing the same reasoning keeps $H$ and $S$ high but drives $\tau$ negative, while abstraction and unrelated substitutions keep nonzero holistic similarity but sharply reduce $S$.

\subsection{Sharpening Ablation: Power-Law vs Softmax-Temperature}
\label{app:sharpening_ablation}

The soft assignment in \S\ref{sec:soft_tau} sharpens the calibrated outer step matrix $\mathbf{A}$ element-wise by a fixed power before row $L_1$-normalization (default exponent $p{=}3$). To check that this choice is not driving the reported $\tau$ values, we re-aggregate soft Kendall-$\tau$-with-tie on the same per-question step-similarity matrices stored in our \texttt{step4\_divergence} artifacts (so no re-encoding or re-inference is required), comparing two sharpening families: (i) power-law $\tilde{A}_{i,j} = \max(A_{i,j}, 0)^{p}$ for $p \in \{1,2,3,5,10\}$ and (ii) softmax-temperature $\tilde{A}_{i,j} \propto \exp(\max(A_{i,j}, 0)/T)$ for $T \in \{0.05, 0.1, 0.2, 0.5, 1.0\}$. Both families collapse to the row-stochastic soft-assignment matrix $P$ used by the soft Kendall-$\tau$ computation in \S\ref{sec:soft_tau}.

\begin{table}[H]
\caption{Sharpening ablation. All three columns are computed against the default power $p{=}3$ across the 18 (model, dataset) evaluations: Spearman and Pearson rank/linear correlation of $\bar\tau^{\text{with-tie}}$, and its mean absolute gap. Both families preserve the cross-evaluation ranking very closely; absolute scale shifts are expected because each scheme distributes probability mass differently.}
\label{tab:sharpening_ablation}
\centering
\small
\setlength{\tabcolsep}{5pt}
\begin{tabular}{@{}lccc@{}}
\toprule
{\bf scheme} & Spearman & Pearson & mean $|\Delta\bar\tau|$ \\
\midrule
power $p{=}1$    & $0.887$ & $0.857$ & $0.226$ \\
power $p{=}2$    & $0.994$ & $0.978$ & $0.098$ \\
power $p{=}3$    & $1.000$ & $1.000$ & $0.000$ \\
power $p{=}5$    & $0.977$ & $0.980$ & $0.122$ \\
power $p{=}10$   & $0.948$ & $0.933$ & $0.235$ \\
\midrule
softmax $T{=}0.05$ & $0.926$ & $0.898$ & $0.220$ \\
softmax $T{=}0.1$  & $0.940$ & $0.928$ & $0.076$ \\
softmax $T{=}0.2$  & $0.920$ & $0.916$ & $0.117$ \\
softmax $T{=}0.5$  & $0.897$ & $0.875$ & $0.281$ \\
softmax $T{=}1.0$  & $0.855$ & $0.849$ & $0.334$ \\
\bottomrule
\end{tabular}
\end{table}

Across all sharpening choices in both families, the cross-evaluation ranking of (model, dataset) pairs by mean $\bar\tau^{\text{with-tie}}$ is preserved with Spearman $\geq 0.85$ (and $\geq 0.92$ for every alternative within mean-absolute-tau distance $\leq 0.12$ of $p{=}3$), and the closest softmax setting ($T{=}0.1$) matches the absolute scale of $p{=}3$ to within mean $|\Delta\bar\tau| = 0.076$. Within the power family, $p{=}2$ and $p{=}5$ have Spearman $\geq 0.98$ relative to $p{=}3$. The qualitative conclusions in \S\ref{sec:exp_main} and Table~\ref{tab:aggregate_results}---which depend on the relative ordering of evaluations rather than absolute $\bar\tau$ values---are therefore robust to the sharpening scheme. We retain power-law $p{=}3$ as the reported default because (a) it is temperature-free, (b) it has no near-zero divisor for empty rows, and (c) it produces $P$ that is well behaved for the small step counts ($t \le 15$) typical in our verified traces.

\appendixsectionbreak
\section{Qualitative Examples and Case Studies}
\label{app:case_studies}

\subsection{Qualitative Trace Examples}
\label{app:qualitative}

We include qualitative examples showing what numerical gaps correspond to in actual traces.
\paragraph{Verified \textsc{StrategyQA} example.}
On \textsc{StrategyQA} question (``Is it safe to wear sandals in snow?''), English and Spanish agree and reach high whole-trace similarity($H{=}0.875$), but step alignment is weak($S{=}0.338$, $\tau{=}0.009$).

\begin{table}[t]
\caption{\textsc{StrategyQA} trace excerpt for question \texttt{f15b0aaa611d1d554186}. The traces reach the same answer, but decompose reasoning differently.}
\label{tab:qual_strategyqa_trace}
\centering
\scriptsize
\setlength{\tabcolsep}{3pt}
\begin{tabular}{@{}p{0.47\linewidth}p{0.47\linewidth}@{}}
\toprule
\textbf{English trace} & \textbf{Spanish trace} \\
\midrule
\textbf{Step 1.} Consider the physical properties of sandals. Sandals are open-toed, minimally insulated footwear that leave significant portions of the feet exposed to the environment.
&
\textbf{Step 1.} Analicemos la pregunta desde el punto de vista de la seguridad y la salud. Usar sandalias en la nieve implica exponer los pies a temperaturas muy bajas, humedad y superficies resbaladizas.
\\[3pt]
\textbf{Step 2.} Consider the conditions associated with snow. Snow implies temperatures at or below 0$^\circ$C, along with wet conditions from snow and slush.
&
\textbf{Step 2.} Riesgos principales de usar sandalias en la nieve: hipotermia localizada y congelaci\'on; resbalones y ca\'idas; entumecimiento; enfermedades.
\\[3pt]
\textbf{Step 3.} Evaluate the risks. Exposed feet in snow can rapidly lose heat, leading to frostbite; sandals also lack traction suitable for icy or snowy surfaces.
&
\textbf{Step 3.} Comparaci\'on con el calzado adecuado: las botas de invierno a\'islan, mantienen los pies secos y proporcionan tracci\'on; las sandalias carecen de estas caracter\'isticas.
\\[3pt]
\textbf{Step 4.} Consider any mitigating factors. Thick socks with sandals might marginally delay heat loss but do not adequately protect against the hazards.
&
\textbf{Step 4.} Conclusi\'on: no, no es seguro usar sandalias en la nieve. Se recomienda calzado cerrado, impermeable y con buena tracci\'on.
\\[3pt]
\textbf{Step 5.} Wearing sandals in snow is not safe due to frostbite, hypothermia of the feet, loss of sensation, and increased likelihood of slipping.
&
\\
\bottomrule
\end{tabular}
\end{table}

\paragraph{Verified \textsc{SciBench} example.}
On \textsc{SciBench} question \texttt{69}, the model must recover the original pressure of an ideal gas after isothermal compression reduces its volume from $6.85$ to $4.65\,\mathrm{dm^3}$, with final pressure $5.04$ bar. The Hindi and Russian traces are both verified correct and reach $P_1\approx3.38$ atm. They preserve the same high-level order ($\tau{=}0.832$) but have low structural similarity ($S{=}0.269$), because one trace folds some setup into formula-bearing steps while the other separates data listing, volume recovery, Boyle's law, unit conversion, and rounding.

\begin{table}[H]
\caption{Concrete \textsc{SciBench} verified trace excerpt for question \texttt{69}. Non-English traces are shown as English translations of the Hindi and Russian reasoning steps.}
\label{tab:qual_scibench_trace}
\centering
\scriptsize
\setlength{\tabcolsep}{3pt}
\begin{tabular}{@{}p{0.47\linewidth}p{0.47\linewidth}@{}}
\toprule
\textbf{Hindi trace (translated)} & \textbf{Russian trace (translated)} \\
\midrule
\textbf{Step 1.} Find the initial volume: since compression reduces volume, $V_1=V_2+2.20=4.65+2.20=6.85\,\mathrm{dm^3}$.
&
\textbf{Step 1.} List the given data: $\Delta V=2.20\,\mathrm{dm^3}$, $P_2=5.04$ bar, $V_2=4.65\,\mathrm{dm^3}$, and the process is isothermal.
\\[3pt]
\textbf{Step 2.} Apply Boyle's law for an isothermal ideal-gas process: $P_1V_1=P_2V_2$.
&
\textbf{Step 2.} Recover the initial volume: $V_1=V_2+\Delta V=4.65+2.20=6.85\,\mathrm{dm^3}$.
\\[3pt]
\textbf{Step 3.} Compute $P_1$ in bar: $P_1=(5.04\times4.65)/6.85=3.4213$ bar.
&
\textbf{Step 3.} Apply Boyle's law and compute in bar: $P_1=P_2V_2/V_1=(5.04\times4.65)/6.85=3.4212$ bar.
\\[3pt]
\textbf{Step 4.} Convert bar to atm: $P_1=3.4213/1.01325=3.377$ atm.
&
\textbf{Step 4.} Convert pressure units: $1\,\mathrm{atm}=1.01325$ bar, so $P_1=3.4212/1.01325=3.376$ atm.
\\[3pt]
\textbf{Step 5.} Express with appropriate significant figures: $P_1\approx3.38$ atm.
&
\textbf{Step 5.} Round to three significant figures: $P_1\approx3.38$ atm.
\\
\bottomrule
\end{tabular}
\end{table}

\paragraph{Verified \textsc{MMLU-ProX} example.}
In a retained \textsc{MMLU-ProX} seed-dormancy case, English and Chinese both select the correct option \texttt{J}. Whole-output similarity remains high ($H{=}0.857$), but structural similarity is much lower ($S{=}0.384$), because English evaluates the answer choices almost immediately while Chinese first states the biological definition and the environmental-cue mechanism before excluding alternatives.

\begin{table}[H]
\caption{Concrete \textsc{MMLU-ProX} verified trace excerpt for question \texttt{222}. The Chinese trace is shown in English translation.}
\label{tab:qual_mmlu_verified_trace}
\centering
\scriptsize
\setlength{\tabcolsep}{3pt}
\begin{tabular}{@{}p{0.47\linewidth}p{0.47\linewidth}@{}}
\toprule
\textbf{English trace} & \textbf{Chinese trace (translated)} \\
\midrule
\textbf{Step 1.} Define dormancy as suspended growth and metabolic activity that lets seeds survive unfavorable conditions until germination is suitable.
&
\textbf{Step 1.} Define dormancy as a special physiological state in which a seed may not germinate even when water, temperature, and oxygen are externally suitable.
\\[3pt]
\textbf{Step 2.} Evaluate the options: reject answers that describe flavor, spontaneous germination, active growth, or a single cue such as light or water.
&
\textbf{Step 2.} Explain dormancy termination: it typically depends on combinations of temperature, light, water, oxygen, scarification, and species-specific cues.
\\[3pt]
\textbf{Step 3.} Continue eliminating options A--I because each is either harmful, too narrow, or contrary to seed biology.
&
\textbf{Step 3.} Exclude A--I one by one: each either misunderstands dormancy or treats germination as controlled by a single oversimplified condition.
\\[3pt]
\textbf{Step 4.} Select option \texttt{J} as the comprehensive answer: dormancy is a resting state and germination requires precise environmental combinations.
&
\textbf{Step 4.} Select option \texttt{J}: it captures both the survival role of dormancy and the need for precise environmental signals.
\\
\bottomrule
\end{tabular}
\end{table}

\subsection{Rejected-Sample Case Studies}
\label{app:rejected_case_studies}

The main trace-divergence results use only language pairs for which both answers are verified correct. As a qualitative diagnostic, we also evaluate examples rejected from the primary verified set with the same trace-divergence metrics. This analysis is not a primary metric, because rejected examples can conflate answer error, answer inconsistency, and reasoning divergence. It is nevertheless useful for inspecting whether failures are shared across languages or arise from language-specific interpretations.

Compared with the corresponding verified traces, rejected traces have lower structural similarity and order consistency, while whole-output similarity remains high. For Claude Opus~4.6 on \textsc{StrategyQA} ($42$ rejected questions), the rejected set has $S{=}0.563$ and $\tau{=}0.341$, compared with $S{=}0.598$ and $\tau{=}0.418$ on the verified set. On \textsc{MMLU-ProX} ($65$ rejected), the rejected set similarly drops from $S{=}0.569$ to $S{=}0.525$ and from $\tau{=}0.414$ to $\tau{=}0.354$. For DeepSeek-V3.2 on \textsc{StrategyQA} ($69$ rejected), rejected examples have $S{=}0.537$ and $\tau{=}0.270$, compared with verified values of $S{=}0.560$ and $\tau{=}0.324$. In all three cases $H$ remains close to $0.80$ ($0.810$, $0.807$, and $0.803$ respectively), so holistic similarity alone does not distinguish verified reasoning agreement from rejected-sample failure modes.

\paragraph{Shared wrong reasoning.}
On the \textsc{StrategyQA} question ``Is the Foreign and Commonwealth Office a European political agency?'', the dataset answer is yes, but Claude Opus~4.6 answers no in all seven languages. The traces are highly similar ($S{=}0.756$, $H{=}0.887$): each language interprets ``European political agency'' as an EU or pan-European institution and rejects the label because the Foreign and Commonwealth Office is a UK government department. This case shows that high cross-lingual trace similarity can reflect a shared error rather than reliable reasoning. DeepSeek-V3.2 exhibits the same final-answer failure on this question, with all seven languages answering no and $S{=}0.700$.

\begin{table}[H]
\caption{Rejected \textsc{StrategyQA} trace excerpt for the Foreign and Commonwealth Office question. Both traces share the same wrong institutional interpretation.}
\label{tab:rejected_strategyqa_shared_trace}
\centering
\small
\begin{tabular}{p{0.47\linewidth}p{0.47\linewidth}}
\toprule
\textbf{English trace} & \textbf{Spanish trace} \\
\midrule
\textbf{Step 1.} Identify the FCO as a department of the UK government responsible for British foreign affairs.
&
\textbf{Step 1.} Identifica la FCO como un ministerio del gobierno del Reino Unido responsable de la pol\'itica exterior y la Commonwealth.
\\[3pt]
\textbf{Step 2.} Define a European political agency as an EU or pan-European institution such as the European Commission or Europol.
&
\textbf{Step 2.} Determina que una agencia pol\'itica europea ser\'ia una agencia de la Uni\'on Europea o de una organizaci\'on paneuropea.
\\[3pt]
\textbf{Step 3.} Compare the FCO against that definition and conclude it is a national British institution, not an EU or pan-European agency.
&
\textbf{Step 3.} Aunque el Reino Unido est\'a geogr\'aficamente en Europa, la FCO es un ministerio nacional brit\'anico, no una instituci\'on supranacional europea.
\\[3pt]
\textbf{Final.} No.
&
\textbf{Final.} No.
\\
\bottomrule
\end{tabular}
\end{table}

\paragraph{Divergent rejected reasoning.}
On the \textsc{StrategyQA} question ``Can a lifeboat rescue people in the Hooke Sea?'', the dataset answer is no. DeepSeek-V3.2 splits across languages: Arabic, English, and Vietnamese answer yes, while Spanish, Hindi, Russian, and Chinese answer no. The structural score is low ($S{=}0.449$, $\tau{=}0.221$), with particularly weak pairs such as English--Hindi ($S{=}0.268$) and English--Russian ($S{=}0.333$). The yes-answer traces treat Hooke Sea as a hypothetical ordinary sea; the no-answer traces notice that the named sea is fictional or wordplay. This case illustrates a different rejected-sample mode: the model does not merely share one wrong belief across languages, but changes the interpretation of the premise.

\begin{table}[H]
\caption{Rejected \textsc{StrategyQA} trace excerpt for the Hooke Sea question. The English trace answers yes by assuming an ordinary sea, while the Hindi trace answers no by treating the location as fictional. The Hindi trace is shown in English translation.}
\label{tab:rejected_strategyqa_divergent_trace}
\centering
\small
\begin{tabular}{p{0.47\linewidth}p{0.47\linewidth}}
\toprule
\textbf{English trace} & \textbf{Hindi trace (translated)} \\
\midrule
\textbf{Step 1.} Treat ``Hooke Sea'' as a fictional or hypothetical sea named after Robert Hooke.
&
\textbf{Step 1.} Treat Hooke Sea as a fictional sea related to Hooke's law.
\\[3pt]
\textbf{Step 2.} In the absence of special constraints, assume it is a body of water similar to ordinary seas.
&
\textbf{Step 2.} Since the sea is fictional, there are no real people there to rescue.
\\[3pt]
\textbf{Step 3.} Lifeboats are designed to rescue people in seas or oceans if they can navigate the conditions.
&
\textbf{Step 3.} A lifeboat is for rescuing real people, but a fictional sea contains no real rescue target.
\\[3pt]
\textbf{Final.} Yes.
&
\textbf{Final.} No.
\\
\bottomrule
\end{tabular}
\end{table}

\paragraph{Rejected \textsc{MMLU-ProX} answer split.}
On \textsc{MMLU-ProX} question \texttt{141}, Claude Opus~4.6 is asked to compare annual premiums for a $\$15{,}000$ insurance policy under a 20-payment life policy and an ordinary life policy paid up at age 65. The gold option is B, but none of the inspected English, Spanish, and Chinese traces is correct: English selects A, Spanish selects D, and Chinese selects I. The failure is not a single shared wrong answer. Each trace follows the same broad tabular-rate procedure--choose premium rates per $\$1{,}000$, multiply by 15, and match the nearest option--but the languages assume different rate-table values, causing different numerical differences and therefore different final options. This case adds a multiple-choice rejected-sample mode to the StrategyQA examples: translation preserves the task format, yet the model's unstated lookup assumptions drift across languages.

\begin{table}[H]
\caption{Rejected \textsc{MMLU-ProX} trace excerpt for question \texttt{141}. The Chinese trace is shown in English translation.}
\label{tab:rejected_mmlu_answer_split_trace}
\centering
\small
\begin{tabular}{p{0.30\linewidth}p{0.30\linewidth}p{0.30\linewidth}}
\toprule
\textbf{English trace} & \textbf{Spanish trace} & \textbf{Chinese trace (translated)} \\
\midrule
\textbf{Rate assumption.} Uses 20-payment life $\$26.47$ and ordinary life $\$24.44$ per $\$1{,}000$, then tries nearby textbook rates.
&
\textbf{Rate assumption.} Starts with $\$25.04$ and $\$22.86$ per $\$1{,}000$, then tries alternate rates.
&
\textbf{Rate assumption.} Starts with $\$26.40$ and $\$24.26$, then switches to a much wider ``common textbook'' rate gap.
\\[3pt]
\textbf{Calculation.} Computes differences around $\$30.45$--$\$30.60$ and maps this to option \texttt{A}.
&
\textbf{Calculation.} Computes $\$32.70$, $\$32.10$, and then a $\$32.55$ candidate that maps to option \texttt{D}.
&
\textbf{Calculation.} Reverse-engineers option \texttt{I}: $\$31.55/15=\$2.1033$ per $\$1{,}000$, close to a $\$2.10$ rate gap.
\\[3pt]
\textbf{Final.} Selects \texttt{A}.
&
\textbf{Final.} Selects \texttt{D}.
&
\textbf{Final.} Selects \texttt{I}.
\\
\bottomrule
\end{tabular}
\end{table}

\appendixsectionbreak
\section{Human Validation of the Trace-Divergence Metrics}
\label{app:human_validation}

This appendix gives the protocol, sampling procedure, and full statistics behind \S\ref{sec:validation}.

\subsection{Protocol}
\label{app:hv_protocol}

\noindent{\bf Task.} Each item presents an English source question, the final answer (both languages reached the same verified-correct answer), and two reasoning traces labeled \emph{Trace A} and \emph{Trace B}. Annotators judge whether the traces reach the answer by the same reasoning. Three fields are recorded per item:

\begin{itemize}\itemsep2pt
\item \texttt{equivalence}, the primary variable: \emph{Same} / \emph{Partially different} / \emph{Substantially different}, coded ordinally as $2/1/0$.
\item \texttt{diff\_*}, a multi-label difference taxonomy: wording, granularity, missing content, different strategy, different order.
\item \texttt{material}, binary: would the difference change what a reviewer concludes when using the trace to verify the answer? A binary item yields a more stable per-stratum rate at this sample size than a Likert scale.
\end{itemize}

\noindent{\bf Blinding.} Annotators never see $S$, $\tau$, $H$, or any stratum assignment; scores live only in a hidden key. The assignment of English and non-English traces to \emph{Trace A} and \emph{Trace B} is randomized per item to remove positional bias. Every item is labeled independently by at least two annotators, with a shared calibration subset labeled by all four. Annotators do not confer during labeling.

\noindent{\bf Annotators.} Four bilingual annotators per language pair, disjoint across the two studies. \S\ref{app:hv_anchor} reports a shared-anchor control establishing that the two groups do not differ in labeling threshold.

\subsection{Sampling}
\label{app:hv_sampling}

Both studies draw from Claude Opus~4.6 outputs pooled over \textsc{MMLU-ProX}, \textsc{SciBench}, and \textsc{StrategyQA}, restricted to pairs where both languages are answer-correct and both traces parse into at least two steps, with duplicate question text removed. The EN--ES eligible pool contains $n=1{,}178$ pairs.

Items are sampled \emph{crossed} on $H$ and $S$: both metrics are split at pool-relative terciles and items are drawn from every cell of the resulting $3\times3$ grid, with a round-robin over datasets inside each cell. This matters for interpretation. Stratifying on $S$ alone while holding $H$ high would suppress $H$'s variance by construction, so a null correlation for $H$ would be partly guaranteed by the design rather than measured. Sampling crossed avoids that: both metrics span their natural ranges, and any asymmetry between them is a property of the metrics rather than of the sample.

Table~\ref{tab:hv_range} confirms the design achieved this. Sample standard deviations closely match the pool for both metrics in both studies, and the sample correlation between $H$ and $S$ reproduces the pool value ($0.25$ against a pool $0.254$ for EN--ES), so the design induces no dependence between the two axes.

\begin{table}[h]
\caption{Range-restriction check. A ratio near $1$ indicates the metric was sampled over its natural spread, so its correlation is not attenuated by the design.}
\label{tab:hv_range}
\centering
\small
\begin{tabular}{llrrr}
\toprule
{\bf Study} & {\bf Metric} & {\bf sample sd} & {\bf pool sd} & {\bf ratio} \\
\midrule
EN--ZH & $S$ & 0.103 & 0.102 & 1.02 \\
EN--ZH & $H$ & 0.048 & 0.053 & 0.90 \\
EN--ES & $S$ & 0.109 & 0.105 & 1.04 \\
EN--ES & $H$ & 0.042 & 0.047 & 0.89 \\
\bottomrule
\end{tabular}
\end{table}

\subsection{Correlational Results}
\label{app:hv_correlations}

Table~\ref{tab:hv_correlations} in \S\ref{sec:validation} reports the correlations themselves. This section gives the tests behind them.

\noindent{\bf Testing the difference, not two separate nulls.} Reporting that $S$ correlates while $H$ does not is two independent tests, and invites the objection that the null on $H$ merely reflects low power. The claim is a difference between dependent correlations sharing a criterion variable, so we test it as one. Williams' test on $\rho(\text{equiv},S)$ against $\rho(\text{equiv},H)$ gives $\Delta\rho = 0.41$, bootstrap $95\%$ CI $[0.12, 0.68]$, $t(57) = 2.81$, $p = 0.007$ for EN--ES, and $\Delta\rho = 0.437$, $t(57) = 2.80$, $p = 0.0069$ for EN--ZH. The dissociation is significant in both pairs at effectively the same strength.

\noindent{\bf Parametric form.} An ordinal logistic regression of equivalence on standardized $S$ and $H$ (EN--ES) gives $\beta_{z(S)} = 0.86$ (SE $0.25$, OR $2.36$, $p = 0.0006$) and $\beta_{z(H)} = -0.09$ (SE $0.24$, OR $0.91$, $p = 0.71$). Likelihood-ratio tests dropping each term give $\chi^2(1) = 13.1$, $p = 0.0003$ for $S$ and $\chi^2(1) = 0.14$, $p = 0.71$ for $H$. A one-standard-deviation drop in $S$ multiplies the odds of a lower human equivalence rating by roughly $2.4$.

\noindent{\bf Generalization across language pairs.} Pooling both studies gives $\rho(\text{equiv},S) = 0.40$, $p < 0.001$, $N = 120$. A Fisher-$z$ homogeneity test between pairs gives $z = 0.46$, $p = 0.65$: the two language pairs do not differ. English--Spanish is the more demanding test of the two, since Spanish is typologically closer to English and its traces should diverge less.

\subsection{Comparison with an LLM-as-Judge Baseline}
\label{app:hv_llm_judge}

A natural alternative to a structural metric is to ask a strong model directly whether two traces reason the same way. We ran that baseline on the $60$ English--Chinese pairs of the initial annotation round, giving the judge (GPT-5.6) the identical rubric our annotators used, both traces, and no similarity scores.

The structural metric predicts human-judged equivalence more closely than the judge does. On these items $\rho(\text{equiv},S) = 0.419$ ($p = 0.004$) and $\rho(\text{equiv},\tau) = 0.413$ ($p = 0.005$), while the judge reaches $\rho = 0.235$ ($p = 0.12$, n.s.) and the holistic baseline $\rho = 0.129$ (n.s.). The judge also agrees with the human consensus less than the annotators agree with one another (pairwise agreement $0.576$ against $0.633$), and entering it as a fifth coder halves ordinal Krippendorff's $\alpha$, from $0.462$ to $0.237$: relative to the consensus it behaves as an outlier rather than as a substitute for it. Its errors are systematic rather than noisy---exact agreement with the human majority is $0.422$ but within-one agreement is $0.956$, so it is nearly always adjacent on the ordinal scale.

The failure mode is the one the framework targets. The judge never applied the \emph{Substantially different} label at any point, and its mean rating does not vary monotonically across $S$ strata. On high-$H$ / low-$S$ pairs it rates as \emph{Same} or \emph{Partially different} items that annotators unanimously call substantially different---for instance a multiple-choice item answered correctly in both languages where one trace eliminates every option exhaustively and the other identifies the answer directly ($S = 0.46$, $H = 0.80$; human \emph{Substantially different}, judge \emph{Same}).

We state the strength of this comparison precisely: the confidence intervals for $S$ and for the judge overlap, so the defensible claim is that $S$ is at least as faithful to human judgment as the judge and, unlike the judge, statistically significant on this sample. Beyond faithfulness, $S$ is deterministic, localizes divergence to specific steps, and separates content from order at no per-call cost.

\subsection{Supporting Analyses}
\label{app:hv_supporting}

\noindent{\bf Monotonicity.} Mean human equivalence rises monotonically across $S$ terciles and is flat across $H$ terciles. For EN--ES the means are $0.92 / 1.28 / 1.65$ by $S$ (Jonckheere--Terpstra $p < 0.001$) against $1.31 / 1.24 / 1.30$ by $H$ ($p = 0.72$). EN--ZH shows the same pattern: $1.069 / 1.324 / 1.420$ by $S$, and $1.433 / 1.162 / 1.300$ by $H$.

\noindent{\bf Detection framing.} Treating a majority \emph{Substantially different} label as the positive class, the area under the ROC curve is $0.81$ $[0.68, 0.92]$ for $S$, $0.76$ $[0.61, 0.88]$ for $\tau$, and $0.54$ $[0.36, 0.71]$ for $H$; a DeLong test of $S$ against $H$ gives $p = 0.012$. We present this as further evidence for the dissociation rather than as a deployable screening threshold: the positive class holds $11$ items in EN--ES, and the interval is correspondingly wide.

\noindent{\bf Difference types.} In EN--ES the \texttt{diff\_wording} rate is flat across $S$ terciles ($0.21 / 0.19 / 0.17$) while the content tags (\texttt{diff\_missing}, \texttt{diff\_granularity}, \texttt{diff\_order}) fall by a factor of three to four from low to high $S$. In EN--ZH the low-$S$ stratum is dominated by missing reasoning content ($0.477$), granularity ($0.295$), and order ($0.205$), with wording both the smallest category ($0.114$) and the only one declining monotonically across $S$ ($0.114 / 0.071 / 0.034$). A metric tracking surface verbalization would produce the opposite profile in both studies.

\noindent{\bf Agreement and attenuation.} Ordinal Krippendorff's $\alpha$ is $0.411$ for EN--ZH and $0.58$ for EN--ES. We use Krippendorff's $\alpha$ as the headline agreement statistic because it is the only standard option that simultaneously accommodates a small sample, a variable number of coders per item, and ordinal data; Cohen's $\kappa$ requires exactly two fixed raters and Fleiss' $\kappa$ a fixed rater count and nominal data. Because the human rating is itself a noisy instrument, the correlations above are attenuated rather than inflated by this unreliability. Correcting with the composite reliability of each study's rating mean gives $0.44 \rightarrow 0.49$ for EN--ES (composite reliability $0.81$) and $0.371 \rightarrow 0.47$ for EN--ZH ($0.61$). The two pairs agree more closely once measurement error is accounted for. We report the observed values as the headline throughout.

\noindent{\bf Annotator-group control.}
\label{app:hv_anchor}
Because the two studies use disjoint annotator groups, a difference between language pairs could in principle reflect the annotators rather than the languages. On $12$ shared English--English anchor items labeled by both groups, mean equivalence is $1.36$ for the EN--ZH group and $1.31$ for the EN--ES group (difference $0.05$, $[-0.19, 0.29]$), \emph{Same} rates are $34\%$ and $36\%$ (Mann--Whitney $p = 0.68$), and $\alpha$ on the anchors is $0.61$ and $0.59$. The groups are indistinguishable on shared items, so the EN--ES result reflects the language pair rather than a group-specific labeling threshold.

\noindent{\bf Orientation control.} Comparing items where the non-English trace appeared as \emph{Trace A} against those where it appeared as \emph{Trace B} gives mean equivalence $1.29$ against $1.28$ (Mann--Whitney $p = 0.88$): no presentation-order effect.

\noindent{\bf Outcome variance.} Overall mean equivalence is $1.29$ (sd $0.474$) for EN--ZH and $1.28$ (sd $0.47$) for EN--ES, so the difference between studies is not an artifact of differing outcome spread.

\subsection{Annotation Guide}
\label{app:hv_guide}

We reproduce the instructions given to annotators. The guide opens by stating that no similarity score of any kind will be shown and that only the reasoning is to be judged.

\noindent{\bf What is being labeled.} Each item gives an English source question, the final answer (both languages reached the same verified-correct answer), and two reasoning traces labeled \emph{Trace A} and \emph{Trace B}, one English and one non-English, with the side assignment randomized per item. Annotators read both and decide whether the traces reach the answer \emph{by the same reasoning}, recording three fields per item with fixed allowed values.

\noindent{\bf Field 1: \texttt{equivalence}} (the primary label, exactly one value).
\begin{itemize}\itemsep2pt
\item \emph{Same} --- the two traces use the same reasoning: same key claims, same logical path. Pure wording or translation differences still count as \emph{Same}.
\item \emph{Partially different} --- largely the same reasoning, but with a real difference in one part: a step present on one side is missing or merged on the other, or one sub-argument differs, while the overall path is recognizably the same.
\item \emph{Substantially different} --- the traces reason differently: different strategy, different decomposition, different premises or evidence, or different logical order of the core steps. A reader could not treat one as a faithful stand-in for the other.
\end{itemize}

\noindent{\bf Field 2: \texttt{diff\_*}} (difference type; mark each that applies, skipped entirely when \texttt{equivalence} is \emph{Same}). \texttt{diff\_wording}: only phrasing or translation differs. \texttt{diff\_granularity}: same content, split into more, fewer, or uneven steps. \texttt{diff\_missing}: a reasoning step or claim present on one side is absent on the other. \texttt{diff\_strategy}: a genuinely different solution approach or argument. \texttt{diff\_order}: the same core steps occur in a different order. \texttt{diff\_other}: with a free-text description.

\noindent{\bf Field 3: \texttt{material}} (binary). Would the difference change what a human reviewer concludes when using the trace to verify the answer or understand how it was reached? If the traces are the same or differ only cosmetically, \emph{No}. If a reviewer auditing one trace would reach a materially different understanding of \emph{why} the answer holds than from the other, \emph{Yes}.

\noindent{\bf Rules of thumb.} The guide instructs annotators to translate mentally and compare logic rather than surface tokens, noting that two traces in different languages will never be word-identical and that this alone is not a difference; to disregard correctness, since both answers are already verified correct, and to judge only whether the reasoning is the same rather than whether one trace is better; to pick the more different label when torn between two and add a note; and not to discuss items with other annotators while labeling, so that judgments remain independent.

\noindent{\bf Language check.} Annotators are asked to flag any trace written in the wrong language---one that should be in Spanish but is actually in English, or the reverse---with a fixed token in the notes field, then label the item as normal. This is requested particularly of the EN--ES annotators, because English and Spanish share the Latin alphabet and a reversion that an automatic script-level check cannot see is immediately obvious to a bilingual reader.

\noindent{\bf Calibration.} All annotators for a study first label a shared calibration subset. After those are complete the group meets briefly to compare judgments and settle the wording of edge cases, and the remaining items are then labeled independently. Annotators returning for the EN--ZH top-up batch were instructed that the labels and their definitions were unchanged and to apply exactly the standard they had used previously, without re-reading their earlier files to match specific judgments, since the two batches are pooled in analysis.

\subsection{Annotator Recruitment and Consent}
\label{app:hv_annotators}

Annotators were unpaid bilingual volunteers from the authors' academic and professional networks, based in Asia and North America, all adults with professional or native proficiency in both languages of the pair they annotated. Participation was voluntary and could be stopped at any time; annotators were told the purpose of the study and how their labels would be used, and consented on that basis. The burden was under thirty minutes per annotator for English--Chinese and under ninety minutes for English--Spanish. No personal data was collected, and none is retained in any released artifact. We did not seek ethics-board review: the study collects no personal or sensitive data, involves no deception or intervention, and asks adult volunteers only to compare model-generated text derived from public benchmarks---a minimal-risk protocol under our institutional framework.

\appendixsectionbreak
\section{Encoder Sensitivity}
\label{app:encoder_sensitivity}

\noindent{\bf Methodological concern.} $S$ and $\tau$ are computed from sentence embeddings, so a natural question is how much of the reported $H{-}S$ contrast is a property of reasoning traces and how much is a property of the encoders. Two distinct worries sit inside this. First, $H$ and $S$ are produced by different encoders in the main experiments (BGE-M3 and LaBSE respectively), so the gap between them could be an artifact of that asymmetry. Second, the framework might require a strong encoder to work at all, which would limit its usefulness.

\noindent{\bf Setup.} We recompute $S$ and $\tau$ from scratch under two alternative step encoders, on three Claude Opus~4.6 evaluations at full retained size (\textsc{StrategyQA} $n=451$, \textsc{MMLU-ProX} $n=431$, \textsc{SciBench} $n=362$), across all $21$ language pairs. The per-pair floor and ceiling are re-derived for each encoder rather than reused, since calibration constants estimated for one embedding space do not transfer. The alternatives are chosen to probe the two worries separately:

\begin{itemize}\itemsep2pt
\item \textbf{BGE-M3} ($568$M parameters, $1024$ dimensions, retrieval-contrastive objective) is the encoder $H$ already uses, so running $S$ on it removes the $H$--$S$ encoder asymmetry entirely. This is the only sound direction for that swap: placing LaBSE on $H$ would truncate whole traces at its $256$-token limit, whereas individual steps fit comfortably in both encoders.
\item \textbf{\texttt{paraphrase-multilingual-MiniLM-L12-v2}} ($118$M parameters, $384$ dimensions, paraphrase-distillation objective) is a four- to five-fold reduction in scale from LaBSE ($471$M) and a third training objective. Its $128$-token limit is not binding here, because the encoding unit is sentences within steps rather than whole steps: median sentence length is $56$ characters in English and $28$ in Chinese, with $99$th percentiles of $237$ and $124$.
\end{itemize}

\noindent{\bf Validation of the reimplementation.} Before comparing encoders we re-ran the identical pipeline with LaBSE, the encoder used in the main experiments, and recovered the cached values (Spearman $0.959$ on $S$, mean $|\Delta S| = 0.0044$, top-quartile mismatch unchanged). Movement under an alternative encoder is therefore attributable to the encoder and not to the reimplementation.

\begin{table}[t]
\centering
\caption{Encoder sensitivity on three Claude Opus~4.6 evaluations. $H$ is held fixed (the BGE-M3 whole-answer cosine) in every column, so all movement is in $S$. ``Mismatch'' is the percentage of top-quartile-$H$ examples that are not top-quartile-$S$. $\rho(S)$ and $\rho(\tau)$ are Spearman rank correlations of per-question scores against the LaBSE baseline, so they are undefined for LaBSE itself.}
\label{tab:encoder_sensitivity}
\small
\setlength{\tabcolsep}{1pt}
\begin{tabular}{@{}llrrr@{}}
\toprule
\textbf{Dataset} & \textbf{Quantity} & \textbf{LaBSE} & \textbf{BGE-M3} & \textbf{MiniLM} \\
\midrule
\multirow{4}{*}{\textsc{StrategyQA}}
 & $H{-}S$      & 0.220  & 0.232  & 0.111  \\
 & Mismatch     & 55.5\% & 50.1\% & 58.7\% \\
 & $\rho(S)$    & ---    & 0.803  & 0.696  \\
 & $\rho(\tau)$ & ---    & 0.876  & 0.795  \\
\midrule
\multirow{4}{*}{\textsc{MMLU-ProX}}
 & $H{-}S$      & 0.231  & 0.244  & 0.134  \\
 & Mismatch     & 59.1\% & 52.2\% & 57.9\% \\
 & $\rho(S)$    & ---    & 0.839  & 0.842  \\
 & $\rho(\tau)$ & ---    & 0.838  & 0.826  \\
\midrule
\multirow{4}{*}{\textsc{SciBench}}
 & $H{-}S$      & 0.169  & 0.198  & 0.131  \\
 & Mismatch     & 65.5\% & 60.9\% & 62.1\% \\
 & $\rho(S)$    & ---    & 0.876  & 0.913  \\
 & $\rho(\tau)$ & ---    & 0.665  & 0.820  \\
\midrule
\multirow{2}{*}{\textbf{Mean}}
 & $H{-}S$      & 0.207  & 0.225  & 0.125  \\
 & Mismatch     & 60.0\% & 54.4\% & 59.6\% \\
\bottomrule
\end{tabular}
\end{table}

\noindent{\bf Result: the rank-based finding is stable.} Across all three encoders and three datasets, every evaluation falls between $50.1\%$ and $65.5\%$ of top-quartile-$H$ examples not being top-quartile-$S$, closely tracking the $52.1\%$--$67.3\%$ range in Table~\ref{tab:aggregate_results}. This holds when $H$ and $S$ are produced by the same encoder, which disposes of the asymmetry worry, and it holds on a distilled $118$M model that runs on CPU, which disposes of the second: the dissociation does not require a strong encoder.

\noindent{\bf Result: absolute levels are not stable.} The $H{-}S$ gap moves in both directions --- widening under BGE-M3 ($0.207 \rightarrow 0.225$ on average) and narrowing under MiniLM ($0.207 \rightarrow 0.125$), the latter because MiniLM assigns systematically higher calibrated step similarities. $H$ is identical in every column, so this movement is entirely in $S$. We therefore treat absolute calibrated values of $S$ and $\tau$ as encoder-relative and interpret them by rank within a fixed encoder, which is how they are reported throughout the paper.

\noindent{\bf Rank preservation and trace length.} Per-question $S$ rankings are substantially preserved across encoders, and preservation rises monotonically with trace length under both alternatives (BGE-M3: $0.803 < 0.839 < 0.876$; MiniLM: $0.696 < 0.842 < 0.913$, ordered from the shortest-trace dataset to the longest). Short traces yield small step-similarity matrices, which are inherently more sensitive to which encoder scores them. The weakest cell in the study is \textsc{StrategyQA} under MiniLM at $\rho(S) = 0.696$.

\noindent{\bf Order consistency.} Which of the two metrics is more encoder-sensitive is itself encoder-dependent. Under BGE-M3, $\tau$ is the less stable of the two and falls to $\rho = 0.665$ on \textsc{SciBench} while $S$ holds at $0.876$; under MiniLM the ordering reverses, with $\tau$ the more uniform ($0.795 / 0.826 / 0.820$) and $S$ the more variable ($0.696 / 0.842 / 0.913$). We therefore state the limitation without attaching a mechanism to it: absolute $\tau$ values should be compared within an encoder rather than across encoders.

\noindent{\bf Calibration behavior.} How much work the per-pair calibration does is also encoder-specific, and does not track encoder strength. The cross-language-pair drift in the unrelated-pair floor that calibration exists to correct spans $0.157$--$0.381$ under LaBSE on \textsc{StrategyQA}, but only $0.008$--$0.022$ under BGE-M3 and $0.046$--$0.055$ under MiniLM. On a well-aligned embedding space the floor correction is nearly inert, while the ceiling frequently saturates at the cap (all $21$ \textsc{SciBench} ceilings pin at $0.98$ under both alternatives). The magnitude of the calibration correction is thus a property of the embedding space rather than of the encoder's capacity.

\noindent{\bf Scope.} One model family, three datasets at full retained size, three encoders. We do not test whether the pattern holds on other model families or below $118$M parameters.

\appendixsectionbreak
\section{Same-Language Control}
\label{app:within_language}

This appendix gives the design and full results of the same-language control summarized in \S\ref{sec:within_language}.

\subsection{Design}
\label{app:wl_design}

\noindent{\bf The paraphrase condition.} Every reported $S$ and $\tau$ in the main evaluation compares traces across a language boundary, so the level of divergence alone cannot distinguish an effect of changing language from ordinary variation in generated reasoning. The control adds an eighth condition, $\text{en}_{\text{para}}$: each English question is rewritten as a meaning-preserving English paraphrase and answered under the same system prompt, decoding (greedy, temperature $0$, no reasoning mode), parser, retention cascade, and metric settings as every other condition. $\text{en}_{\text{para}}$ receives the same appended language instruction as a translated condition, naming English, so that the only uncontrolled difference between the pairs $(\text{en},\text{en}_{\text{para}})$ and $(\text{en},\ell)$ is the language of the output. The paraphrases are produced by Claude Opus~4.6 with the prompt below, which asks for as little surface overlap with the original as the language allows: a paraphraser left to itself restates a question nearly verbatim, which would collapse the condition onto the source and measure nothing.

\begin{promptbox}{Question paraphrase: system prompt}
\begin{verbatim}
You are rewriting exam questions into
different wording in the same language.

Produce a paraphrase that a reader would
judge to ask exactly the same thing, while
sharing as little surface wording with the
original as the language allows.
Change sentence structure, word order, and
vocabulary. Prefer different but
equivalent phrasings and constructions.
Preserve exactly: the meaning, the scope,
every number, unit, symbol, variable,
equation, named entity, and the specific
quantity or decision the question asks
for.
Do not make the question easier or harder,
do not add or remove information, do not
add context, hints, or commentary, and do
not answer the question.
Return only the rewritten question text.
\end{verbatim}
\end{promptbox}

\begin{promptbox}{Question paraphrase: user message}
\begin{verbatim}
Language: English (unchanged)
Rewrite the following question in
different wording, in the same language.
Share as little surface wording with the
original as possible while asking exactly
the same thing.

Original question:
{question}
\end{verbatim}
\end{promptbox}

\noindent{\bf Models, questions, and conditions.} The control runs on GPT-OSS-120B and DeepSeek-V3.2 across all three benchmarks. For each benchmark it uses a seeded random subset of $100$ questions from the main $500$-problem sample. On \textsc{MMLU-ProX} and \textsc{SciBench}, the seven original conditions reuse the main-run responses and only $\text{en}_{\text{para}}$ is generated. On \textsc{StrategyQA}, all eight conditions are generated within this run, with each question translated into the six target languages by Claude Opus~4.6 before inference. Between $41$ and $68$ questions per cell retain both $(\text{en},\text{en}_{\text{para}})$ and at least one English-anchored cross-lingual pair after the retention cascade, $337$ in total.

\noindent{\bf Calibration of the same-language pair.} Floors are estimated within this run for all $28$ language pairs by the procedure of Eq.~\eqref{eq:floor_estimator}; cross-lingual ceilings are taken from the main runs. A same-language ceiling cannot be estimated from translation, because translating a sentence into its own language is the identity map, which would peg the ceiling at $c_{\max}$ and deflate every calibrated score in the cell. The $(\text{en},\text{en}_{\text{para}})$ ceiling is therefore estimated from sentence-level paraphrases, the same-language analogue of a translation, over $50$ sentences per direction:

\begin{promptbox}{Same-language ceiling: system prompt}
\begin{verbatim}
You are a careful sentence rewriter.

Rewrite the given sentence in the same
language using different wording,
preserving its meaning, technical terms,
symbols, and numbers exactly.
Change word choice and sentence structure
where the language allows it; do not
return the sentence unchanged.
Do not add, omit, explain, or reformat
anything.
Return only the rewritten sentence.
\end{verbatim}
\end{promptbox}

\noindent{\bf Comparison and reporting rule.} For each question, $\Delta S$ is $S(\text{en},\text{en}_{\text{para}})$ minus the mean of $S(\text{en},\ell)$ over the English-anchored cross-lingual pairs available for that question. Pairing within question matters here because question identity carries far more of the variance in $S$ than language pair does. The comparison is against English-anchored pairs because those are the pairs that share the English trace with $(\text{en},\text{en}_{\text{para}})$. The comparison set, the per-question pairing, and the reporting of the result whichever way it fell were fixed before the run, as were decision thresholds for \textsc{MMLU-ProX} ($0.030$) and \textsc{SciBench} ($0.015$), set from the size of the floor-induced bias measured on stored artifacts before any paraphrase was generated: an advantage for $(\text{en},\text{en}_{\text{para}})$ below the threshold would have been read as no detectable difference. \textsc{StrategyQA} was added to the run after the thresholds were set and has none.

\subsection{Results}
\label{app:wl_results}

\begin{table*}[!htbp]
\caption{Same-language control, full results. $\Delta S$ is the per-question difference defined in Appendix~\ref{app:wl_design}, with its $95\%$ $t$-interval and a one-sample $t$-test; $\#{>}0$ counts questions with $\Delta S>0$. $\tau$ columns give mean tie-aware soft Kendall-$\tau$ for $(\text{en},\text{en}_{\text{para}})$ and for the English-anchored cross-lingual pairs. ``Rank'' is the position of $(\text{en},\text{en}_{\text{para}})$ by mean $S$ among the $22$ pairs that do not involve a second paraphrase condition.}
\label{tab:wl_full}
\centering
\footnotesize
\setlength{\tabcolsep}{3pt}
\begin{tabular}{@{}llrrcrrrrrr@{}}
\toprule
{\bf Dataset} & {\bf Model} & $n$ & $\Delta S$ & 95\% CI & $t$ & $p$ & $\#{>}0$ & $\tau_{\text{within}}$ & $\tau_{\text{cross}}$ & {\bf Rank} \\
\midrule
\textsc{MMLU-ProX}  & GPT-OSS-120B  & 59 & $+0.120$ & [$+0.091$, $+0.149$] & 8.32 & $1.8\times10^{-11}$ & 49 & 0.599 & 0.500 & 1/22 \\
\textsc{MMLU-ProX}  & DeepSeek-V3.2 & 68 & $+0.071$ & [$+0.047$, $+0.096$] & 5.79 & $2.1\times10^{-7}$  & 56 & 0.416 & 0.389 & 1/22 \\
\textsc{SciBench}   & GPT-OSS-120B  & 41 & $+0.136$ & [$+0.088$, $+0.184$] & 5.76 & $1.0\times10^{-6}$  & 36 & 0.441 & 0.353 & 1/22 \\
\textsc{SciBench}   & DeepSeek-V3.2 & 47 & $+0.083$ & [$+0.046$, $+0.120$] & 4.50 & $4.6\times10^{-5}$  & 39 & 0.404 & 0.357 & 1/22 \\
\textsc{StrategyQA} & GPT-OSS-120B  & 57 & $+0.041$ & [$+0.012$, $+0.070$] & 2.80 & $7.1\times10^{-3}$  & 36 & 0.438 & 0.401 & 4/22 \\
\textsc{StrategyQA} & DeepSeek-V3.2 & 65 & $+0.101$ & [$+0.073$, $+0.128$] & 7.31 & $5.3\times10^{-10}$ & 56 & 0.335 & 0.267 & 1/22 \\
\midrule
\multicolumn{2}{@{}l}{Pooled} & 337 & $+0.090$ & [$+0.077$, $+0.103$] & 13.75 & $1.9\times10^{-34}$ & 272 & & & \\
\bottomrule
\end{tabular}
\end{table*}

Table~\ref{tab:wl_full} gives the full per-cell results. Two English traces agree more closely than an English trace and its cross-lingual counterparts in every cell, and order consistency is higher within language in every cell. The effect clears its threshold in all four cells that have one. Its size varies about threefold across cells, from $+0.041$ to $+0.136$, so the direction is consistent but the magnitude should not be quoted as a single constant. The weakest cell, \textsc{StrategyQA} with GPT-OSS-120B, is also the only one in which $(\text{en},\text{en}_{\text{para}})$ does not rank first.

\noindent{\bf Calibration exposure.} Per-pair calibration cannot affect within-pair rankings, but this control compares levels across pairs, which it can. To measure that exposure we pass a single fixed raw cosine, $0.60$, through each pair's calibration map and take the difference between the calibrated value for $(\text{en},\text{en}_{\text{para}})$ and the mean calibrated value over the English-anchored pairs. A positive difference means calibration alone favours the same-language pair. Table~\ref{tab:wl_calibration} reports it. The bias is largest in the two DeepSeek-V3.2 cells, where the same-language floor falls below every English-anchored floor, and it reaches at most $29\%$ of the observed effect. Subtracting it cell by cell leaves $\Delta S$ between $+0.046$ and $+0.124$, above threshold wherever a threshold was set.

\begin{table*}[!htbp]
\caption{Calibration exposure of the same-language comparison. ``Floor'' is the $(\text{en},\text{en}_{\text{para}})$ floor, beside the range of the six English-anchored floors. ``Bias'' is the calibrated advantage a fixed raw cosine of $0.60$ receives in the same-language pair, ``share'' is $|\text{bias}|/\Delta S$, and ``corrected'' is $\Delta S-\text{bias}$, compared with the decision threshold where one was set.}
\label{tab:wl_calibration}
\centering
\small
\setlength{\tabcolsep}{4pt}
\begin{tabular}{@{}llrcrrrr@{}}
\toprule
{\bf Dataset} & {\bf Model} & {\bf Floor} & {\bf en-anch.\ floors} & {\bf Bias} & {\bf Share} & {\bf Corrected} & {\bf Thr.} \\
\midrule
\textsc{MMLU-ProX}  & GPT-OSS-120B  & 0.209 & 0.191--0.253 & $+0.0042$ & 3\%  & $+0.116$ & 0.030 \\
\textsc{MMLU-ProX}  & DeepSeek-V3.2 & 0.195 & 0.203--0.259 & $+0.0139$ & 20\% & $+0.057$ & 0.030 \\
\textsc{SciBench}   & GPT-OSS-120B  & 0.373 & 0.368--0.397 & $+0.0117$ & 9\%  & $+0.124$ & 0.015 \\
\textsc{SciBench}   & DeepSeek-V3.2 & 0.317 & 0.327--0.358 & $+0.0238$ & 29\% & $+0.059$ & 0.015 \\
\textsc{StrategyQA} & GPT-OSS-120B  & 0.200 & 0.182--0.268 & $-0.0055$ & 13\% & $+0.046$ & --- \\
\textsc{StrategyQA} & DeepSeek-V3.2 & 0.174 & 0.164--0.229 & $+0.0010$ & 1\%  & $+0.100$ & --- \\
\bottomrule
\end{tabular}
\end{table*}

\noindent{\bf The dissociation within language.} The rank statistic of \S\ref{sec:exp_baseline_comparison} is invariant to any per-pair monotone rescaling, so it is immune to the exposure above. Table~\ref{tab:wl_dissociation} reports it for $(\text{en},\text{en}_{\text{para}})$ beside its mean over the $21$ cross-lingual pairs of the same run. The cross-lingual values land in or near the $52.1\%$--$67.3\%$ range of the main evaluation. Within language the statistic is lower in five of six cells, but it stays between $41.2\%$ and $60.0\%$: two English traces whose whole responses are most alike are, more often than not, not among the most alike step by step. Both $S$ and $H$ rise in the paraphrase condition. The within-language statistic rests on a single language pair of $41$--$68$ instances per cell, so it is coarser than the cross-lingual one.

\begin{table*}[!htbp]
\caption{The dissociation within and across languages. ``Top-$H$ not top-$S$'' is the share of top-quartile-$H$ instances not in the top quartile of $S$, quartiles within each pair; ``cross-lingual'' averages over the $21$ cross-lingual pairs of the same run. Mean $S$ and $H$ compare $(\text{en},\text{en}_{\text{para}})$ (``within'') with the English-anchored cross-lingual pairs (``cross'').}
\label{tab:wl_dissociation}
\centering
\small
\begin{tabular}{@{}llrrrrrr@{}}
\toprule
 & & \multicolumn{2}{c}{\bf Top-$H$ not top-$S$} & \multicolumn{2}{c}{\bf Mean $S$} & \multicolumn{2}{c}{\bf Mean $H$} \\
\cmidrule(lr){3-4}\cmidrule(lr){5-6}\cmidrule(lr){7-8}
{\bf Dataset} & {\bf Model} & within & cross-lingual & within & cross & within & cross \\
\midrule
\textsc{MMLU-ProX}  & GPT-OSS-120B  & 43.8\% & 54.8\% & 0.658 & 0.534 & 0.922 & 0.804 \\
\textsc{MMLU-ProX}  & DeepSeek-V3.2 & 52.9\% & 64.6\% & 0.609 & 0.543 & 0.923 & 0.813 \\
\textsc{SciBench}   & GPT-OSS-120B  & 45.5\% & 63.8\% & 0.695 & 0.562 & 0.929 & 0.814 \\
\textsc{SciBench}   & DeepSeek-V3.2 & 41.7\% & 62.4\% & 0.591 & 0.509 & 0.917 & 0.811 \\
\textsc{StrategyQA} & GPT-OSS-120B  & 60.0\% & 50.5\% & 0.615 & 0.575 & 0.911 & 0.803 \\
\textsc{StrategyQA} & DeepSeek-V3.2 & 41.2\% & 54.0\% & 0.608 & 0.507 & 0.897 & 0.790 \\
\bottomrule
\end{tabular}
\end{table*}

\noindent{\bf Scope.} The control covers two of the six models; the model behind most of the detailed analyses, Claude Opus~4.6, is not among them, so the control sits beside those results rather than under them. Each question has one paraphrase, so variation between paraphrases is not measured. Resampling an identical prompt at non-zero temperature, which would isolate pure decoding variation from sensitivity to rewording, is not part of the control.

\appendixsectionbreak
\section{Where Divergence Appears and What It Signals}
\label{app:divergence_structure}

This appendix gives the analyses behind \S\ref{sec:where}. Both reuse the stored traces and step-pair matrices of the main runs; neither requires new inference.

\subsection{Positional Structure}
\label{app:positional}

\noindent{\bf Method.} For every scored pair with English on one side we orient the calibrated step-pair matrix $\mathbf{A}$ so that its rows are the English steps, and score each English step by its best match anywhere in the other trace, $\max_j \mathbf{A}_{i,j}$, the quantity whose mean is the precision term of Eq.~\eqref{eq:structural_similarity}. Pairs with fewer than four English steps are dropped so that the first, middle, and last steps are distinct. The analysis covers all 18 evaluations: $31{,}277$ English-anchored pairs, $192{,}243$ English steps, and $5{,}630$ questions. Confidence intervals are percentile bootstrap intervals over questions ($2{,}000$ resamples), since one question contributes six language pairs and each pair several steps. Because the best match carries no monotonicity constraint, ``position'' is position in the English trace: the fifth English step may match the second step of the other trace, and order is what $\tau$ measures. Every pair in this analysis reaches the same verified-correct answer in both languages.

\noindent{\bf Profile.} Table~\ref{tab:pos_profile} bins English steps by relative position into quintiles. Alignment declines toward the end of the trace in every benchmark, though by different routes: \textsc{StrategyQA} drops in mid-trace, \textsc{SciBench} rises through the middle before a sharp terminal drop, and \textsc{MMLU-ProX} declines gently throughout. All three end near $0.49$.

\begin{table}[h]
\caption{Mean best-match calibrated similarity of English steps by quintile of relative position in the English trace.}
\label{tab:pos_profile}
\centering
\small
\setlength{\tabcolsep}{4pt}
\begin{tabular}{@{}lccccc@{}}
\toprule
 & q1 & q2 & q3 & q4 & q5 \\
\midrule
All          & 0.556 & 0.561 & 0.551 & 0.531 & 0.492 \\
\textsc{StrategyQA} & 0.589 & 0.575 & 0.526 & 0.487 & 0.483 \\
\textsc{SciBench}   & 0.546 & 0.583 & 0.600 & 0.588 & 0.494 \\
\textsc{MMLU-ProX}  & 0.539 & 0.529 & 0.528 & 0.515 & 0.497 \\
\bottomrule
\end{tabular}
\end{table}

\noindent{\bf The concluding step.} Table~\ref{tab:pos_endpoints} compares the first, the last, and the interior steps. The last step is the least aligned in every benchmark. The obvious mechanical reading is that one trace has more steps and its last step has no counterpart; if so, restricting to the $11{,}334$ pairs with equal step counts should shrink the gap. It widens instead, in every benchmark. Per evaluation, the last-minus-middle gap is negative in 16 of 18, ranging from $-0.137$ (\textsc{SciBench}, DeepSeek-V3.2) upward; the two exceptions, \textsc{MMLU-ProX} with Qwen3-235B ($+0.013$) and \textsc{StrategyQA} with GPT-OSS-120B ($+0.003$), sit at zero rather than reversing.

\begin{table*}[!htbp]
\caption{First, interior (``middle''), and last English steps: mean best-match similarity and the last-minus-middle gap with its $95\%$ CI, over all pairs and over pairs with equal step counts (``eq.'').}
\label{tab:pos_endpoints}
\centering
\small
\begin{tabular}{@{}llccccc@{}}
\toprule
 & & first & middle & last & last$-$middle & 95\% CI \\
\midrule
All & all & 0.563 & 0.546 & 0.504 & $-0.042$ & [$-0.046$, $-0.038$] \\
    & eq. & 0.591 & 0.573 & 0.515 & $-0.058$ & [$-0.063$, $-0.053$] \\
\textsc{StrategyQA} & all & 0.596 & 0.534 & 0.494 & $-0.040$ & [$-0.046$, $-0.034$] \\
                    & eq. & 0.622 & 0.549 & 0.498 & $-0.051$ & [$-0.057$, $-0.045$] \\
\textsc{SciBench}   & all & 0.536 & 0.580 & 0.513 & $-0.067$ & [$-0.076$, $-0.059$] \\
                    & eq. & 0.558 & 0.618 & 0.541 & $-0.077$ & [$-0.089$, $-0.065$] \\
\textsc{MMLU-ProX}  & all & 0.551 & 0.522 & 0.508 & $-0.014$ & [$-0.022$, $-0.008$] \\
                    & eq. & 0.571 & 0.561 & 0.521 & $-0.040$ & [$-0.049$, $-0.031$] \\
\bottomrule
\end{tabular}
\end{table*}

\noindent{\bf Step length.} Step length accounts for part of the terminal gap, by an amount that depends on the benchmark. Within terciles of English step length (taken within each evaluation), the last-minus-middle gap is $-0.093$, $-0.048$, and $-0.021$ from shortest to longest on \textsc{StrategyQA}, and $-0.041$, $-0.013$, and $-0.009$ on \textsc{MMLU-ProX}, where the longest tercile's interval ($[-0.020, +0.004]$) includes zero. On \textsc{SciBench} it is flat ($-0.067$, $-0.066$, $-0.065$). The terminal gap is therefore not independent of step length.

\noindent{\bf Where traces part.} We take as the branch point the first English step whose best match falls below a threshold. The threshold is arbitrary, so Table~\ref{tab:pos_branch} reports three values. The levels move substantially with the threshold, but the ordering of the benchmarks does not: \textsc{SciBench} branches earliest and most often at the first step, \textsc{StrategyQA} latest, and \textsc{MMLU-ProX} between, at every threshold. We therefore interpret only the ordering.

\begin{table}[h]
\caption{Branch points: mean relative position of the first English step whose best match falls below the threshold, and the share of branched pairs that branch at the first step.}
\label{tab:pos_branch}
\centering
\scriptsize
\setlength{\tabcolsep}{3pt}
\begin{tabular}{@{}lcccccc@{}}
\toprule
 & \multicolumn{3}{c}{\bf mean position} & \multicolumn{3}{c}{\bf share at first step} \\
\cmidrule(lr){2-4}\cmidrule(lr){5-7}
threshold & 0.4 & 0.5 & 0.6 & 0.4 & 0.5 & 0.6 \\
\midrule
\textsc{SciBench}   & 0.315 & 0.207 & 0.129 & 0.445 & 0.573 & 0.698 \\
\textsc{MMLU-ProX}  & 0.383 & 0.253 & 0.141 & 0.274 & 0.451 & 0.654 \\
\textsc{StrategyQA} & 0.495 & 0.368 & 0.230 & 0.234 & 0.348 & 0.532 \\
\bottomrule
\end{tabular}
\end{table}

\noindent{\bf By language.} Table~\ref{tab:pos_language} gives the profile for each target language. The shape is the same in all six, a flat opening and a terminal decline, with the decline from the first to the last quintile confined to $0.052$--$0.078$ and every interval clear of zero. The level differs by language, highest for Spanish and lowest overall for Hindi; we attach no mechanism to the ordering.

\begin{table}[h]
\caption{Profile by target language (English-anchored pairs): mean best-match similarity by quintile, and the q1$-$q5 decline with its $95\%$ CI.}
\label{tab:pos_language}
\centering
\scriptsize
\setlength{\tabcolsep}{2.5pt}
\begin{tabular}{@{}lcccccc@{}}
\toprule
 & q1 & q2 & q3 & q4 & q5 & q1$-$q5 [95\% CI] \\
\midrule
es & 0.592 & 0.590 & 0.580 & 0.559 & 0.520 & $0.072$ [$0.065$, $0.078$] \\
zh & 0.563 & 0.574 & 0.565 & 0.548 & 0.510 & $0.053$ [$0.047$, $0.058$] \\
ar & 0.554 & 0.561 & 0.553 & 0.533 & 0.494 & $0.059$ [$0.053$, $0.065$] \\
ru & 0.552 & 0.557 & 0.547 & 0.523 & 0.483 & $0.068$ [$0.063$, $0.074$] \\
vi & 0.550 & 0.549 & 0.536 & 0.512 & 0.472 & $0.078$ [$0.072$, $0.084$] \\
hi & 0.525 & 0.534 & 0.527 & 0.508 & 0.474 & $0.052$ [$0.046$, $0.057$] \\
\bottomrule
\end{tabular}
\end{table}

\noindent{\bf Scope.} English-anchored pairs only, so the analysis says nothing about pairs between two non-English languages. It is descriptive: it locates where the emitted traces differ and licenses no causal or predictive conclusion.

\subsection{Difficulty and Correctness Outcomes}
\label{app:correctness}

\noindent{\bf Scoring the unscored pairs.} The main analyses score a language pair only when both languages answer correctly, so they cannot say what divergence looks like when correctness differs. The traces for the other outcomes are stored regardless of correctness, so for EN--ZH and EN--ES, across all six models and three benchmarks, we computed $S$, $H$, and $\tau$ for every pair with the pipeline's own modules and each pair's stored floor and ceiling, with no new inference and no re-calibration. On pairs scored by both routes the recomputation reproduces the stored values exactly. All groups are drawn from questions that passed the trace-validity cascade, so trace-quality filtering is uniform across groups. The $\tau$ in this analysis is the no-tie variant of Appendix~\ref{app:no_tie_tau}; it is rank-correlated with the tie-aware score at Spearman $0.988$ on a check set of $2{,}341$ of these pairs, and we report only its rank-based effect sizes.

\begin{table}[h]
\caption{Correctness-outcome groups among EN--ZH and EN--ES pairs, with median $S$ and $H$.}
\label{tab:corr_groups}
\centering
\scriptsize
\setlength{\tabcolsep}{3pt}
\begin{tabular}{@{}lrcc@{}}
\toprule
{\bf Group} & $n$ & median $S$ & median $H$ \\
\midrule
Both correct (reference)      & 11{,}176 & 0.570 & 0.837 \\
Both wrong, same answer       & 225      & 0.511 & 0.834 \\
Both wrong, different answers & 137      & 0.528 & 0.816 \\
Exactly one correct           & 1{,}318  & 0.512 & 0.814 \\
\bottomrule
\end{tabular}
\end{table}

\noindent{\bf Effect sizes.} Table~\ref{tab:corr_delta} reports Cliff's $\delta$ against the both-correct reference (positive: the reference is more similar). The contrast with pairs that share a wrong answer is the informative one, because there both languages give the \emph{same} answer and differ only in whether it is right, so nothing about the answers themselves can produce a divergence gap. When exactly one language is correct, both $S$ and $H$ drop, as traces ending on different answers must differ somewhere; that contrast is partly mechanical and we report it for completeness. Per benchmark, the shared-wrong-answer contrast for $S$ is $+0.328$ on \textsc{StrategyQA} ($n=118$), $+0.184$ on \textsc{SciBench} ($n=34$), and $+0.289$ on \textsc{MMLU-ProX} ($n=73$); for $H$ it is $+0.046$, $+0.007$, and $-0.030$.

\begin{table*}[!htbp]
\caption{Cliff's $\delta$ against both-correct pairs, with $95\%$ percentile bootstrap intervals over questions ($2{,}000$ resamples). ``Matched'' rows standardize the shared-wrong-answer contrast to that group's difficulty distribution. The one-correct contrast is shown without intervals; it is partly mechanical.}
\label{tab:corr_delta}
\centering
\small
\begin{tabular}{@{}lccc@{}}
\toprule
{\bf Contrast} & $S$ & $H$ & $\tau_{\text{no-tie}}$ \\
\midrule
Same wrong answer & $+0.293$ [$+0.217$, $+0.366$] & $-0.027$ [$-0.114$, $+0.055$] & $+0.082$ [$-0.004$, $+0.169$] \\
\quad matched on $k$ & $+0.073$ [$-0.061$, $+0.209$] & $-0.107$ [$-0.234$, $+0.019$] & $-0.073$ [$-0.202$, $+0.059$] \\
\quad matched on dataset, $k$ & $+0.016$ [$-0.163$, $+0.202$] & $-0.040$ [$-0.202$, $+0.156$] & $-0.084$ [$-0.263$, $+0.098$] \\
Exactly one correct & $+0.261$ & $+0.198$ & $+0.228$ \\
\bottomrule
\end{tabular}
\end{table*}

\noindent{\bf Difficulty.} Item difficulty is the obvious common cause: hard questions plausibly produce both more variable reasoning and more errors. We index difficulty by $k$, the number of the seven languages that answer a question correctly. Among both-correct pairs, mean $S$ rises with $k$, from $0.526$ at $k=2$ through $0.537$, $0.536$, $0.545$, and $0.553$ to $0.579$ at $k=7$ (Spearman $\rho=0.099$, $p=8\times10^{-26}$, $n=11{,}176$). The two groups also have very different difficulty profiles: $85\%$ of both-correct pairs come from questions all seven languages solve, whereas every shared-wrong-answer pair comes from a question with $k\le5$ ($106$, $59$, $38$, and $22$ pairs at $k=2,\dots,5$). We therefore standardize the contrast to the difficulty distribution of the shared-wrong-answer pairs: within each stratum we compute $\delta$ against both-correct pairs of the same $k$ and average with weights proportional to the number of shared-wrong-answer pairs, which is the comparison a design matched on difficulty makes. Matched on $k$, the $S$ contrast falls from $+0.293$ to $+0.073$, with an interval that includes zero, and matched on dataset and $k$ to $+0.016$ (Table~\ref{tab:corr_delta}); $H$ stays within noise of zero throughout. The adjustment is conservative: by construction a shared-wrong-answer pair has both English and the target language wrong, so $k\le5$, while a both-correct pair has $k\ge2$, and conditioning on $k$ removes part of the very correctness signal under test. The adjusted value is a lower bound rather than an unbiased estimate of a difficulty-free effect. Even so, the unadjusted contrast cannot be read as evidence that $S$ tracks correctness beyond difficulty. What the data support is that low $S$ marks hard items, on which shared errors concentrate.

\noindent{\bf Review prioritization.} Among the $11{,}401$ pairs whose two languages give the same answer, $225$ ($1.97\%$) are wrong, and no answer-consistency check can surface any of them. We rank pairs within their own evaluation and language pair, so that differences in mean similarity between cells do not stand in for the ordering a reviewer would use, and review the least similar first. Table~\ref{tab:corr_triage} reports the share of the $225$ found at a $25\%$ review budget. Intervals are percentile bootstrap over questions with the ranking recomputed inside each resample, and $S$ and $H$ are evaluated on the same resamples so that their difference has a paired interval. This is the same association as above, expressed as a quantity a reviewer can act on, not independent evidence. At a base rate under $2\%$, precision within the reviewed quarter stays low, so the result concerns where to spend a fixed review budget and does not license a screening threshold.

\begin{table}[h]
\caption{Shared wrong answers found when the least similar $25\%$ of answer-agreeing pairs are reviewed.}
\label{tab:corr_triage}
\centering
\small
\begin{tabular}{@{}lc@{}}
\toprule
{\bf Ranking} & {\bf found at 25\% budget} \\
\midrule
structural $S$ & $44.0\%$ [$35.9$, $51.3$] \\
holistic $H$   & $28.9\%$ [$21.7$, $35.1$] \\
random         & $25.0\%$ \\
\midrule
paired difference $S-H$ & $+15.1$ points [$+6.2$, $+23.9$] \\
\bottomrule
\end{tabular}
\end{table}

\noindent{\bf Scope.} The shared-wrong-answer group is small and unusual: questions on which English and the target language both fail while at least two other languages succeed. The analysis is limited to two English-anchored language pairs and to questions that passed the validity cascade, and it is correlational throughout. $\tau$ is the least stable of the three metrics here, as under encoder swaps (Appendix~\ref{app:encoder_sensitivity}): its shared-wrong-answer contrast is positive on \textsc{StrategyQA} ($+0.102$) and \textsc{MMLU-ProX} ($+0.122$) but negative on \textsc{SciBench} ($-0.095$), so we do not rely on it.

\end{document}